\documentclass[10pt]{article} 
\usepackage[accepted]{tmlr}

\usepackage{amsmath,amsfonts,bm}

\def\eqref#1{equation~\ref{#1}}

\def\1{\bm{1}}

\DeclareMathAlphabet{\mathsfit}{\encodingdefault}{\sfdefault}{m}{sl}
\SetMathAlphabet{\mathsfit}{bold}{\encodingdefault}{\sfdefault}{bx}{n}

\definecolor{CitationColor}{RGB}{0, 0, 150}
\definecolor{LinkColor}{RGB}{200, 0, 0}
\usepackage{threeparttable}
\usepackage{url}
\usepackage{hyperref}
\usepackage{makecell}
\hypersetup{colorlinks=true, citecolor=CitationColor, 
    linkcolor=LinkColor,
    urlcolor=CitationColor,
    pdfpagemode=FullScreen,linktocpage=true}
\usepackage{graphicx}
\usepackage{booktabs}
\usepackage{longtable}
\usepackage{adjustbox}
\usepackage[table]{xcolor}
\usepackage{array}
\usepackage{ragged2e}
\usepackage{tabularx}
\usepackage{pdflscape}
\usepackage{tikz}
\usetikzlibrary{arrows.meta}
\usepackage{lmodern}
\usepackage[T1]{fontenc}
\usepackage{pifont}
\newcommand{\cmark}{\ding{51}} 
\newcommand{\xmark}{\ding{55}} 
\newcommand{\circmark}{\ding{109}}
\usepackage{forest}
\usepackage[dvipsnames, svgnames]{xcolor}
\usepackage{amssymb}
\usetikzlibrary{arrows.meta}

\definecolor{AgnosticDark}{RGB}{180, 30, 45}      
\definecolor{AgnosticLight}{RGB}{255, 210, 215}
\definecolor{BayesianDark}{RGB}{31, 73, 125}      
\definecolor{BayesianLight}{RGB}{189, 215, 238}
\definecolor{ParametricDark}{RGB}{197, 90, 17}  
\definecolor{ParametricLight}{RGB}{252, 228, 214}
\definecolor{DistRegDark}{RGB}{84, 130, 53}       
\definecolor{DistRegLight}{RGB}{226, 239, 218}
\definecolor{GenerativeDark}{RGB}{110, 40, 180}   
\definecolor{FoundationDark}{RGB}{64,64,64}
\definecolor{GenerativeLight}{RGB}{234, 217, 243}
\definecolor{ReservoirDark}{RGB}{92, 64, 51}    
\definecolor{ReservoirLight}{RGB}{234, 222, 201}
\definecolor{Decision}{HTML}{FFF8DC}
\definecolor{FoundationLight}{RGB}{220,220,220}

\newcommand{\rfoundation}
{\rowcolor{FoundationLight}}
\newcommand{\ragnostic}{\rowcolor{AgnosticLight}}
\newcommand{\rbayesian}{\rowcolor{BayesianLight}}
\newcommand{\rparametric}{\rowcolor{ParametricLight}}
\newcommand{\rdistreg}{\rowcolor{DistRegLight}}
\newcommand{\rgenerative}{\rowcolor{GenerativeLight}}

\definecolor{AgnosticRoot}  {RGB}{217, 142, 150}
\definecolor{Agnostic}  {RGB}{255, 210, 215}
\definecolor{AgnosticLeaf}  {RGB}{255, 235, 238}
\definecolor{Intrinsic}  {RGB}{250, 235, 195}
\definecolor{BayesianRoot}  {RGB}{145, 178, 208}
\definecolor{Bayesian}  {RGB}{189, 215, 238}
\definecolor{BayesianLeaf}  {RGB}{235, 242, 250}
\definecolor{BayesianLeafL}  {RGB}{245, 248, 252}
\definecolor{ParametricRoot}  {RGB}{244, 203, 179}
\definecolor{Parametric}  {RGB}{252, 228, 214}
\definecolor{DistRegRoot}  {RGB}{195, 215, 183}
\definecolor{DistReg}  {RGB}{226, 239, 218}
\definecolor{GenerativeRoot}  {RGB}{206, 180, 220}
\definecolor{Generative}  {RGB}{234, 217, 243}
\definecolor{GenerativeLeaf}  {RGB}{250, 244, 253}
\definecolor{EvaluationRoot}  {RGB}{175, 155, 125}
\definecolor{Evaluation}  {RGB}{234, 222, 201}
\definecolor{EvaluationLeaf}  {RGB}{252, 248, 242}
\definecolor{DemoRoot}{RGB}{230, 248, 245}
\definecolor{DemoLeaf}{RGB}{246, 252, 251}
\definecolor{ChallengeRoot}  {RGB}{210,210,210}
\definecolor{Challenge}  {RGB}{240, 240, 240}
\definecolor{Cedge}   {RGB}{60,60,60}

\forestset{
  guidance tree/.style={
    for tree={
      grow=south,
      parent anchor=south,
      child anchor=north,
      anchor=center,
      align=center,
      draw=Cedge,
      line width=0.5pt,
      rounded corners=1.5pt,
      font=\LARGE,
      inner sep=2pt,
      l sep=50mm,
      s sep=2mm,
    },
    edge path={
      \noexpand\path[\forestoption{edge}]
      (!u.parent anchor) -- ++(0,-3mm) -| (.child anchor)
      \forestoption{edge label};
    },
  },
}

\title{Beyond Point Forecasts: A Survey on Probabilistic Forecasting for Time Series and Spatiotemporal Data}

\author{\name Donia Besher$^{1,*}$ \email A00021999@sorbonne.ae \AND
      \name Rajdeep Pathak$^{1,2,*}$ \email rajdeep.pathak@sorbonne.ae \AND
      \name Madhurima Panja$^1$ \email madhurima.panja@sorbonne.ae \AND
      \name Tanujit Chakraborty$^{1,2}$ \email tanujit.chakraborty@sorbonne.ae \AND
      \addr $^1$SAFIR, Sorbonne University Abu Dhabi, United Arab Emirates \\
      \addr $^2$SCAI, Sorbonne Universit\'{e}, Paris, France\\
      \addr $^*$Joint first authors\\}

\def\month{11}  
\def\year{2026} 
\def\openreview{\url{https://openreview.net/forum?id=F140ECa1HO}} 

\begin{document}

\maketitle

\begin{abstract}
Probabilistic forecasting is central to decision-making under uncertainty, yet its methodological landscape has become increasingly fragmented across temporal and spatiotemporal forecasting, statistical modeling, machine learning, and deep generative modeling. This survey develops a unified perspective by organizing probabilistic forecasting methods according to where and how uncertainty is introduced into the forecasting pipeline. Our taxonomy connects model-agnostic approaches including ensembles and distribution-free calibration, with model-intrinsic approaches spanning Bayesian modeling, parametric predictive distributions, distributional regression, and modern generative models, and further examines the emerging role of time series foundation models. Beyond methodological synthesis, we identify the assumptions, computational demands, and forms of uncertainty represented by different paradigms, and translate these distinctions into data-driven and domain-specific guidance for method selection. We complement the survey with a cross-paradigm empirical study on univariate, multivariate, and spatiotemporal forecasting tasks. The results reveal that no single uncertainty-quantification paradigm dominates across settings. Calibration, sharpness, predictive accuracy, and computational efficiency can lead to substantially different model preferences, while expressive generative models and zero-shot foundation models exhibit markedly different accuracy-efficiency trade-offs. Lastly, we identify unresolved challenges surrounding uncertainty in evolving dependency structures, physics-informed predictive distributions, forecasting extreme events, handling count-valued, directional, and continuous-time series, and the development of unified software resources. Our survey provides both a conceptual framework and a practical roadmap for probabilistic forecasting research. Code and supporting materials are available at \url{https://github.com/PyCoder913/ProbForecasting}.

\end{abstract}

\tableofcontents
\newpage

\section{Introduction}
Forecasting plays a key role in decision-making across epidemiology, energy systems, transportation, finance, and environmental monitoring \citep{hyndman2021forecasting}. In many of these application domains, however, a single prediction of the future is insufficient as decisions depend critically on the uncertainty surrounding that prediction \citep{gneiting2014probabilistic}. For instance, health authorities planning hospital capacity require reliable uncertainty estimates for future case counts \citep{stojanovic2019bayesian, karami2026comparative}; power-system operators must account for uncertainty when balancing supply and demand \citep{hong2016probsurvey, xie2023probabilistic}; financial risk management depends on the tails of predictive distributions rather than expected outcomes \citep{adrian2019vulnerable, blasco2024survey}; environmental and hydrological applications require reliable assessment of rare events such as floods \citep{pasche2024eqrn, Papacharalampous2022hydrological}; and transportation systems rely on uncertainty-aware forecasts of congestion and travel times \citep{cheng2024recent, pal2021rnnpf}. These requirements motivate probabilistic forecasting, whose objective is to characterize the range and likelihood of possible future outcomes instead of providing a single-point prediction. 


These challenges become substantially more pronounced when moving from time series to spatiotemporal forecasting. Even in purely temporal settings, predictive uncertainty is shaped by serial dependence, nonstationarity, and the propagation of uncertainty across forecast horizons. Spatiotemporal problems additionally require accounting for interactions across variables and locations, where the underlying dependency structure may be complex, asymmetric, time-varying, and influenced by external drivers. 
In practice, these difficulties are further compounded by 
rare but consequential extremes that may invalidate conventional distributional assumptions \citep{coles2001extremes}. 
Probabilistic forecasting in such contexts must therefore 
characterize uncertainty while respecting the temporal and spatial structure of the data. Their quality is consequently governed by the complementary principles of calibration and sharpness, which require predictive distributions to be statistically reliable yet sufficiently informative \citep{gneiting2007probForecastandCalibration}. These challenges have motivated a broad and rapidly evolving range of probabilistic forecasting methodologies.

\begin{figure}[p]         
\centering
\resizebox{\textwidth}{!}{
\begin{forest}
  for tree={
    grow'           = 0,
    draw            = Cedge,
    line width      = 0.65pt,
    rounded corners = 2pt,
    parent anchor   = east,      
    child anchor    = west,       
    anchor          = west,
    edge path={
      \noexpand\path[\forestoption{edge}] 
      (!u.parent anchor) -- ++(4mm,0) |- (.child anchor); 
    }, 
    font            = \rmfamily\scriptsize,
    align           = center,
    inner xsep      = 5pt,
    inner ysep      = 3pt,
    outer sep       = 0pt,
    l sep           = 10mm,       
    s sep           = 2.7mm,      
  },
  where level=0{
    text width=2.7cm,
    font=\rmfamily\large\bfseries,
    inner ysep=8pt,
    parent anchor = east,         
  }{},
  where level=1{
    font=\rmfamily\normalsize\bfseries,
    text width=3.1cm, inner ysep=5pt,
  }{},
  where level=2{
    font=\rmfamily\normalsize\bfseries,
    text width=4.5cm, inner ysep=4pt,
  }{},
  where level=3{
    font=\rmfamily\small\bfseries,
    text width=8.2cm, inner ysep=3pt,
  }{},
  where level=4{
    font=\rmfamily\small\bfseries,
     text width=3.8cm, inner ysep=3pt,
  }{},
  where level=5{
    font=\rmfamily\small\bfseries,
     text width=2.5cm, inner ysep=3pt,
  }{},
[{Probabilistic\\ Forecasting}, align=center
  [{Model-Agnostic\\Methods (Sec.~\ref{sec_model_agnostic})}, fill=AgnosticRoot
    [{Pre-Control Limits \\(Sec.~\ref{sec_precontrol_limits})}, fill=Agnostic,
    ]
    [{Ensemble-Based Methods\\ (Sec.~\ref{sec_ensemble})}, fill=Agnostic
      [{Resampling-Based Ensembles (Sec.~\ref{sec_resampling_based})}, fill=AgnosticLeaf]
      [{Randomization-Based Ensembles (Sec.~~\ref{sec_randomization_based})}, fill=AgnosticLeaf]
      [{Deep Ensembles (Sec.~~\ref{sec_deep_ensemble})}, fill=AgnosticLeaf]
      [{Bayesian Model Averaging (Sec.~\ref{sec_bma})}, fill=AgnosticLeaf]
    ]
    [{Distribution-Free \\Calibration (Sec.~~\ref{sec_distribution_free})}, fill=Agnostic
      [{Conformal Prediction  (Sec.~\ref{sec:agnostic_conformal})}, fill=AgnosticLeaf]
      [{Conformalized Quantile Regression (Sec.~\ref{sec_cqr})}, fill=AgnosticLeaf]
    ]
  ]
  [{Model-Intrinsic\\Methods (Sec.~\ref{sec_model_intrinsic})}, fill=Intrinsic
    [{Bayesian Modeling\\ (Sec.~\ref{sec_bayesian})}, fill=BayesianRoot
      [{Classical Bayesian Models (Sec.~\ref{sec_classical_bayesian})}, fill=Bayesian
        [{Bayesian Hierarchical\\ Models (Sec.~\ref{sec_bhm})}, fill=BayesianLeaf]
        [{Bayesian State Space\\ Models  (Sec.~\ref{sec_bayesian_state_space})}, fill=BayesianLeaf]
      ]
      [{Bayesian Nonparametric Methods (Sec.~\ref{sec_bayesian_nonparametric})}, fill=Bayesian
        [{Gaussian Processes\\ (Sec.~\ref{sec_gp})}, fill=BayesianLeaf]
        [{Dirichlet Processes\\ (Sec.~\ref{sec_Dirichlet})}, fill=BayesianLeaf]
        [{Bayesian Additive \\ Regression Trees\\ (Sec.~\ref{sec_BART})}, fill=BayesianLeaf]
      ]
      [{Bayesian Neural Networks (Sec.~\ref{sec_bnn})}, fill=Bayesian]
        [{Approximate Bayesian Inference (Sec.~\ref{sec_approximate_bayesian})}, fill=Bayesian
          [{Markov Chain Monte\\ Carlo (Sec.~\ref{sec_approximate_mcmc})}, fill=BayesianLeaf],
          [{Monte Carlo Dropout\\ (Sec.~\ref{sec_approximate_drop})}, fill=BayesianLeaf],
          [{Variational Inference\\ (Sec.~\ref{sec_approximate_vi})}, fill=BayesianLeaf]
          [{Integrated Nested \\ Laplace Approximation \\(Sec.~\ref{sec_approximate_inla})}, fill=BayesianLeaf]
        ]
    ]
    [{Parametric Predictive \\ Distributions (Sec.~\ref{sec_parametric})}, fill=ParametricRoot
      [{Single Parametric Distributions (Sec.~\ref{sec_parametric_single})}, fill=Parametric]
      [{Mixture Density Networks (Sec.~\ref{sec_parametric_mdn})}, fill=Parametric]
    ]
    [{Distributional Regression\\ (Sec.~\ref{sec_distributional})}, fill=DistRegRoot
      [{Quantile Regression (Sec.~\ref{sec_distributional_quantile})}, fill=DistReg]
      [{Engression (Sec.~\ref{sec_distributional_engression})}, fill=DistReg]
      [{Loss-Driven Interval Estimation (Sec.~\ref{sec_loss_driven})}, fill=DistReg]
    ]
    [{Generative Models \\(Sec.~\ref{sec_generative})}, fill=GenerativeRoot, 
      [{Copula-Based Models (Sec.~\ref{sec_generative_copula})}, fill=Generative
      ] 
      [{Hidden Markov Models (Sec.~\ref{sec_generative_hmm})}, fill=Generative
      ]
      [{Hierarchical Discretization (Sec.~~\ref{sec_distributional_discretization})}, fill=Generative]
      [{Variational Autoencoders (Sec.~\ref{sec_generative_vae})}, fill=Generative
      ]
      [{Generative Adversarial Networks (Sec.~\ref{sec_generative_gan})}, fill=Generative
      ]
      [{Normalizing Flows (Sec.~\ref{sec_generative_nf})}, fill=Generative]
      [{Diffusion Models  (Sec.~\ref{sec_generative_diffusion})}, fill=Generative
        [{Denoising Diffusion\\ Probabilistic Models\\ (Sec.~\ref{sec_diffusion_ddpm})}, fill=GenerativeLeaf]
        [{Score-Based Diffusion\\ Models (Sec.~\ref{sec_diffusion_score})}, fill=GenerativeLeaf]
        [{Latent Diffusion\\ Models (Sec.~\ref{sec_diffusion_ldm})}, fill=GenerativeLeaf]
      ]
      [{Parametric Prior Mapping (Sec.~\ref{sec_generative_prior})}, fill=Generative
      ]   
      [{Probabilistic Scenarios (Sec.~\ref{sec_generative_scenario})}, fill=Generative
      ]   
      ]      
  ]
  [{Foundation \\ Models (Sec.~\ref{sec_foundation_models})}, fill=Challenge
  ]
]
\end{forest}
}
\caption{Taxonomy and overview of the survey on probabilistic forecasting methods for time series and spatiotemporal data.}
\label{fig:Methods_Diagram}
\end{figure}
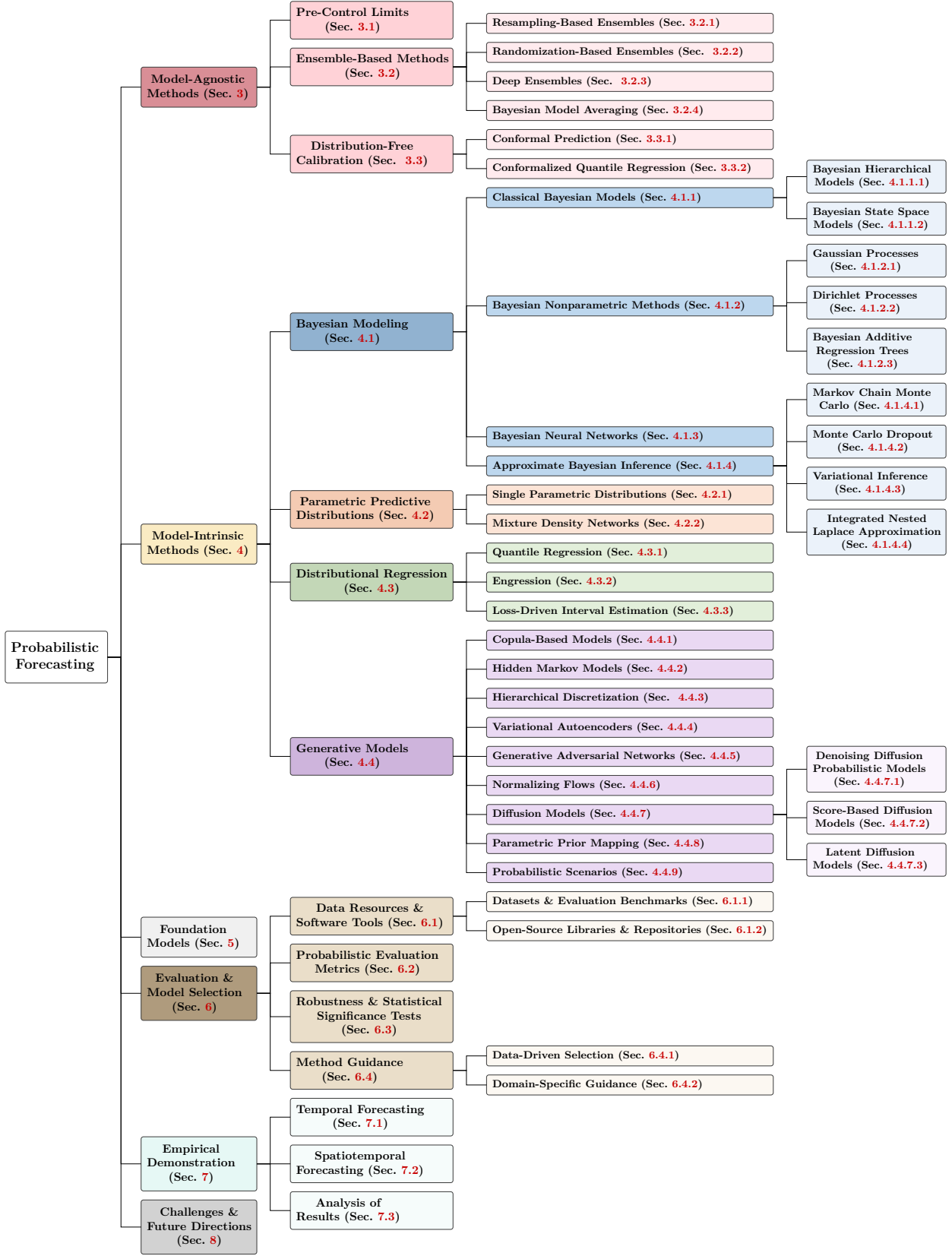

We organize these methods into two overarching paradigms according to \emph{where and how uncertainty is introduced into the forecasting pipeline}: \emph{model-agnostic} and \emph{model-intrinsic} approaches. Model-agnostic methods quantify uncertainty around the predictions of an existing forecasting model without fundamentally modifying its architecture or learning objective. Early approaches often relied on independent and identically distributed (i.i.d.) Gaussian residuals and fixed-width prediction intervals \citep{panja2023epicasting}, whereas subsequent developments introduced resampling, ensemble aggregation \citep{elliott2006handbook}, and distribution-free calibration through conformal prediction \citep{angelopoulos2023conformal}. In contrast, model-intrinsic methods learn predictive uncertainty as part of the forecasting model itself. These approaches range from parametric predictive distributions \citep{salinas2020deepar} and Bayesian formulations \citep{qiu2018mbsts,coscia2025barnn} to distributional regression \citep{koenker1978regression} and increasingly expressive generative models \citep{rasul2021timegrad,rasul2021multivariate}. More recently, time series foundation models have introduced a new setting in which probabilistic forecasts may be produced with little or no task-specific training, raising important questions regarding calibration, transferability, and their extension to structured spatiotemporal systems.

Despite this methodological breadth, the corresponding survey literature remains surprisingly fragmented. Foundational work has established key principles of probabilistic forecasting, including calibration, sharpness, scoring rules, and distributional regression \citep{gneiting2014probabilistic}. However, much of the existing literature focuses on specific model families \citep{gao2022gan, Chakraborty2024gan, yang2026diffusion}; examines uncertainty quantification within the broader context of deep learning, with limited attention to temporal and spatial dependence \citep{abdar2021review, gawlikowski2023survey}; reviews forecasting architectures without treating probabilistic forecasting and uncertainty quantification as their primary focus \citep{lim2021forecasting, benidis2022deeplearning, wang2022deeplearning}; or concentrates on particular application domains \citep{yuan2021survey, kaur2022energy, Papacharalampous2022hydrological, xie2023probabilistic, rickard2026financial}.
As detailed in Section~\ref{sec_literature_review}, these perspectives provide substantial depth individually, but none offers a unified, architecture-agnostic treatment of probabilistic forecasting that simultaneously spans temporal and spatiotemporal settings. As a result, developments that address the same underlying forecasting problem often remain separated across statistical, machine learning, spatiotemporal, and application-specific literatures.

This survey aims to bridge these literatures. Rather than organizing methods primarily by forecasting architecture or application domain, we classify them according to the mechanism through which predictive uncertainty is represented and estimated. The resulting taxonomy, illustrated in Fig.~\ref{fig:Methods_Diagram}, connects model-agnostic and model-intrinsic approaches for uncertainty quantification. 
Representative methods are summarized in Table~\ref{tab:UQ_Methods_Summary}. Importantly, our objective is not merely to catalogue existing techniques, but to analyze the assumptions and trade-offs that distinguish them. Specifically, we analyze the type of uncertainty they represent, the structural assumptions they impose, their computational requirements, calibration properties, and the forecasting domains in which they are most appropriate. Our main contributions are summarized as follows:

\begin{itemize}

    \item \textbf{A unified methodological view of probabilistic forecasting.}
    We connect classical statistical approaches with recent machine learning developments through a taxonomy based on the mechanism used to represent predictive uncertainty. The survey spans model-agnostic calibration, Bayesian modeling, parametric predictive distributions, distributional regression, and modern generative approaches, while comparing their assumptions, design choices, computational properties, and limitations. We further examine probabilistic time series foundation models and discuss the methodological challenges involved in extending them to spatiotemporal forecasting. Finally, we provide a discussion on cross-paradigm synthesis of uncertainty representation in the aforementioned model families.

    \item \textbf{From methodological taxonomy to practical model selection.}
    We complement the review with benchmark datasets, open-source model implementations, probabilistic evaluation measures, and statistical comparison procedures. More importantly, we synthesize the methodological differences into two complementary guidance frameworks: a data-driven framework in Fig.~\ref{fig:model_guidance_flowchart}, which links modeling choices to data characteristics, uncertainty requirements, interpretability, and computational constraints; and a domain-specific framework that connects probabilistic forecasting paradigms to recurring requirements across application areas.

    \item \textbf{A cross-paradigm empirical study.}
    We evaluate representative approaches from the taxonomy under a common protocol on univariate, multivariate, and spatiotemporal forecasting problems. The goal is not to identify a universally superior model, but to examine how methodological choices translate into empirical behavior. We therefore analyze point accuracy together with probabilistic performance, calibration, sharpness, computational cost, and statistical robustness. 

    \item \textbf{Future research directions on probabilistic forecasting.}
    Drawing on the reviewed literature and empirical evidence, we identify several problems that remain insufficiently addressed: uncertainty in evolving dependency structures, incorporation of physical knowledge into predictive distributions, forecasting of rare extremes, zero-inflated spatiotemporal counts, directional data, and continuous time series, and the lack of unified software implementations for probabilistic forecasting. We additionally discuss emerging opportunities in probabilistic foundation models, adaptive forecasting systems, and agentic probabilistic forecasting.

\end{itemize}

\begin{table}[!htbp]
\caption{Representative model-agnostic and model-intrinsic probabilistic forecasting methods organized by taxonomy. A dash (-) indicates methods without an explicit name in their original publication.}

\begin{adjustbox}{width=\textwidth, max height=\textheight}
   \centering
    \begin{tabular}{ccccccc} \toprule
    Method Name & Uncertainty Quantification Method & Data Type & Application Domain & Venue & Year & Reference\\ 
    \multicolumn{7}{l}{\cellcolor{AgnosticDark}\textbf{\textcolor{white}{%
  Model-Agnostic}}}\\
  \ragnostic EWNet  & Pre-Control Limits & Univariate & Epidemiology & Neural Networks & 2023 & \cite{panja2023epicasting}\\ 

  \ragnostic  - & Resampling-based Ensemble & Univariate & Energy & Solar Energy & 2016 & \cite{grantham2016nonparametric}\\ 

  \ragnostic S-ESN & Randomization-based Ensemble & Spatiotemporal & Energy & JRSS C & 2022 & \cite{huang2022esn} \\ 

  \ragnostic D-EESN  & Randomization-based Ensemble & Spatiotemporal  & Meteorology & Environmetrics & 2019 & \cite{McDermott2019deen}\\ 

  \ragnostic TimeSeriesXAI & Deep Ensemble & Univariate & Healthcare & IEEE JBHI & 2021 & \cite{wickstrom2021uncertainty}\\

  \ragnostic DESQRUQ  & Deep Ensemble & Spatiotemporal & Transportation  & IEEE T-ITS & 2024 & \cite{mallick2024desqruq} \\ 

  \ragnostic -  & Bayesian Model Averaging & Spatiotemporal  & Meteorology & Urban Science & 2025 & \cite{qin2025urban}\\

  \ragnostic -  & Bayesian Model Averaging & Spatiotemporal  & Meteorology & JASA & 2010 & \cite{Sloughter2010probabilistic}\\ 


  \ragnostic STACI & Conformal Prediction & Spatiotemporal & Environmental & NeurIPS & 2026 & \cite{feng2026staci}\\ 
  
  \ragnostic CSP & Conformal Prediction & Univariate & General & ArXiv & 2026 & \cite{valery2026csp}\\ 

  \ragnostic E-STGCN & Conformal Prediction & Spatiotemporal & Environmental & JRSS A & 2026 & \cite{panja2026estgcn}\\ 


  \ragnostic MTSA-QRN & Conformalized Quantile Regression & Spatiotemporal & Energy & Sustainability & 2026 & \cite{wang2026cqr}\\ 

  \ragnostic EnCQR  & Conformalized Quantile Regression & Multivariate & Energy & IEEE TNNLS & 2024 & \cite{jensen2024ensqr}\\ 

  \ragnostic CTCQRN  & Conformalized Quantile Regression & Multivariate  & Energy & Energy & 2022 & \cite{hu2022ctc}\\

    \multicolumn{7}{l}{\cellcolor{BayesianDark}\textbf{\textcolor{white}{%
  Bayesian Modeling}}}\\

    \rbayesian BSTI  & Bayesian Hierarchical Model & Spatiotemporal  & Transportation & IEEE Access & 2021 & \cite{cui2021bus}\\ 

    \rbayesian -  & Bayesian Hierarchical Model & Spatiotemporal  & Energy & AOAS & 2020 & \cite{lenzi2020wind}\\ 



    \rbayesian AGCGRU+flow & State Space Model 
    & Spatiotemporal & Traffic & ICML & 2021 & \cite{pal2021rnnpf}\\ 

    \rbayesian MBSTS & State Space Model & Multivariate & Econometrics & JMLR & 2018 & \cite{qiu2018mbsts}\\


    \rbayesian LKGP & Gaussian Process & Spatiotemporal & General & ICML & 2025 & \cite{lin2025scalable}\\ 



    \rbayesian Space-Time.DeepKriging & Gaussian Process 
    & Spatiotemporal & Environmental & Spatial Statistics & 2023 & \cite{nag2023DeepKriging} \\ 

    \rbayesian DGGP & Gaussian Process & Spatiotemporal & Traffic  & IEEE T-ITS & 2022 & \cite{jiang2022dggp}\\ 
    
    \rbayesian ST-SVGP & Gaussian Process & Spatiotemporal & General & NeurIPS & 2021 & \cite{hamelijnck2021stvgp}\\ 

    \rbayesian GpGp & Gaussian Process & Spatiotemporal & General & Technometrics & 2018 & \cite{guinness2018gpgp}\\ 

    \rbayesian -  & Dirichlet Process & Univariate  & General & IEEE ICDMW & 2023 & \cite{ren2023infinite}\\

    \rbayesian Sticky HDP-HMM  & Dirichlet Process & Multivariate & Speech Processing & AOAS & 2011 & \cite{fox2011sticky}\\

    \rbayesian ARDL-BART  & Bayesian Additive Regression Tree & Multivariate  & Macroeconomics & AM (SCIRP) & 2024 & \cite{mahdavi2024measuring}\\

    \rbayesian BART  & Bayesian Additive Regression Tree & Multivariate  & Macroeconomics & JOF & 2019 & \cite{pruser2019forecasting}\\

    \rbayesian BNN  & Bayesian Neural Network & Multivariate & Macroeconomics & J. Econom. & 2025 & \cite{Hauzenberger2025bnn}\\

    \rbayesian BARNN  & Bayesian Neural Network & Univariate & Synthetic & ICML & 2025 & \cite{coscia2025barnn}\\ 

    \rbayesian BAST-RNN  & Bayesian Neural Network & Spatiotemporal  & General & Entropy & 2019 & \cite{mcdermott2019bayesianrnn}\\ 

    \rbayesian -  & Approximate Bayesian Inference (MCMC) & Spatiotemporal & Environmental  & Environmetrics & 2023 & \cite{johnson2023bayesian}\\ 

    \rbayesian BSTIM  & Approximate Bayesian Inference (MCMC) & Spatiotemporal & Epidemiology  & PLOS ONE & 2019 & \cite{stojanovic2019bayesian}\\ 

    \rbayesian UA-STTFF  & Approximate Bayesian Inference (MC Dropout) & Spatiotemporal & Transportation  & ICKECS & 2026 & \cite{hemavathi2026traffic}\\ 

    \rbayesian MVE+D  & Approximate Bayesian Inference (MC Dropout) & Univariate & Energy & CIARP & 2019 & \cite{serpell2019mc}\\ 



    \rbayesian BayesNF  & Approximate Bayesian Inference (VI) & Spatiotemporal & General & Nat. Commun. & 2024 & \cite{saad2024scalable}\\ 

    \rbayesian STNN-VB  & Approximate Bayesian Inference (VI) & Spatiotemporal  & Energy & Applied Energy & 2020 & \cite{liu2020bayesvariational}\\ 

    \rbayesian -  & Approximate Bayesian Inference (INLA) & Spatiotemporal  & Criminology & JQC & 2021 & \cite{mahfoud2021burglary}\\

    \rbayesian Spliced Gamma-GP  & Approximate Bayesian Inference (INLA) & Spatiotemporal  & Energy & JABES & 2019 & \cite{castro2019spliced}\\ 

    \multicolumn{7}{l}{\cellcolor{ParametricDark}\textbf{\textcolor{white}{%
  Parametric Predictive Distributions}}}\\

  \rparametric - & Single Parametric Distribution & Multivariate & General & NeurIPS  & 2024 & \cite{zheng2024correlated} \\ 


  \rparametric DeepAR & Single Parametric Distribution & Multivariate & General & IJF & 2020 & \cite{salinas2020deepar} \\

  \rparametric EMOS & Single Parametric Distribution & Spatiotemporal & Environmental & JRSS A & 2010 & \cite{thorarinsdottir2010wind} \\ 

  \rparametric RST & Single Parametric Distribution & Spatiotemporal & Energy & JASA & 2006 & \cite{gneiting2006rst} \\ 
  

  \rparametric HSMPF & Mixture Density Networks & Spatiotemporal & Energy  & IEEE TSTE & 2026 & \cite{he2026hybrid} \\ 


  \rparametric WA-IMDN  & Mixture Density Networks & Univariate & Energy  & IEEE TSTE & 2022 & \cite{yang2022waimdn} \\ 


  \rparametric IDMDN  & Mixture Density Networks & Spatiotemporal & Energy  & IEEE TPS & 2020 & \cite{zhang2020idmdn} \\ 


  \rparametric - & Mixture Density Networks & Spatiotemporal & Transportation & IEEE RA-L & 2019 & \cite{zhi2019directional} \\

  \multicolumn{7}{l}{\cellcolor{DistRegDark}\textbf{\textcolor{white}{%
  Distributional Regression}}}\\ 

  \rdistreg EQRN & Quantile Regression & Spatiotemporal & Hydrology & AOAS & 2024 & \cite{pasche2024eqrn} \\



  \rdistreg SQR & Quantile Regression & Spatiotemporal & Energy & IEEE IAS & 2019 & \cite{yu2019quantile} \\ 


  \rdistreg LSTM-Engression & Engression & Multivariate & Hydrology  & GRL & 2026 & \cite{kraft2026engression} \\ 


  \rdistreg Enformer & Engression & Multivariate & General & ArXiv & 2026 & \cite{pathak2026entransformer} \\ 

  \rdistreg GCEN/STEN & Engression & Spatiotemporal & Epidemiology   & TMLR & 2026 & \cite{pathak2026engression} \\ 




  \rdistreg Tube Loss & Loss-Driven Interval Prediction & Multivariate & General & TMLR & 2026 & \cite{anand2026tube} \\

  \rdistreg DeepPIPE & Loss-Driven Interval Prediction & Multivariate & General & Neurocomputing & 2020 & \cite{wang2020DeepPIPE} \\

  \multicolumn{7}{l}{\cellcolor{GenerativeDark}\textbf{\textcolor{white}{%
  Generative Models}}}\\


  \rgenerative STMPNet & Copula-Based Model & Spatiotemporal & Traffic & IEEE TKDE  & 2025 & \cite{an2025stmpnet} \\

  \rgenerative TACTiS-2 & Copula-Based Model & Multivariate & General & ICLR  & 2024 & \cite{ashok2024tactis} \\

  \rgenerative GP-Copula & Copula-Based Model & Multivariate & General & NeurIPS  & 2019 & \cite{salinas2019gpcopula} \\

  \rgenerative DE-HMM & Hidden Markov Model & Spatiotemporal & Autonomous Systems & Transportmetrica A & 2026 & \cite{zhou2026dehmm} \\ 

   \rgenerative IOHMM & Hidden Markov Model & Spatiotemporal & Transportation & J. Adv. Transp. & 2026 & \cite{liu2026iohmm} \\

   \rgenerative MOHMM & Hidden Markov Model & Spatiotemporal & Energy & Energies & 2025 & \cite{zhang2025mohmm} \\ 

   \rgenerative InfHMM & Hidden Markov Model & Univariate & Energy & Solar Energy & 2022 & \cite{frimane2022infhmm} \\ 

   \rgenerative C2FAR & Hierarchical Discretization & Univariate & General & NeurIPS & 2022 & \cite{bergsma2022c2far} \\ 


  \rgenerative HDT & Hierarchical Discretization & Multivariate & General & AAAI & 2025 & \cite{shibo2025hdt} \\ 

  \rgenerative K$^2$VAE & Variational Autoencoder & Multivariate & General & ICML & 2025 & \cite{wu2025kvae} \\ 

  \rgenerative Graph-EFM & Variational Autoencoder & Spatiotemporal & Meteorology & NeurIPS & 2024 & \cite{oskarsson2024probabilistic} \\


  \rgenerative - & Generative Adversarial Networks & Spatiotemporal & Environmental & JMLR & 2024 & \cite{pacchiardi2024gan} \\

  \rgenerative STI-GAN & Generative Adversarial Networks & Spatiotemporal & Autonomous Systems & IEEE Access & 2021 & \cite{huang2021stigan} \\ 


  \rgenerative ProFITi  & Normalizing Flows & Multivariate & General & AAAI & 2025 & \cite{yalavarthi2025ProFITi} \\ 

  \rgenerative STGNF  & Normalizing Flows & Spatiotemporal & Transportation  & CIKM & 2024 & \cite{an2024graphnf} \\

  \rgenerative DyLand  & Normalizing Flows & Spatiotemporal & Geoscience & IEEE Trans. Cybern. & 2024 & \cite{xu2024dyland} \\

  \rgenerative MotionFlow  & Normalizing Flows & Spatiotemporal & Computer Vision & IEEE TPAMI & 2023 & \cite{Zand2023stsp} \\ 

  \rgenerative LSTM/Transformer-MAF & Normalizing Flows & Multivariate & General & ICLR & 2021 & \cite{rasul2021multivariate} \\ 

  \rgenerative NsDiff & Denoising Diffusion Probabilistic Model & Multivariate & General & ICML & 2025 & \cite{ye2025nsdiff} \\ 
  
  \rgenerative SpecSTG & Denoising Diffusion Probabilistic Model & Spatiotemporal & Transportation & IJCNN & 2025 & \cite{lin2025SpecSTG} \\ 

  
  \rgenerative DST-DDPM & Denoising Diffusion Probabilistic Model & Spatiotemporal & Environmental & Environ. Res. & 2024 & \cite{Chen2024diffusion} \\ 

  \rgenerative MG-TSD & Denoising Diffusion Probabilistic Model & Multivariate & General & ICLR & 2024 & \cite{fan2024mgtsd} \\ 

  \rgenerative DiffSTG & Denoising Diffusion Probabilistic Model & Spatiotemporal & General & SIGSPATIAL & 2023 & \cite{wen2023diffstg} \\

  \rgenerative TimeDiff  & Denoising Diffusion Probabilistic Model & Multivariate & General & ICML & 2023 & \cite{shen2023nonautoregressive} \\ 
  
  \rgenerative TimeGrad & Denoising Diffusion Probabilistic Model & Multivariate & General & ICML & 2021 & \cite{rasul2021timegrad} \\ 

   \rgenerative GenCast & Score-based Diffusion Model & Spatiotemporal & Meteorology & Nature  & 2025 & \cite{price2024gencast} \\ 



  \rgenerative - & Score-based Diffusion Model & Multivariate & General & ICML & 2023 & \cite{bilos2023diffusion} \\

  \rgenerative DYffusion & Score-based Diffusion Model & Spatiotemporal & General & NeurIPS & 2023 & \cite{cachay2023dyffusion} \\ 


  \rgenerative SHADECast & Latent Diffusion Model & Spatiotemporal & Energy  & Applied Energy & 2025 & \cite{Carpentieri2025SHADECast} \\ 

  \rgenerative LDT & Latent Diffusion Model & Multivariate & General & AAAI & 2024 & \cite{feng2024ldt} \\ 

  \rgenerative LADM & Latent Diffusion Model & Spatiotemporal & Transportation & IEEE TIM & 2024 & \cite{lv2024ladm} \\ 

  \rgenerative PPM & Parametric Prior Mapping & Multivariate & General & ICML & 2026 & \cite{li2026ppm} \\ 

  \rgenerative TimePrism & Probabilistic Scenarios & Multivariate & General & ICLR & 2026 & \cite{dai2026timeprism} \\
  
  \rgenerative TimePre & Multiple Choice Learning & Multivariate & General & TMLR & 2026 & \cite{jiang2026timepre} \\

  \rgenerative TimeMCL & Multiple Choice Learning & Multivariate & General & ICML & 2025 & \cite{cortes2025timemcl} \\
  \bottomrule
    \end{tabular}
    \label{tab:UQ_Methods_Summary}
    \end{adjustbox}
\end{table}

The remainder of this survey is organized as follows. Section~\ref{sec_background} introduces the notation and statistical foundations, while Section~\ref{sec_literature_review} reviews related surveys and positions this work within the broader landscape. Sections~\ref{sec_model_agnostic} and~\ref{sec_model_intrinsic} cover model-agnostic and model-intrinsic uncertainty quantification methods, respectively. 
Section~\ref{sec_foundation_models} discusses probabilistic time series foundation models and their potential extension to spatiotemporal settings. Section~\ref{sec_cross_paradigm_synthesis} offers a systematic cross-paradigm comparison of fundamental assumptions, theoretical guarantees, predictive uncertainty representations, computational overhead, and practical limitations. Section~\ref{sec_evaluation_model_selection} presents datasets, software resources, probabilistic evaluation measures, statistical comparison procedures, and guidance for method selection. Section~\ref{sec_emp_demo} presents the empirical study of representative approaches across temporal and spatiotemporal forecasting tasks. Section~\ref{sec_challenges} discusses open challenges and future research directions, and Section~\ref{sec_conclusion} concludes the survey.

\subsection{Background and Notation}\label{sec_background}
A time series consists of observations recorded at uniformly spaced intervals \citep{hyndman2021forecasting}. Let $\mathcal{S} = \{1, 2, \dots, N\}$ define a set of discrete spatial locations and $\mathcal{T} = \{1, 2, \dots, T\}$ denote a sequence of equally spaced time points. A univariate time series at a specific location $s \in \mathcal{S}$ is defined as the scalar sequence $\{Y(s, t)\}_{t \in \mathcal{T}}$, where $Y(s, t) \in \mathbb{R}$ represents the target variable at time $t$. Extending this to a multivariate context, we define a sequence of column vectors $\{y_t\}_{t=1}^{T}$, where $y_t = [Y(1, t), \dots, Y(N, t)]^\top \in \mathbb{R}^N$, stacking the observations across all $N$ locations at each time step $t$. The complete collection of these variables can be represented as a matrix $\mathbf{Y}_{1:T} = [y_1^\top, \ldots, y_T^\top]^\top \in \mathbb{R}^{T \times N}$. This matrix constitutes a spatiotemporal dataset capturing both temporal dynamics and spatial interdependencies, if additional spatial information such as connectivity among the $N$ locations is exploited. Many real-world spatiotemporal systems are described by discrete relational structures that encode interactions through connectivity. In systems such as transportation networks and power grids, observations are defined over nodes of a graph $\mathcal{G} = (\mathcal{V},\mathcal{E})$, where $\mathcal{V}$ is the set of $N$ spatial locations (nodes) and $\mathcal{E}$ represents the connections (edges) between them. 
Graph neural networks (GNNs), including graph convolutional and graph attention-based models, propagate information along edges to capture these relational dependencies within a spatiotemporal framework \citep{kipf2017gcn, velickovic2018graphattention}. A more general form of a spatiotemporal dataset measures $D$ features for $N$ spatial locations across $T$ time steps, rendering a three-dimensional tensor $\mathcal{Y} \in \mathbb{R}^{T \times N \times D}$. A sequence of exogenous covariates $\{x_t\}_{t \in \mathcal{T}}$ may be incorporated alongside the historical data to provide information on external factors influencing the target variable, thereby improving predictive accuracy.

The forecasting task is to predict future states $\{y_{T+1}, \dots, y_{T+H}\}$ for all $N$ locations over a horizon of $H$ steps, given the observed history $\mathbf{Y}_{1:T}$. A point forecasting model maps this history to produce a single deterministic estimate $\widehat{y}_{T+h} \in \mathbb{R}^N$ for each future step $h \in \{1, \dots, H\}$. In contrast, probabilistic forecasting seeks to characterize the uncertainty inherent in future states \citep{gneiting2014probabilistic, hyndman2021forecasting}. Rather than producing a single fixed vector, a probabilistic model estimates the conditional predictive probability distribution $p(y_{T+h} \mid \mathbf{Y}_{1:T})$ for each future step $h \in \{1, \dots, H\}$, or jointly over the entire horizon. Point forecasts, prediction intervals, and quantiles can then be obtained as summaries of the predictive distribution \citep{gneiting2014probabilistic}. Throughout the remainder of this work, we adhere to the notations established in this section. Unless explicitly stated otherwise, all standard mathematical operations involving vector-valued quantities, such as the addition or squaring of $\widehat{y}_{T+h}$, are to be interpreted as element-wise operations across the $N$ spatial dimensions.

\section{Literature Review}\label{sec_literature_review}

The growing forecasting literature has produced numerous surveys focusing primarily on point estimation and predictive accuracy. Early reviews established neural networks as viable alternatives to classical forecasting models and laid the foundation for subsequent developments in deep learning-based forecasting \citep{zhang1998ann}. More recent surveys have provided comprehensive overviews of modern architectures, including recurrent networks, convolutional models, Transformers, and hybrid forecasting frameworks \citep{lim2021forecasting, benidis2022deeplearning}. While these reviews have significantly advanced the understanding of forecasting methodologies, probabilistic forecasting is often treated as supplementary. The literature on spatiotemporal forecasting follows a similar pattern. Existing surveys provide extensive coverage of architectures designed to capture spatial and temporal dependencies, including convolutional recurrent networks, attention mechanisms, and GNNs \citep{wang2022deeplearning,yuan2021survey, jin2024gnn}. However, these surveys primarily address modeling architectures and deterministic prediction, with probabilistic forecasting and uncertainty quantification forming only part of their broader scope. Recent surveys have also examined foundation models for time series analysis \citep{liang2024foundation} and large models for both temporal and spatiotemporal data \citep{jin2026large}. These surveys cover model architectures, pre-training and adaptation strategies, data modalities, and downstream tasks, but pay limited attention to predictive uncertainty. On the other hand, surveys addressing uncertainty quantification in deep learning often treat it as a broader machine learning problem. For instance, \cite{abdar2021review} and \cite{gawlikowski2023survey} cover techniques such as Bayesian neural networks and ensembles, but provide limited attention to forecasting-specific dependence, multi-horizon uncertainty, and probabilistic evaluation. Similarly, while \cite{gneiting2014probabilistic} established rigorous principles for probabilistic forecasting and \citet{cressie2011statistics} provided a comprehensive statistical treatment of temporal and spatiotemporal processes, these foundational works predate the recent expansion of conformal methods in forecasting, modern deep probabilistic and generative models, and time series foundation models.


\begin{table}[!t]
\caption{Qualitative comparison of this survey with existing works. (\textcolor{ForestGreen}{\ding{51}}) denotes substantive treatment of the column's topic, (\textcolor{Blue}{\circmark}) denotes partial or secondary treatment, and (\textcolor{red}{\ding{55}})
denotes no identified substantive treatment. ``Probabilistic'' concerns predictive uncertainty, rather than point accuracy alone. ``TS Forecasting'' and ``ST Forecasting'' concern whether the survey spans forecasting future outcomes in temporal and spatially structured temporal settings, respectively. ``Architecture Agnostic'' denotes comparison of multiple modeling or uncertainty mechanisms rather than one forecasting architecture; ``General Scope'' denotes coverage across application domains rather than one designated forecasting domain.}

\begin{adjustbox}{width=\textwidth}
   \centering
    \begin{tabular}{cccccc} \toprule
    \textbf{Survey} & \textbf{Probabilistic} & \textbf{TS Forecasting} & \textbf{ST Forecasting} & \textbf{Architecture Agnostic} & \textbf{General Scope}  \\\midrule

  \cite{gneiting2014probabilistic}   & \textcolor{ForestGreen}{\ding{51}} & \textcolor{ForestGreen}{\ding{51}} & \textcolor{Blue}{\circmark}  & \textcolor{ForestGreen}{\ding{51}}  & \textcolor{Blue}{\circmark} \\

  \cite{hong2016probsurvey} & \textcolor{ForestGreen}{\ding{51}} & \textcolor{ForestGreen}{\ding{51}} & \textcolor{red}{\ding{55}}   & \textcolor{ForestGreen}{\ding{51}}  & \textcolor{red}{\ding{55}}  \\

  \cite{nowotarski2018electricity} & \textcolor{ForestGreen}{\ding{51}} & \textcolor{ForestGreen}{\ding{51}} & \textcolor{red}{\ding{55}}   & \textcolor{ForestGreen}{\ding{51}}  & \textcolor{red}{\ding{55}}  \\

  \cite{li2020solar} & \textcolor{ForestGreen}{\ding{51}} & \textcolor{ForestGreen}{\ding{51}} & \textcolor{Blue}{\circmark}   & \textcolor{ForestGreen}{\ding{51}}  & \textcolor{red}{\ding{55}}  \\

  \cite{bazionis2021electricity} & \textcolor{Blue}{\circmark} & \textcolor{ForestGreen}{\ding{51}} & \textcolor{Blue}{\circmark}   & \textcolor{ForestGreen}{\ding{51}}  & \textcolor{red}{\ding{55}}  \\

  \cite{abdar2021review} & \textcolor{ForestGreen}{\ding{51}}  & \textcolor{Blue}{\circmark} & \textcolor{Blue}{\circmark} & \textcolor{ForestGreen}{\ding{51}}  & \textcolor{ForestGreen}{\ding{51}} \\

   \cite{lim2021forecasting} & \textcolor{red}{\ding{55}}   & \textcolor{ForestGreen}{\ding{51}} & \textcolor{red}{\ding{55}} & \textcolor{Blue}{\circmark}  & \textcolor{ForestGreen}{\ding{51}} \\

   \cite{yuan2021survey} & \textcolor{red}{\ding{55}}   & \textcolor{ForestGreen}{\ding{51}} & \textcolor{ForestGreen}{\ding{51}} & \textcolor{ForestGreen}{\ding{51}}  & \textcolor{red}{\ding{55}} \\

   \cite{kaur2022energy} & \textcolor{Blue}{\circmark}   & \textcolor{ForestGreen}{\ding{51}} & \textcolor{red}{\ding{55}} & \textcolor{ForestGreen}{\ding{51}}  & \textcolor{red}{\ding{55}} \\

   \cite{benidis2022deeplearning} & \textcolor{Blue}{\circmark}   & \textcolor{ForestGreen}{\ding{51}} & \textcolor{red}{\ding{55}} & \textcolor{Blue}{\circmark}  & \textcolor{ForestGreen}{\ding{51}} \\

   \cite{gao2022gan} & \textcolor{Blue}{\circmark}  & \textcolor{Blue}{\circmark} & \textcolor{Blue}{\circmark} & \textcolor{red}{\ding{55}}  & \textcolor{ForestGreen}{\ding{51}} \\ 

   \cite{Papacharalampous2022hydrological} & \textcolor{ForestGreen}{\ding{51}}  & \textcolor{ForestGreen}{\ding{51}} & \textcolor{red}{\ding{55}} & \textcolor{ForestGreen}{\ding{51}}  & \textcolor{red}{\ding{55}} \\

   \cite{wang2022deeplearning} & \textcolor{red}{\ding{55}}  & \textcolor{Blue}{\circmark} & \textcolor{Blue}{\circmark} & \textcolor{ForestGreen}{\ding{51}}  & \textcolor{ForestGreen}{\ding{51}} \\

   \cite{gawlikowski2023survey} & \textcolor{ForestGreen}{\ding{51}} & \textcolor{Blue}{\circmark} & \textcolor{red}{\ding{55}} & \textcolor{ForestGreen}{\ding{51}}  & \textcolor{ForestGreen}{\ding{51}} \\ 

   \cite{xie2023probabilistic} &  \textcolor{Blue}{\circmark}   & \textcolor{ForestGreen}{\ding{51}} & \textcolor{Blue}{\circmark} & \textcolor{ForestGreen}{\ding{51}}  & \textcolor{red}{\ding{55}} \\

   \cite{yu2024deep} & \textcolor{Blue}{\circmark}   & \textcolor{ForestGreen}{\ding{51}} & \textcolor{ForestGreen}{\ding{51}} & \textcolor{Blue}{\circmark}  & \textcolor{red}{\ding{55}} \\ 

   \cite{Chakraborty2024gan} & \textcolor{Blue}{\circmark}   & \textcolor{Blue}{\circmark} & \textcolor{red}{\ding{55}} & \textcolor{red}{\ding{55}}  & \textcolor{ForestGreen}{\ding{51}} \\ 

   \cite{liu2024deep} & \textcolor{Blue}{\circmark}  & \textcolor{ForestGreen}{\ding{51}} & \textcolor{red}{\ding{55}} & \textcolor{Blue}{\circmark}  & \textcolor{ForestGreen}{\ding{51}} \\ 

    \cite{blasco2024survey} & \textcolor{ForestGreen}{\ding{51}}  & \textcolor{ForestGreen}{\ding{51}} & \textcolor{red}{\ding{55}} & \textcolor{Blue}{\circmark}  & \textcolor{red}{\ding{55}} \\ 

   \cite{jin2024gnn} & \textcolor{Blue}{\circmark}   & \textcolor{ForestGreen}{\ding{51}} & \textcolor{ForestGreen}{\ding{51}} & \textcolor{red}{\ding{55}}  & \textcolor{ForestGreen}{\ding{51}} \\ 

   \cite{de2025beyond} & \textcolor{ForestGreen}{\ding{51}}   & \textcolor{ForestGreen}{\ding{51}} & \textcolor{red}{\ding{55}} & \textcolor{ForestGreen}{\ding{51}}  & \textcolor{red}{\ding{55}} \\ 

   \cite{rickard2026financial} & \textcolor{ForestGreen}{\ding{51}}  & \textcolor{ForestGreen}{\ding{51}} & \textcolor{red}{\ding{55}} & \textcolor{ForestGreen}{\ding{51}}  & \textcolor{red}{\ding{55}} \\ 

   \cite{yang2026diffusion} & \textcolor{ForestGreen}{\ding{51}}  & \textcolor{Blue}{\circmark} & \textcolor{Blue}{\circmark} & \textcolor{red}{\ding{55}}  & \textcolor{ForestGreen}{\ding{51}} \\ 

    \midrule
    \textbf{Our survey} & \textcolor{ForestGreen}{\ding{51}} & \textcolor{ForestGreen}{\ding{51}} & \textcolor{ForestGreen}{\ding{51}} & \textcolor{ForestGreen}{\ding{51}} & \textcolor{ForestGreen}{\ding{51}} \\\bottomrule
  
    \end{tabular}
    \label{tab:survey_comparison}
    \end{adjustbox}
\end{table}

The growing importance of probabilistic forecasting has also motivated a number of domain-specific reviews. In energy systems, \cite{hong2016probsurvey} provides a tutorial on probabilistic load forecasting, while \cite{li2020solar} reviews the integration of probabilistic solar forecasting into power system operations, and \cite{kaur2022energy} surveys smart grid forecasting with an emphasis on probabilistic methods. In wind energy, \cite{bazionis2021electricity} and \cite{xie2023probabilistic}  survey deterministic and probabilistic methods, offering practical guidance for practitioners. In finance, \cite{nowotarski2018electricity} reviews probabilistic electricity price forecasting, \cite{rickard2026financial} examines probabilistic artificial intelligence methods for financial time series, and \cite{blasco2024survey} surveys uncertainty quantification in deep learning for financial markets. In hydrology, \cite{Papacharalampous2022hydrological} reviews machine learning approaches for probabilistic post-processing and forecasting, while \cite{de2025beyond} provides a comprehensive review of uncertainty quantification methods across short-to-seasonal horizons. These works offer valuable depth within their respective domains, but their scope is shaped by domain-specific datasets, assumptions, and evaluation protocols. They do not establish a methodological perspective that carries across different temporal and spatiotemporal problems. More recently, dedicated surveys have examined generative adversarial networks (GANs) broadly \citep{Chakraborty2024gan}, GANs for spatiotemporal data \citep{gao2022gan}, and diffusion models for temporal and spatiotemporal applications \citep{yang2026diffusion}. These reviews have advanced understanding of rapidly evolving model families and their applications to forecasting, imputation, and anomaly detection. However, by focusing on a single methodological paradigm, they offer limited guidance on how generative approaches compare to alternative probabilistic forecasting frameworks. Therefore, as summarized in Table~\ref{tab:survey_comparison}, there is currently no comprehensive survey that jointly covers probabilistic forecasting across both temporal and spatiotemporal settings while spanning the full range of approaches from classical statistical methods through to modern generative models. Building on the foundational works \citep{cressie2011statistics, gneiting2014probabilistic}, this survey addresses the literature gap by presenting a unified review of probabilistic forecasting for time series and spatiotemporal data, connecting classical statistical approaches with contemporary calibration, deep generative, and foundation-model methods through a unified taxonomy of how uncertainty is introduced and handled.

\section{Model-Agnostic Uncertainty Quantification}\label{sec_model_agnostic}
\begin{figure}[!t]
    \centering
    \includegraphics[width=\linewidth]{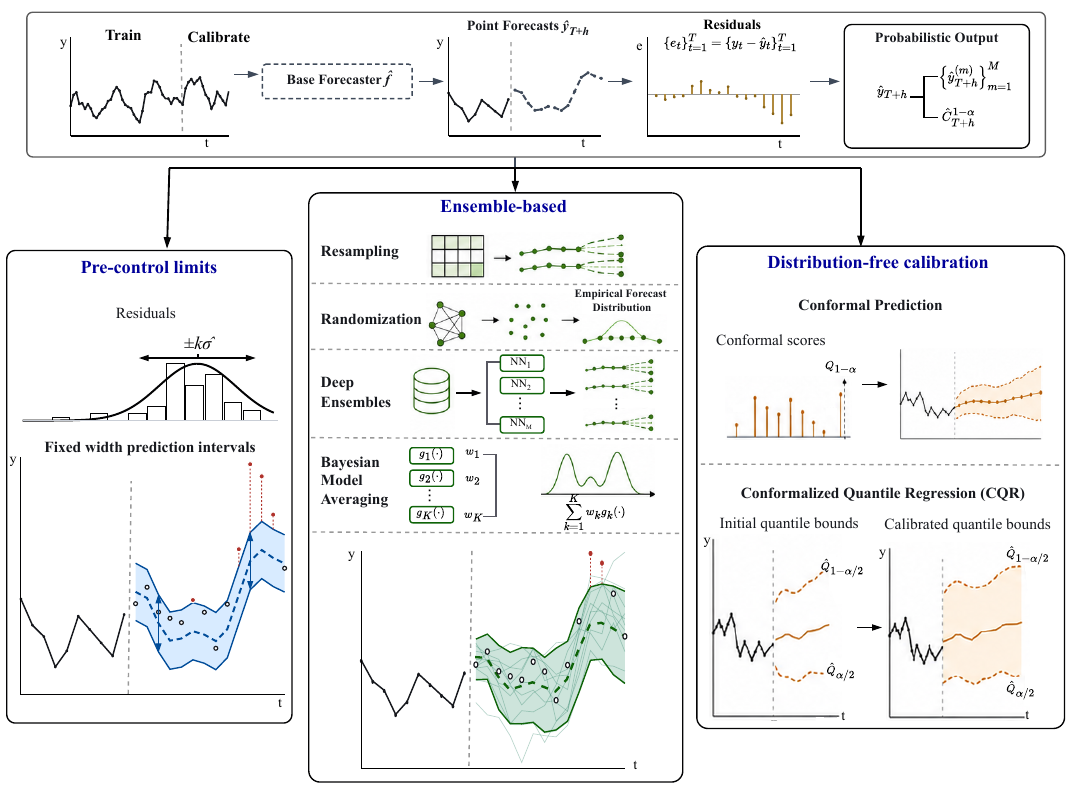}
    \caption{Overview of model-agnostic uncertainty quantification strategies applied to a base forecasting model, comprising residual-based pre-control limits, ensemble-based predictive distributions, and distribution-free calibration.}
    \label{fig:Model_Agnostic_Overview}
\end{figure}

Model-agnostic uncertainty quantification methods estimate predictive uncertainty through procedures that can be applied across different forecasting architectures and model classes \citep{efron1993bootstrap,vovk2005conformal,angelopoulos2023conformal}. These methods operate largely independently of the underlying forecaster and can therefore be used with a wide range of models, from classical statistical methods to modern deep neural networks. Depending on the method, they may be applied post hoc or require repeated model fitting or an uncertainty-aware base estimator.
In contrast to the model-intrinsic approaches reviewed in Section~\ref{sec_model_intrinsic}, these methods do not require uncertainty to be encoded in a single forecasting architecture. Depending on the method, predictive uncertainty is obtained through residual transformation, resampling, refitting, model randomization, forecast combination, or post-hoc calibration. 
The approaches reviewed in this section are grouped into three broad families, illustrated in Fig.~\ref{fig:Model_Agnostic_Overview}: pre-control limits (Section~\ref{sec_precontrol_limits}), ensemble-based approaches (Section~\ref{sec_ensemble}), and distribution-free calibration methods (Section~\ref{sec_distribution_free}). Their main characteristics are summarized in Table~\ref{tab:model_agnostic_comparison}.

\begin{table}[!htbp]
\caption{Comparison of model-agnostic methods by coverage guarantee, parametric assumptions, adaptive width, and computational cost. Check marks (\ding{51}) denote yes, crosses (\ding{55}) denote no.}

\begin{adjustbox}{width=\textwidth}
   \centering
    \begin{tabular}{lcccccl} \toprule
    \textbf{Method} & \textbf{Coverage Guarantee} & \textbf{Parametric Assumptions} & \textbf{Adaptive Width} & \textbf{Computational Cost} \\\midrule

  Pre-control Limits & \ding{55} & \ding{51} & \ding{55} & Low \\

   Resampling-based Ensembles & Asymptotic\textsuperscript{$\dagger$} & \ding{55} & \ding{55} & Low \\

   Randomization-based Ensembles & \ding{55} & \ding{55} & \ding{51} & Moderate \\

   Deep Ensembles & \xmark & \cmark & \cmark & High\textsuperscript{$\ddagger$} \\

   Bayesian Model Averaging & \ding{55} & \ding{51} & \ding{51} & Moderate \\

    Conformal Prediction & Finite-sample marginal\textsuperscript{$\S$} & \ding{55} & \ding{51} & Low-Moderate\\



   Conformalized Quantile Regression & Finite-sample marginal\textsuperscript{$\S$} & \ding{55} & \ding{51} & Moderate \\

    \bottomrule
    \end{tabular}
    \label{tab:model_agnostic_comparison}
    \end{adjustbox}
\footnotesize{\textsuperscript{$\dagger$}Coverage is unguaranteed for the i.i.d. bootstrap under serial dependence; block-bootstrap variants provide asymptotic consistency under stationarity and mixing conditions \citep{kunsch1989jackknife}.\\
\textsuperscript{$\ddagger$}Scales linearly with the number of ensemble members.\\
\textsuperscript{$\S$}For standard split conformal formulations, the guarantee requires exchangeability of calibration and test scores and the finite-sample quantile correction. It is marginal, not automatically conditional or simultaneous across forecast horizons or spatial locations.}

\end{table}


\subsection{Pre-Control Limits}\label{sec_precontrol_limits}
Pre-control limits are a simple model-agnostic approach to uncertainty quantification, originating in statistical process control for monitoring process variability \citep{rahimifard2020precontrol}. They belong to a broader class of process control methods that quantify variation using control charts and fixed dispersion-based thresholds \citep{castillo2002statistical, montgomery2020introduction}. In forecasting, the same principle can be applied to construct prediction intervals around point forecasts by assuming that errors are i.i.d. and have constant variance. Given a forecasting model that produces point forecasts $\widehat{y}_{T+h}$ and historical in-sample residuals $\mathcal{E} = \{y_t - \widehat{y}_t\}_{t=1}^T$, the prediction interval for a future observation $\widehat y_{T+h}$ is constructed as $\widehat y_{T+h} \pm k \cdot \widehat{\sigma}$, where $\widehat{\sigma}$ is the standard deviation of the historical residuals $\mathcal{E}$ and $k$ is a fixed multiplier. Under a Gaussian assumption, $k = 1.5$ yields an expected coverage of approximately 86\% \citep{panja2023epicasting}. The primary advantage of pre-control limits is their computational simplicity since the intervals are a transformation of the point forecast that requires a single scalar estimate of residual variance. However, the method relies on the strong assumptions of stationarity and homoskedasticity (constant variance) of forecast errors. In temporal data, these assumptions are often violated, leading to severely miscalibrated intervals, especially for multi-step-ahead forecasts as observed in epidemic forecasting \citep{panja2023epicasting}. These restrictive assumptions have limited the use of pre-control limits in modern probabilistic forecasting and motivate more flexible approaches for representing predictive uncertainty. 


\subsection{Ensemble-Based Methods}\label{sec_ensemble}
Ensemble-based methods extend the model-agnostic perspective by representing predictive uncertainty through variation across multiple models or forecast trajectories, rather than through a fixed estimate of residual dispersion. These methods can be broadly divided into two categories: constructing an ensemble of forecasting models and extracting an ensemble of forecasts from a single model. The latter characterizes predictive uncertainty through an ensemble of plausible future trajectories rather than a single point forecast, using the variance among these ensemble members to quantify forecast uncertainty \citep{elliott2006handbook}. This inherent flexibility has led to the widespread adoption of ensemble methods across statistical modeling, machine learning, and environmental forecasting \citep{dietterich2000ml,gneiting2005calibrated}. Existing approaches differ primarily in how ensemble diversity is generated, including resampling-based strategies (Section~\ref{sec_resampling_based}), model randomization (Section~\ref{sec_randomization_based}), deep ensembles (Section~\ref{sec_deep_ensemble}), and forecast combination methods such as Bayesian model averaging (Section~\ref{sec_bma}). Although conceptually simple, applying ensemble methods to temporal and spatiotemporal forecasting requires careful treatment of temporal and spatial dependence structures, which are often violated by standard ensemble construction procedures \citep{cressie2011statistics}. 

\subsubsection{Resampling-Based Ensembles}\label{sec_resampling_based}
Resampling-based methods provide one of the most established mechanisms for generating ensemble diversity. Bootstrap approaches approximate the predictive distribution by repeatedly resampling the observed data or fitted residuals and generating a collection of plausible future trajectories \citep{efron1993bootstrap}.
Given a forecasting model that produces a point forecast $\widehat{y}_{T+h}$ and a set of historical forecast errors $\mathcal{E} = \{e_t\}_{t=1}^T$, where $e_t = y_t - \widehat{y}_t$, residual bootstrap methods generate an ensemble of horizon-specific forecast samples by resampling from these past residuals:
$$\widehat{y}_{T+h}^{(m)} = \widehat{y}_{T+h} + \tilde{e}_{h}^{(m)}, \quad m = 1, \dots, M,$$
where $\tilde{e}_{h}^{(m)}$ is a residual drawn randomly with replacement from the set $\mathcal{E}$. The resulting ensemble $\{\widehat{y}_{T+h}^{(m)}\}_{m=1}^M$ approximates the predictive distribution for the $h$-step-ahead time point, from which prediction intervals can be obtained using empirical quantiles.

However, preserving the temporal dependence structure remains a central challenge. Standard bootstrap procedures rely on independent sampling and may destroy temporal correlations present in the data. Block bootstrap methods address this issue by resampling groups of consecutive observations to preserve temporal structure \citep{kunsch1989jackknife}. 
An additional limitation of the direct application of residual bootstrap methods is the assumption that historical errors are representative of future uncertainty. This assumption is violated when the error distribution changes over time or depends on external factors. To address this limitation, \cite{grantham2016nonparametric} proposed a conditional bootstrap approach for solar radiation forecasting, where residuals are partitioned into bins according to covariates that influence variance, then sampled from the appropriate bin for each forecast. This method produces prediction intervals that better reflect the heteroskedastic nature of the forecasting problem. For spatiotemporal forecasting, however, additional strategies are required to maintain spatial dependence, making uncertainty estimation considerably more challenging in high-dimensional settings \citep{cressie2011statistics}.

\subsubsection{Randomization-Based Ensembles}\label{sec_randomization_based}
Resampling observations or residuals can distort the temporal and spatial dependence structure. Randomization-based ensembles overcome this by 
generating diversity through stochasticity in forecasting model. 
Echo state networks (ESNs) provide an early example of this idea in forecasting \citep{jaeger2001echo}. ESNs belong to the family of reservoir computing models, where the reservoir weights are randomly initialized and kept fixed during training, while only the output layer is trained. 
Despite their effectiveness in modeling nonlinear temporal dynamics, ESNs yield a single deterministic forecast for a given reservoir realization. To obtain uncertainty estimates, \cite{McDermott2017qesn} introduced the ensemble ESN (EESN), where an ensemble of ESN models is generated through independent reservoir realizations. 
For each future step $h \in \{1, \dots, H\}$, this process generates an ensemble of forecasts $\{\widehat{y}_{T+h}^{(m)}\}_{m=1}^{M}$. The resulting empirical predictive distribution for the $h$-step-ahead time point is typically summarized by the ensemble mean and variance:
$$\bar{y}_{T+h} = \frac{1}{M} \sum_{m=1}^{M} \widehat{y}_{T+h}^{(m)},  \qquad \text{and} \qquad \widehat \sigma^2_{T+h} = \frac{1}{M-1} \sum_{m=1}^M \left( \widehat{y}_{T+h}^{(m)} - \bar{y}_{T+h} \right)^2,$$
respectively, where the mean ($\bar{y}_{T+h}$) serves as the point forecast and the variance ($\widehat \sigma^2_{T+h}$) provides a measure of predictive uncertainty. 
Subsequent studies extended this idea to deep reservoir architectures capable of representing spatiotemporal dynamics \citep{McDermott2019deen} and incorporated spatial dimension reduction strategies for large-scale spatiotemporal forecasting \citep{huang2022esn}. 
However, 
generating sufficiently diverse ensembles can become computationally demanding, particularly for complex forecasting models and high-dimensional spatiotemporal applications \citep{cressie2011statistics,huang2022esn}. 

Another conventional randomization-based framework is the random forest \citep{breiman2001random}, which trains an ensemble of decision trees in parallel using bootstrap samples and randomized feature subsets, thereby combining data and model randomization. The efficacy of such ensembles relies on aggregating strong, uncorrelated models. This diversity is classically achieved through bagging (bootstrapping), where individual members are fit to distinct resampled subsets of the training data. Standard random forests approximate the conditional mean of a response variable. To capture the conditional predictive distribution, \cite{meinshausen2006quantile} extended random forests through quantile regression forests (QRF), which estimate conditional quantiles from the forest-induced distribution of responses. QRF therefore lies at the intersection of ensemble-based methods and distributional regression (discussed in Section \ref{sec_distributional}). 
This methodology was recently adapted for sequential temporal data via the time series QRF (tsQRF) \citep{shiraishi2024time}. Drawing conceptual similarities from QRF, \cite{hasson2021probabilistic} introduced the level set forecaster (LSF), a model-agnostic approach capable of transforming any point-prediction algorithm into a probabilistic forecaster. LSF accomplishes this by grouping training instances that yield sufficiently similar point forecasts, subsequently utilizing the empirical distribution of the ground-truth values within these localized bins to formulate the final predictive density.

\subsubsection{Deep Ensembles}\label{sec_deep_ensemble}
Bagging-based ensembles \citep{breiman2001random} can be inefficient for deep neural networks because each base learner is trained on only a subset ($\sim63\%$) of the unique training data. Deep ensembles instead induce diversity through independent parameter initialization and stochastic training while allowing each network to use the full training dataset \citep{lakshminarayanan2017simple}. This combination preserves data efficiency and model capacity while providing a practical means of quantifying predictive uncertainty from variation across independently trained models. Recent work has extended deep ensembles to address uncertainty estimation in increasingly complex forecasting settings.
For clinical time series, \cite{wickstrom2021uncertainty} quantify uncertainty in temporal relevance scores by measuring their variation across independently trained neural networks, providing uncertainty estimates alongside model explanations.
\cite{maulik2023quantifying} use genetic algorithms and Bayesian optimization to construct diverse ensembles of high-performing neural architectures and decompose predictive variability to quantify uncertainty. 
For spatiotemporal forecasting, \cite{mallick2024desqruq} combine deep ensembles with distribution-free simultaneous regression loss to characterize predictive uncertainty, inducing ensemble diversity through hyperparameter configurations sampled from a Gaussian copula (discussed in Section~\ref{sec_generative_copula}) generative model. The utility of such strategies is empirically supported by \cite{willard2025machine}, who evaluate six classical and deep learning ensemble construction methods, demonstrating that varying input data and combining diverse models significantly enhances both the accuracy and uncertainty quantification of spatiotemporal hydrologic forecasts. Deep ensembles are also widely used as a practical approximation to epistemic uncertainty by measuring variability across independently trained models. Their forecasts or predictive distributions are aggregated to estimate the overall predictive uncertainty. 

\subsubsection{Bayesian Model Averaging}\label{sec_bma}
Bayesian model averaging (BMA) provides a probabilistic ensemble framework for addressing model uncertainty in statistical inference \citep{hoeting1999bma}. It combines the predictive distributions of multiple competing models using weights that reflect their posterior support, thereby accounting for their relative contributions to the combined forecast \citep{raftery2005bma}.
Given point forecasts from $K$ distinct models, denoted as $\widehat{y}_{T+h, 1}, \dots, \widehat{y}_{T+h, K}$, the BMA predictive density for the $h$-step-ahead future state $y_{T+h}$ is formulated as the conditional mixture distribution:
$$p(y_{T+h} \mid \widehat{y}_{T+h, 1}, \dots, \widehat{y}_{T+h, K}) = \sum_{k=1}^K w_k \, g_k(y_{T+h} \mid \widehat{y}_{T+h, k}),$$
where $g_k(y_{T+h} \mid \widehat{y}_{T+h, k})$ represents the conditional predictive density associated with the $k^{th}$ model, and the weights are constrained such that $w_k \geq 0$ and $\sum_{k=1}^K w_k = 1$. Under this framework, the weights $w_k$ correspond to posterior probabilities that reflect each model's relative predictive performance over a designated training horizon \citep{raftery2005bma}. In practice, for continuous variables such as temperature, the conditional densities are typically modeled as Gaussian distributions centered at the bias-corrected forecasts.
For non-Gaussian variables, the component densities are tailored to the support of the target domain. For instance, skewed continuous variables such as wind speed are modeled using gamma distributions \citep{Sloughter2010probabilistic}. Conversely, precipitation, which exhibits both zero-inflation and a heavy upper tail, is accommodated via a discrete-continuous mixture comprising a point mass at zero and a power-transformed gamma distribution \citep{Sloughter2007bma}. Parameter estimation is typically conducted via the Expectation-Maximization (EM) algorithm \citep{dempster1977em}. 

In spatiotemporal forecasting, BMA has been applied to weather and climate forecasting, where multiple numerical simulation systems are often available. For instance, \cite{raftery2005bma} employed BMA for weather ensemble post-processing, while \cite{qin2025urban} applied the framework to combine climate model projections for long-term environmental forecasting. However, most existing implementations estimate mixture parameters independently across spatial locations, limiting their ability to represent spatial dependence and coherent uncertainty across large spatial domains. Moreover, employing BMA in large-scale spatiotemporal domains introduces significant modeling and computational hurdles, rendering standard optimization routines computationally prohibitive.

\subsection{Distribution-Free Calibration Methods}\label{sec_distribution_free}
Although ensemble-based methods provide flexible representations of predictive uncertainty, they do not inherently ensure calibration, as their reliability depends on how ensemble diversity is generated and forecasts are combined. This limitation motivates distribution-free calibration methods, which construct predictive intervals by calibrating model outputs against observed forecast errors without specifying a parametric form for the underlying data distribution. 
This section reviews two major approaches to distribution-free calibration: conformal prediction (Section~\ref{sec:agnostic_conformal}) and  conformalized quantile regression (Section~\ref{sec_cqr}).

\subsubsection{Conformal Prediction}\label{sec:agnostic_conformal}
Conformal prediction (CP) provides a rigorous, distribution-free framework for uncertainty quantification, providing finite-sample marginal coverage under exchangeability for a target miscoverage rate $\alpha \in (0,1)$. We refer readers to \cite{vovk2005conformal, shafer2008conformal} for a comprehensive discussion of foundational CP mechanics. To reduce the computational burden of standard CP, split conformal prediction \citep{papadopoulos2002split} partitions data into training ($\mathcal{I}_1$) and calibration ($\mathcal{I}_2$) sets, generating intervals based on the finite-sample adjusted empirical quantile $Q_{1-\alpha}$ of the calibration conformity scores. Split CP transforms point estimators into distribution-free prediction intervals, but its theoretical guarantees rely on exchangeability, which is often violated in temporal and spatiotemporal domains due to serial and spatial correlations. Historical calibration scores may therefore provide poor estimates of future forecast errors when the data exhibit dependence or distributional changes, leading to degraded empirical coverage \citep{xu2021conformal,gibbs2021adaptive}. Consequently, recent work has focused on adapting conformal prediction to dependent data through dynamic and adaptive calibration procedures.
\begin{figure}
    \centering
    \includegraphics[width=\linewidth]{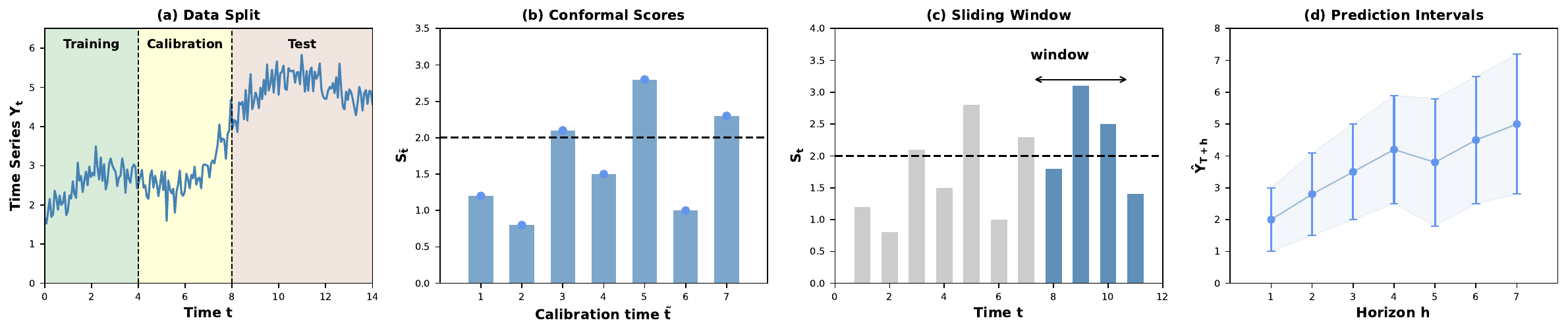}
    \caption{Illustration of rolling-window conformal prediction for temporal data. Panel $(a)$ shows the data split into training (green), calibration (yellow), and test (red) sets. Panel $(b)$ displays the conformal scores $S_{\tilde{t}}$ computed on the calibration set, with the empirical quantile marked as a dashed black line. Panel $(c)$ demonstrates the sliding window mechanism where only the most recent scores (highlighted in blue) contribute to computing the conformal quantile. Panel $(d)$ shows the resulting prediction intervals.}
    \label{fig:Conformal_Illustration}
\end{figure}

A common approach for temporal forecasting uses weighted sliding windows to emphasize recent observations, as illustrated in Fig.~\ref{fig:Conformal_Illustration}. Instead of assigning equal weight to all historical errors, this method computes calibration scores from recent model performance and normalizes them against an underlying uncertainty estimator. A rolling window then retains the most recent conformity scores to determine an adaptive threshold. This allows the resulting prediction intervals to adjust to local variability and nonstationary dynamics without relying on potentially outdated historical observations. 
Another conformal approach designed for dependent time series is ensemble batch prediction intervals (EnbPI) \citep{xu2021conformal}, which combines leave-one-out predictions from bootstrap ensembles with a sliding window of recent residuals. EnbPI continuously updates the interval using recent forecasting errors and provides approximately valid marginal coverage for time series with strongly mixing errors. 
Recent developments on conformal seasonal pools (CSP) \citep{valery2026csp} instead exploits recurring seasonal structure when constructing prediction intervals. Rather than relying solely on recent residuals, CSP combines seasonally matched historical observations with residual-based adjustments around a seasonal baseline forecast,
making it particularly suitable for seasonal time series.

Extending conformal prediction to spatiotemporal forecasting introduces an additional challenge, because dependence must be addressed simultaneously across both space and time. One strategy replaces global calibration with localized calibration. The spatiotemporal aleatoric conformal inference (STACI) framework \citep{feng2026staci}, for instance, constructs calibration sets using observations located within a local spatiotemporal neighborhood of the prediction target, relying on the weaker assumption of local exchangeability. A complementary direction integrates conformal calibration with graph-based forecasting architectures. Methods such as CP-STGCN \citep{zhao2025cpstgcn} leverage spatiotemporal graph convolutional network (STGCN) \citep{yu2018spatio} to capture spatial dependencies while using conformal prediction as a post-hoc calibration layer.
The practical appeal of CP approach has led to its adoption across diverse forecasting settings, including clinical and public-health time series \citep{stankeviciute2021conformal, goswami2026longesn}, electricity-price forecasting \citep{zaffran2022adaptive}, financial-volatility forecasting under distribution shift \citep{gibbs2021adaptive}, and air quality prediction \citep{panja2026estgcn}.

\subsubsection{Conformalized Quantile Regression}\label{sec_cqr}
Conformalized quantile regression (CQR) extends the conformal prediction framework by incorporating conditional quantile estimation into the calibration process \citep{romano2019cqr}. 
CQR first estimates lower and upper conditional quantile functions, denoted by $\widehat q_{\alpha_{\text{lo}}}(x_{t+h})$ and $\widehat q_{\alpha_{\text{hi}}}(x_{t+h})$, respectively, where $x_{t+h}$ denotes a set of features (or lagged variables) available at time $t$ to predict horizon $h$. This is done by minimizing the pinball loss \citep{ingo2011pinball}. These quantile estimates are subsequently calibrated using a held-out calibration set $\mathcal{I}_2$. 
The $(1-\alpha)^{th}$ empirical quantile, denoted by $\widehat{Q}_{1-\alpha, h}(\mathcal{I}_2)$, is then computed over the horizon-specific set of non-conformity calibration scores. For a forecast of the future state $y_{T+h}$ at horizon $h$ given features $x_{T+h}$, the final adjusted prediction interval is constructed as:
$$\widehat{C}_{T+h}^{1-\alpha} = \left[ \widehat q_{\alpha_{\text{lo}}}(x_{T+h}) - \widehat{Q}_{1-\alpha, h}(\mathcal{I}_2), \; \widehat q_{\alpha_{\text{hi}}}(x_{T+h}) + \widehat{Q}_{1-\alpha, h}(\mathcal{I}_2) \right].$$

The resulting interval retains the finite-sample marginal coverage guarantee of split conformal prediction under exchangeability and the finite-sample quantile correction, while using quantile regression to adapt interval width to different forecasting conditions. This property has made CQR suitable for temporal forecasting problems characterized by nonstationarity and heteroskedasticity. For example, the conformalized temporal convolutional quantile regression network (CTCQRN) integrates CQR with temporal convolution networks (TCNs) for wind power interval forecasting \citep{hu2022ctc}. The TCN architecture captures long-range dependence, while the conformal calibration layer corrects the quantile estimates to provide reliable uncertainty bounds. 
More recent approaches, such as ensemble conformalized quantile regression (EnCQR), combine ensembles of quantile regression models with adaptive calibration strategies to further improve robustness under evolving temporal dynamics \citep{jensen2024ensqr}. 

For relational and spatiotemporal forecasting, CQR has been extended to exploit dependencies across multiple correlated time series. Conformal relational prediction (CoRel) \citep{cini2025relational} uses graph-based learning to estimate residual quantiles across related series, incorporating an adaptive calibration mechanism to accommodate non-exchangeability and localized distributional changes. 
In photovoltaic power forecasting, the multi-scale temporal-spatial attention quantile regression network (MTSA-QRN) combines a dual-pathway quantile regression architecture with adaptive conformal calibration \citep{wang2026cqr}. Its temporal pathway uses temporal convolutions and attention mechanisms, while its graph-attention pathway models dependencies among variables. 
Physics-informed constraints have also been incorporated into spatiotemporal CQR. The physics-driven dynamic wake forecasting CQR (PDWF-CQR) framework \citep{wang2026physics} constructs a time-varying directed graph from turbine wake interactions and imposes power-curve-based regularization on the quantile boundaries. 

The model-agnostic methods reviewed above address different stages of uncertainty construction. Residual-based procedures translate past forecast errors into uncertainty estimates, ensembles use variation across models or resamples, and calibration methods adjust predictive intervals with respect to a stated coverage level. Their main advantage is compatibility with a wide range of forecasters. Their reliability, however, depends on the quality and stability of the residuals, ensemble members, or calibration data on which they are based.

\section{Model-Intrinsic Uncertainty Quantification}
\label{sec_model_intrinsic}
In contrast to model-agnostic methods, model-intrinsic uncertainty quantification refers to approaches in which uncertainty is represented within the model formulation, learning objective, or posterior inference procedure. The model is trained to produce probabilistic outputs as part of the forecasting process. We organize these methods into four families: Bayesian modeling (Section~\ref{sec_bayesian}), parametric predictive distributions (Section~\ref{sec_parametric}), distributional regression (Section~\ref{sec_distributional}), and generative models (Section~\ref{sec_generative}). 
Bayesian models propagate uncertainty over parameters, functions, or latent states through the posterior predictive distribution. Parametric predictive models estimate the parameters of an assumed distributional family, whereas distributional regression directly learns quantiles, intervals, or predictive distributions through loss-based objectives. Generative models instead learn a stochastic mechanism from which plausible future trajectories can be sampled.

\subsection{Bayesian Modeling}\label{sec_bayesian}
\begin{figure}[!t]
    \centering
    \includegraphics[width=\linewidth]{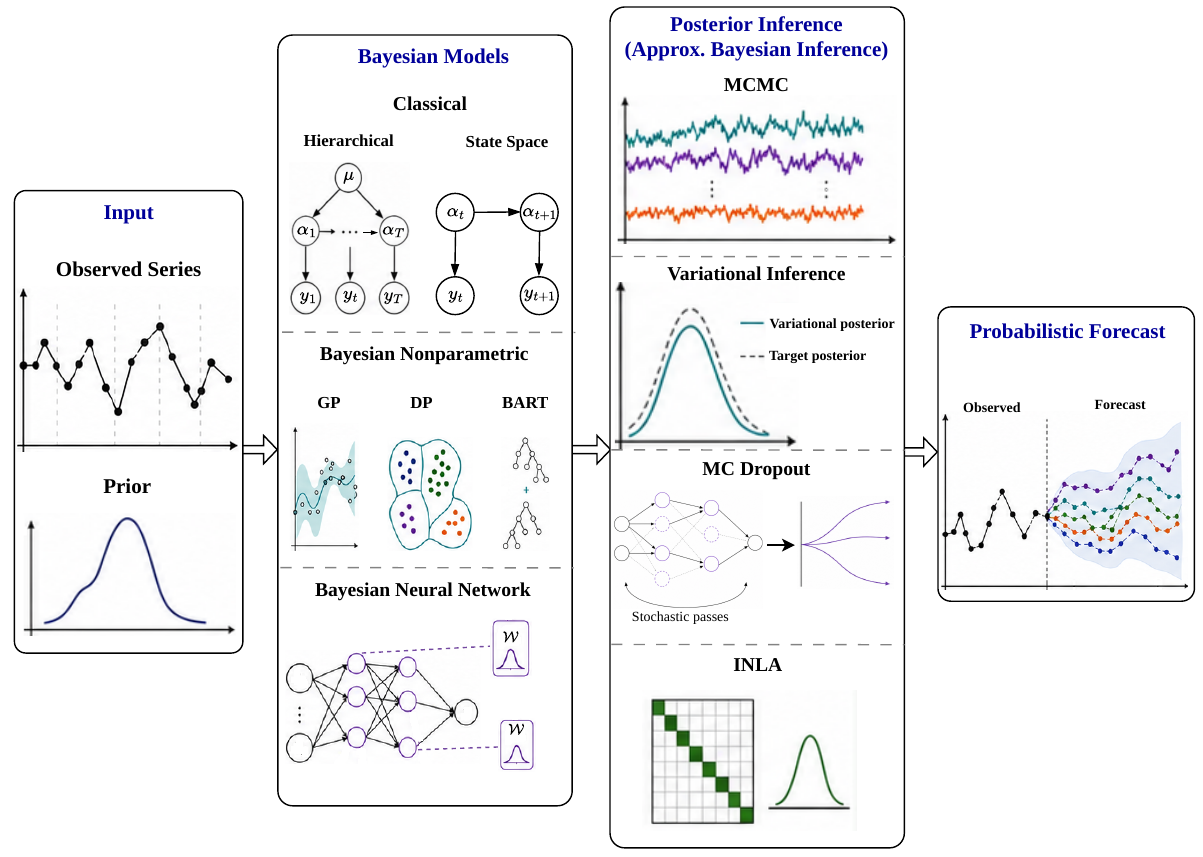}
    \caption{Overview of Bayesian methods for probabilistic forecasting, covering classical Bayesian models, Bayesian nonparametric methods, and Bayesian neural networks. Posterior predictive distributions are obtained through exact inference when tractable or through approximate inference methods.}
    \label{fig:Bayesian_Overview}
\end{figure}

Bayesian modeling provides a framework for probabilistic forecasting by placing prior distributions over unknown quantities and updating them with observed data to obtain a posterior predictive distribution, as depicted in Fig.~\ref{fig:Bayesian_Overview}. 
Bayesian approaches can be broadly categorized into three classes. Classical Bayesian models (Section~\ref{sec_classical_bayesian}) explicitly represent dependence through hierarchical structures, latent state processes, or covariance functions. Bayesian nonparametric methods (Section~\ref{sec_bayesian_nonparametric}) relax fixed parametric assumptions by placing priors over infinite-dimensional spaces. Finally, Bayesian neural networks (Section~\ref{sec_bnn}) extend these principles to deep learning by treating network weights as random variables. Across these model classes, exact posterior inference is rarely tractable beyond simple linear Gaussian settings. Consequently, Bayesian forecasting relies heavily on the approximate inference methods reviewed in Section~\ref{sec_approximate_bayesian}.

\subsubsection{Classical Bayesian Models}\label{sec_classical_bayesian}
Classical Bayesian models construct predictive distributions by placing priors over unknown quantities and updating them using observed data. This framework enables uncertainty to propagate coherently from parameter estimation to forecasting. Two foundational classes are reviewed in this section. Bayesian hierarchical models (Section~\ref{sec_bhm}) decompose the joint distribution into observation, process, and parameter components, whereas Bayesian state space models (Section~\ref{sec_bayesian_state_space}) capture dynamics through an evolving latent state process. 
In both cases, dependence structures are specified explicitly, supporting principled uncertainty quantification but at the expense of scalability. 

\paragraph{Bayesian Hierarchical Models.}\label{sec_bhm} 

Bayesian hierarchical models (BHMs) decompose the joint distribution into a data model $p(y \mid z,\mathcal{W}_D)$ describing the observation process, a process model $p(z \mid\mathcal{W}_P)$ characterizing the latent dynamics, and a parameter model $p(\mathcal{W}_D,\mathcal{W}_P)$ specifying prior distributions \citep{wikle2023statistical}. For observed data $y$ and latent process $z,$ the structure is given by:
$$
p(z, \mathcal{W}_D, \mathcal{W}_P \mid y) \;\propto\; p(y \mid z, \mathcal{W}_D)\;p(z \mid \mathcal{W}_P)\;p(\mathcal{W}_D, \mathcal{W}_P),
$$
where $\mathcal{W}_D$ and $\mathcal{W}_P$ denote data and process parameters, respectively. This modular construction allows each component to be specified independently and adapted to the data characteristics \citep{wikle2023statistical, lovegrove2023improving}. For example, count observations in epidemiological applications are accommodated through Poisson or Negative Binomial likelihoods that handle overdispersion \citep{almanji2025predicting}, while spatial and temporal dependencies can be introduced through conditional autoregressive models \citep{cui2021bus,louzada2025bayesian}. 


A common limitation of BHMs in spatiotemporal applications is the use of a separable covariance structure, where joint spatiotemporal dependence is represented through distinct spatial and temporal components \citep{cressie2011statistics}. \cite{cui2021bus} addressed this limitation by introducing explicit spatiotemporal interaction (STI) terms for online bus speed prediction. A second limitation is the assumption of spatial isotropy, under which dependence is determined by Euclidean distance and does not vary with direction. This can be restrictive when spatial dependence is directional or varies across the spatial domain \citep{fuglstad2025nonstationary}. To accommodate nonstationary and anisotropic spatial dependence, \cite{castruccio2018spatiotemporalbrain} proposed a multi-resolution BHM for functional magnetic resonance imaging (fMRI) data that forms a spatially weighted combination of local, geometrically anisotropic Matérn processes across subregions. 
BHMs have also been developed for multivariate spatiotemporal processes, where dependencies among multiple variables must be modeled jointly. For instance, \cite{lenzi2020wind} proposed a bivariate model for wind vector forecasting based on a linear model of coregionalization, expressing each orthogonal wind component as a weighted combination of latent spatiotemporal processes to induce joint space-time cross-covariance structures. 


\paragraph{Bayesian State Space Models.}\label{sec_bayesian_state_space}
Bayesian state space models (SSMs) provide an alternative framework for probabilistic forecasting by representing observations through an evolving latent state process. The (univariate) framework is defined by an observation equation and a state transition equation:
\begin{equation}
y_t = Z_t^\top \alpha_t + \beta^\top x_t + \epsilon_t, \quad \epsilon_t \sim \mathcal{N}(0, \Sigma_t), \qquad \text{and} \qquad \alpha_{t+1} = T_t \alpha_t + R_t \eta_t, \quad \eta_t \sim \mathcal{N}(0, Q_t),
\label{eq_bssm_observation}
\end{equation}
where $\alpha_t$ denotes the latent state vector, $Z_t, T_t, R_t$ are model matrices, $x_t$ represents exogenous covariates with coefficients $\beta$, and $\Sigma_t$ and $Q_t$ represent the observation and state noise covariance matrices, respectively \citep{scott2014bsts}. Kalman filter \citep{kalman1960filter, diebold1992kalman} and smoother enable exact inference under Gaussian assumptions by recursively estimating the filtering distribution $p(\alpha_t \mid \mathbf{Y}_{1:t})$ and the smoothing distribution $p(\alpha_t \mid \mathbf{Y}_{1:T})$, respectively. Probabilistic forecasts are then obtained by propagating the posterior state distribution through the transition dynamics, yielding the full predictive distribution $p(y_{T+h} \mid \mathbf{Y}_{1:T})$ over future observations. 

The Bayesian structural time series (BSTS) framework extends the state-space formulation through structural decomposition and spike-and-slab regression, allowing large sets of candidate covariates to be incorporated. \citep{scott2014bsts}. BSTS additively decomposes the target series into level, trend, seasonal, and regression components. 
Posterior inference is performed via a Markov Chain Monte Carlo (MCMC) algorithm (Section~\ref{sec_approximate_mcmc}) using simulation smoothing and stochastic search variable selection \citep{george1997ssvs}, while accounting for model uncertainty through Bayesian model averaging (Section~\ref{sec_bma}). 
BSTS has been adopted across applications including economic and financial forecasting \citep{qiu2018mbsts,katarina2023bsts}, climate policy uncertainty \citep{besher2026cpu}, and tourism demand prediction \citep{andrews2023bsts}. 
The multivariate BSTS (MBSTS) \citep{qiu2018mbsts} generalizes BSTS by extending the univariate formulation (Eq.~\ref{eq_bssm_observation}) to jointly model correlated series by assigning series-specific state components and capturing cross-series dependence through the observation noise covariance matrix $\Sigma_t$. MBSTS further incorporates cyclical components with damping factors to capture transient effects, which improved forecasting performance for financial time series during periods of market instability \citep{qiu2018mbsts}. Despite these extensions, the state space formulation (Eq. \ref{eq_bssm_observation}) remains limited for spatiotemporal systems due to restrictive assumptions of linearity and Gaussian noise. \cite{pal2021rnnpf} addressed this limitation through a nonlinear SSM that combines graph convolutional recurrent units \citep{bai2020gcrn} for spatiotemporal dependency learning with particle flow inference 
for posterior approximation, improving probabilistic forecasting in traffic applications.

\subsubsection{Bayesian Nonparametric Methods} \label{sec_bayesian_nonparametric}
Bayesian nonparametric (BNP) methods provide a flexible framework for probabilistic temporal and spatiotemporal forecasting by relaxing the assumption of a fixed parametric model. Instead, they place priors over infinite-dimensional probability measures or function spaces, allowing model complexity to adapt to the observed data \citep{hjort2010bayesian}. 
The choice of prior defines the resulting modeling framework, giving rise to different BNP approaches. This section discusses three prominent BNP approaches: Gaussian Processes (Section~\ref{sec_gp}), Dirichlet Processes (Section~\ref{sec_Dirichlet}), and Bayesian Additive Regression Trees (Section~\ref{sec_BART}).

\paragraph{Gaussian Processes.}\label{sec_gp} 
A Gaussian process (GP) is a stochastic process defined such that any finite collection of function values follows a joint Gaussian distribution. Instead of assuming a fixed parametric form, GPs specify a prior over functions, offering a flexible nonparametric Bayesian framework for probabilistic forecasting \citep{rasmussen2004gp}. 
GP is defined as $\widetilde{f} \sim \mathcal{GP}(\Upsilon(t), \Omega(t,t')),$ where $\Upsilon(t)$ is the mean function and $\Omega(t,t')$ is the covariance kernel that determines the dependency structure of the process.
The kernel specification controls properties such as smoothness, periodicity, and stationarity of the resulting forecasts, and its hyperparameters are learned from data \citep{williams1995gp}.
Given the observed history 
and a GP prior with a Gaussian likelihood,
the posterior distribution over future states remains Gaussian with closed-form predictive mean and covariance. 
Exact GP inference, however, is limited by cubic computational complexity
$\mathcal{O}(T^3)$ arising from covariance matrix inversion. 
Sparse variational GPs address this limitation by introducing a set of $M$ inducing points to summarize the training data, reducing the computational cost to $\mathcal{O}(TM^2+M^3)$ \citep{wilkinson2021sparse}. 
For the spatiotemporal setting, methods such as latent Kronecker GPs accommodate missing observations by operating on an underlying complete grid representation \citep{lin2025scalable}. State-space formulations provide another scalable alternative by exploiting Markovian structure, enabling linear-time inference through Kalman filtering and smoothing \citep{hamelijnck2021stvgp}. 

GP-based forecasting has been applied across a wide range of domains, including epidemiological forecasting \citep{agyemang2025gaussian}, intermittent demand forecasting with TweedieGP \citep{damato2025tweedie}, and short-term traffic forecasting with deep graph Gaussian processes (DGGP) \citep{jiang2022dggp}.
In large-scale spatiotemporal applications, sparse variational GP frameworks have been developed for tasks such as crime and air quality forecasting by combining GP priors with structured approximations \citep{flaxman2015fast,schoucair2025nowcasting}. Related efforts have also explored replacing covariance-based interpolation altogether. For instance, \cite{nag2023DeepKriging} proposed Space-Time.DeepKriging, which uses neural networks with spatial and temporal basis functions to achieve scalable spatiotemporal forecasting while retaining uncertainty estimates through quantile-based prediction. For irregular spatial observations, scalable inference can instead be achieved through Vecchia approximations \citep{vecchia1988estimation}, which condition each observation on a small set of neighboring observations rather than the full dataset. GpGp \citep{guinness2018gpgp} adopts this strategy, enabling scalable probabilistic forecasting on large spatiotemporal datasets where Kronecker-based methods are less suitable. 

\paragraph{Dirichlet Processes.}\label{sec_Dirichlet}
In contrast to GPs, which define priors over function spaces, Dirichlet processes (DPs) place priors over probability measures, enabling highly adaptive mixture modeling \citep{ferguson1973bayesian}. A DP, denoted as $\mathcal{DP}(\alpha, G_0)$, is characterized by a concentration parameter $\alpha$ and a base distribution $G_0$, acting as a distribution over discrete distributions. In time series and spatiotemporal forecasting, this property is primarily leveraged through DP mixture models (DPMMs) to dynamically infer an unknown or theoretically infinite number of latent components, temporal regimes, or spatial clusters directly from the data \citep{emonet2013temporal}. For temporal modeling, the hierarchical DP hidden Markov model (HDP-HMM) \citep{beal2001ihmm} extends the HDP \citep{teh2006hierarchical} to sequential data by coupling state-specific transition distributions through a shared hierarchical prior, allowing the number of latent states to be inferred from the data (HMMs are discussed in Section~\ref{sec_generative_hmm}). Standard HDP-HMMs, however, often over-segment sequences because they rapidly switch among redundant states. The sticky HDP-HMM \citep{fox2011sticky} addresses this limitation by introducing a self-transition bias that encourages state persistence while retaining nonparametric state inference and allowing each state's observation distribution to be modeled flexibly via DP mixtures. Recent works have also integrated DPs to enhance ensemble forecasting \citep{ren2023infinite}. 

\paragraph{Bayesian Additive Regression Trees.}\label{sec_BART}
Bayesian additive regression trees (BART) provide a flexible BNP regression framework based on ensembles of regression trees \citep{chipman2010bart}. BART represents the regression function as a sum of weak learners, where regularization priors constrain individual trees to avoid overfitting while allowing the ensemble to capture complex nonlinear relationships. Although originally developed for classical machine learning tasks, BART has been extended to probabilistic forecasting through multiple variants.
For instance, \cite{pruser2019forecasting} applied BART to macroeconomic forecasting in predictor-rich settings, demonstrating its ability to model nonlinear temporal relationships, while \cite{mahdavi2024measuring} integrated BART into autoregressive distributed lag models to improve forecasting under nonlinear dynamics. Minnesota BART \citep{lima2025minnesota} introduces sparsity-inducing priors and structured shrinkage for high-dimensional vector autoregressions. More recently, implicit quantile BART (IQ-BART) \citep{o2025generative} places a BART prior on the conditional quantile function and uses a check-loss likelihood to learn it jointly across quantile levels. The resulting quantile function supports predictive sampling and can capture multimodal distributions that may be missed by conventional parametric models.

\subsubsection{Bayesian Neural Networks}\label{sec_bnn} 
Bayesian neural networks (BNNs) extend classical neural networks by treating the network weights as random variables rather than deterministic parameters \citep{neal1996bayesian,ghosh2019model}. 
Given the observed history $\mathbf{Y}_{1:T}$, inference is performed over the posterior distribution of the weights, $p(\mathcal{W} \mid \mathbf{Y}_{1:T})$. Probabilistic forecasts for a future state $y_{T+h}$ are then obtained by marginalizing over this posterior:$$p(y_{T+h} \mid \mathbf{Y}_{1:T}) = \int p(y_{T+h} \mid \mathbf{Y}_{1:T}, \mathcal{W}) \, p(\mathcal{W} \mid \mathbf{Y}_{1:T}) \, d\mathcal{W},$$where $p(y_{T+h} \mid \mathbf{Y}_{1:T}, \mathcal{W})$ represents the conditional predictive distribution parameterized by a specific weight configuration.
Exact posterior inference is generally intractable because of the high-dimensional and nonlinear parameter space. Consequently, BNNs rely on approximate inference techniques (Section~\ref{sec_approximate_bayesian}), most commonly variational inference and MCMC methods \citep{graves2011practical}. 
The choice of prior distribution is also important, particularly in highly overparameterized settings, where hierarchical shrinkage priors such as the horseshoe prior can improve regularization by reducing the contribution of irrelevant parameters \citep{ghosh2019model}. 
The flexibility of neural networks, combined with their ability to represent parameter uncertainty, has motivated the use of BNNs across several forecasting domains, including macroeconomic prediction using shrinkage priors and stochastic volatility modeling \citep{Hauzenberger2025bnn} and spatiotemporal urban flood forecasting \citep{chi2024bnn}. 

BNNs have also been extended to recurrent architectures by placing priors over recurrent weights and performing inference over their posterior distribution, modeling uncertainty in both the sequential observations and network parameters.  
For instance, the Bayesian autoregressive and RNN (BARNN) framework \citep{coscia2025barnn} models the joint distribution of observations and time-varying network weights autoregressively, sampling the weights at each time step from a variational posterior conditioned on past states. 
Similarly, \cite{mcdermott2019bayesianrnn} proposed BAST-RNN, a BRNN model for spatiotemporal forecasting by placing priors over recurrent weight matrices and applying MCMC-based inference. Recent work has improved BNNs for large-scale spatiotemporal forecasting through scalable inference and informative priors, reducing computational cost and sensitivity to prior specification \citep{duncker2023scalable}. Moving beyond weight-space posterior inference altogether, prior-data fitted networks (PFNs) \citep{muller2022pfn} amortize Bayesian prediction into in-context prediction without task-specific weight updates, a paradigm that has recently gained prominence in foundation models for time series forecasting (see Section~\ref{sec_foundation_uq}).

\subsubsection{Approximate Bayesian Inference}\label{sec_approximate_bayesian} 
Exact Bayesian inference is generally intractable for the complex models encountered in modern temporal and spatiotemporal forecasting. Approximate Bayesian inference addresses this challenge by replacing exact posterior computation with scalable approximations. 
Existing methods differ fundamentally in both their computational efficiency and the quality of their posterior approximations. This section reviews four representative approaches: Markov Chain Monte Carlo (Section~\ref{sec_approximate_mcmc}), Monte Carlo Dropout (Section~\ref{sec_approximate_drop}), Variational Inference (Section~\ref{sec_approximate_vi}), and Integrated Nested Laplace Approximation (Section~\ref{sec_approximate_inla}). Their characteristics and trade-offs are summarized in Table~\ref{tab:bayesian_approx_comparison}.

\begin{table}[!htbp]
\caption{Comparison of approximate Bayesian inference methods.
Check marks (\ding{51}) denote restricted applicability; crosses (\ding{55}) denote no restriction.}
\begin{adjustbox}{width=\textwidth}
   \centering
    \begin{tabular}{lccccc} \toprule
    \textbf{Method} & \textbf{Approximation Strategy} & \textbf{Model Restrictions} & \textbf{Posterior Flexibility} & \textbf{Computational Cost} \\\midrule
   Markov Chain Monte Carlo & Posterior Sampling & \ding{55} & High & High \\
   
   Monte Carlo Dropout & Dropout-based Variational & \ding{51}\textsuperscript{$\dagger$} & Low &  Low \\
   
   Variational Inference & Optimization-based & \ding{55} & Moderate-High\textsuperscript{$\S$} & Moderate \\
   
   Integrated Nested Laplace Approximation & Deterministic Approximation & \ding{51}\textsuperscript{$\ddagger$} & Low-Moderate & Low-Moderate\textsuperscript{$\P$} \\
    \bottomrule
    \end{tabular}
    \label{tab:bayesian_approx_comparison}
    \end{adjustbox}
\footnotesize{\textsuperscript{$\dagger$}Requires a network trained with dropout regularization; not a post-hoc wrapper for arbitrary models.\\
\textsuperscript{$\S$}Depends on the chosen variational family.\\
\textsuperscript{$\ddagger$}Restricted to latent Gaussian models with exponential-family likelihoods.\\
\textsuperscript{$\P$}Sub-quadratic complexity, though cost remains substantial for very large datasets.}
\end{table}

\paragraph{Markov Chain Monte Carlo.}\label{sec_approximate_mcmc} MCMC methods represent the classical approach to approximate Bayesian inference by drawing samples from posterior distributions that cannot be evaluated analytically \citep{jones2022mcmc}. 
MCMC 
constructs a Markov chain whose stationary distribution is the target posterior, allowing posterior summaries and predictive quantities to be estimated from Monte Carlo samples. The Metropolis-Hastings algorithm proposes candidate parameter values and accepts them with a probability determined by the posterior-density ratio, adjusted for asymmetry in the proposal distribution \citep{hastings1970mh}. 
Metropolis-Hastings provides a general inference framework and has been widely adopted for spatiotemporal forecasting, including Bayesian epidemiological models \citep{stojanovic2019bayesian} and precipitation nowcasting \citep{johnson2023bayesian}. Alternatively, when full conditional distributions are analytically tractable, Gibbs sampling becomes the standard inference approach \citep{geman1984stochastic}. This method sequentially updates each parameter by drawing directly from its respective conditional posterior, thereby eliminating the need for an acceptance step. The method's simplicity and computational efficiency have led to its widespread adoption in frameworks such as BSTS models via simulation smoothing \citep{scott2014bsts,qiu2018mbsts}, Gaussian mixture models for travel time forecasting \citep{chen2023prob}, and Bayesian hierarchical spatiotemporal models based on Gaussian Markov random fields \citep{rue2005gaussian}. Although MCMC provides asymptotically exact Bayesian inference, its computational cost can be prohibitive for large spatiotemporal forecasting problems with millions of parameters. Numerous extensions have been proposed to improve scalability, including parameter expansion \citep{mcdermott2019bayesianrnn} and ensemble Kalman smoothing within Gibbs sampling \citep{johnson2023bayesian}. These limitations have motivated approximate inference methods that trade exact posterior sampling for greater computational efficiency.

\paragraph{Monte Carlo Dropout.}\label{sec_approximate_drop} 
Monte Carlo dropout (MCD) offers a scalable alternative by reinterpreting dropout as approximate Bayesian inference. \cite{gal2016dropout} established that a network trained with dropout at rate $p$ implicitly minimizes the Kullback-Leibler (KL) divergence between a variational distribution $q(\mathcal{W})$ induced by the dropout masks and the true posterior $p(\mathcal{W} \mid \mathbf{Y}_{1:T})$. This leads to variational inference without requiring any modification to the training objective beyond $\ell_2$ regularization. 
At inference time, dropout is kept active and $M$ stochastic forward passes are performed. For a given forecast horizon $h$, each pass applies an independently sampled mask to yield a prediction $\widehat{y}_{T+h}^{(m)}$. The resulting ensemble $\{\widehat{y}_{T+h}^{(m)}\}_{m=1}^{M}$ captures predictive variability induced by the approximate weight posterior. When combined with an output likelihood that accounts for observation noise, it provides an empirical approximation to the posterior predictive distribution $p(y_{T+h} \mid \mathbf{Y}_{1:T})$. From this ensemble, the predictive mean and the epistemic variance are estimated as the sample mean and sample variance across the $M$ forward passes, respectively. For recurrent architectures, variational dropout applies a fixed dropout mask across all time steps within a sequence, preserving temporal dependencies while maintaining the Bayesian interpretation of the model \citep{gal2016recurrent}. 
MCD has been combined with mean variance estimation to jointly model epistemic and aleatoric uncertainty for wind speed, wind power, and load forecasting \citep{serpell2019mc}. 
The method has also been applied to financial time series prediction using Bayesian LSTMs for Bitcoin price forecasting \citep{hassan2024bitcoin} and to traffic breakdown probability forecasting using variational LSTMs \citep{zechin2023traffic}. For spatiotemporal forecasting, MCD has been integrated with GNNs to quantify uncertainty in traffic flow forecasting \citep{hemavathi2026traffic} and with convolutional architectures for quantitative precipitation forecasting \citep{kim2026quantifying}. 
However, the framework is constrained by the restrictive variational family induced by the dropout mechanism, which underestimates posterior uncertainty and offers limited flexibility for incorporating informative priors.

\paragraph{Variational Inference.}\label{sec_approximate_vi} 
Variational inference (VI) addresses the scalability limitations of MCMC and the restricted approximation family of MC Dropout by formulating posterior inference as an optimization problem \citep{blei2017vi}. Given a family of tractable distributions $\mathcal{Q}$, VI seeks to find the member $q^\star(\mathcal{W}) \in \mathcal{Q}$ that minimizes the KL divergence to the true posterior $p(\mathcal{W} \mid \mathbf{Y}_{1:T})$. Because this divergence is itself intractable, optimization is performed by maximizing the evidence lower bound (ELBO):
$$\operatorname{ELBO}(q) = \mathbb{E}_{q(\mathcal{W})} \left[\log p\left(\mathbf{Y}_{1:T} \mid \mathcal{W}\right)\right]- \operatorname{KL} \left(q(\mathcal{W}) \|\ p(\mathcal{W}) \right),$$ which decomposes into a data-fit term and a regularization term that penalizes deviation from the prior. 
The most common variational family is the mean-field approximation $q(\mathcal{W}) = \prod_j q_j (\mathcal{W}_j)$, which factorizes over parameters and admits efficient coordinate ascent updates \citep{blei2017vi}. In deep neural networks, inference is performed using stochastic variational inference \citep{hoffman2013svi}, where MC estimates of the ELBO gradients are computed via the reparameterization trick \citep{kingma2014autoencoding,rezende2014stochastic}, enabling VI to scale to high-dimensional parameter spaces \citep{graves2011practical}. 

For spatiotemporal forecasting, \cite{liu2020bayesvariational} proposed the variational Bayesian spatial temporal neural network (STNN-VB), which places a Gaussian mixture variational posterior over the weights of a convolutional gated recurrent unit (ConvGRU) for wind speed forecasting. 
Alternatively, \cite{saad2024scalable} introduced the Bayesian neural field (BayesNF) framework, embedding mean-field variational inference within a hierarchical BNN to map continuous spatiotemporal coordinates to the target field. 
VI is also used in the deep latent variable models discussed in Section~\ref{sec_generative_vae}. Despite its scalability advantages, VI has several limitations. The widely used mean-field approximation assumes independence between parameters, which can underestimate posterior uncertainty and result in overconfident predictions \citep{macKay2003information,bishop2006prml}. Similar issues arise in deep latent variable models, where an overly restrictive approximation can cause posterior collapse, leading the latent variables to become uninformative as the approximate posterior converges toward the prior \citep{maldonado2026analysis}. 

\paragraph{Integrated Nested Laplace Approximation.}\label{sec_approximate_inla} Integrated Nested Laplace Approximation (INLA) provides a deterministic alternative to MCMC for latent Gaussian models \citep{rue2009approximate}. It assumes a three-level hierarchical structure in which observations $Y(s,t)$ at spatial locations $s$ for $s \in \{1,\ldots,N\}$ are conditionally independent, given a latent Gaussian field $\mathbf{X}$ comprising components $X(s,t)$ and likelihood hyperparameters $\theta_1$. The observation model is typically assumed to belong to the exponential family. The latent field is modeled as a Gaussian Markov random field (GMRF) with mean $\Upsilon(\theta_2)$ and precision matrix $\Sigma^{-1}(\theta_2)$ that captures the spatiotemporal dependence structure of the data \citep{rue2005gaussian}. A prior distribution is then assigned to the combined hyperparameters $\theta = (\theta_1, \theta_2)$.
Given the observed history $\mathbf{Y}_{1:T}$, the resulting joint posterior is formulated as:
$$p(\mathbf{X}, \theta \mid \mathbf{Y}_{1:T}) \propto p(\theta) \, p(\mathbf{X} \mid \theta_2) \prod_{s = 1}^N \prod_{t = 1}^T p(Y(s,t) \mid X(s,t), \theta_1),$$
where $p(\theta)$ denotes the prior on the hyperparameters and $p(\mathbf{X} \mid \theta_2)$ is the GMRF prior on the latent field.
The key advantage of INLA over MCMC is that it provides a fast and accurate approximation of posterior marginals without requiring sampling \citep{rue2009approximate}. 

A key enabler of INLA for spatiotemporal forecasting is the stochastic partial differential equation (SPDE) approach of \cite{lindgren2011approximate}, which represents Gaussian fields with Matérn covariance as GMRFs defined over a finite element mesh. 
The formulation extends to spatiotemporal settings by allowing the spatial field to evolve through a first-order autoregressive process \citep{cameletti2013spde}. Consequently, the resulting INLA-SPDE framework has become the dominant approach for many Bayesian spatiotemporal forecasting models.
For instance, \cite{mahfoud2021burglary} forecasted residential burglary across Amsterdam neighborhoods, while \cite{castro2019spliced} combined a spliced Gamma-Generalized Pareto likelihood with an SPDE-based spatiotemporal field for short-term extreme wind speed forecasting.
INLA has also been widely applied in epidemiology and environmental monitoring, including COVID-19 hotspot detection using exceedance probabilities \citep{jaya2021covid}, PM$_{2.5}$ interpolation through spatial prediction models \citep{jaya2024pm}, and dengue forecasting 
in Costa Rica \citep{chou2023dengue} and Yogyakarta \citep{salim2025dengue}. 
Despite its efficiency, INLA is restricted to latent Gaussian models with observation likelihoods from the exponential family, limiting its applicability to more general non-Gaussian and deep latent models \citep{rue2009approximate}. 

\subsection{Parametric Predictive Distributions}\label{sec_parametric}
Parametric predictive distributions quantify uncertainty by specifying a distributional form for future observations and estimating its parameters from data. In these models, the forecasting architecture outputs the parameters of the assumed distribution, which are then used to derive predictive intervals and quantiles. This section reviews two main approaches which differ in flexibility of the predictive distribution. Single parametric distributions (Section~\ref{sec_parametric_single}) assume a fixed family, such as the Gaussian distribution, whereas mixture density networks (Section~\ref{sec_parametric_mdn}) extend this idea by learning input-dependent combinations of multiple distributions.

\subsubsection{Single Parametric Distributions}\label{sec_parametric_single}
The classical formulation of this approach models forecast uncertainty through a predefined distributional family, with its parameters estimated from historical observations. 
Modern frameworks directly parameterize distributions inherently tailored to the data. For instance, \cite{thorarinsdottir2010wind} use heteroskedastic censored (tobit) regression and truncated normal distributions for wind speed forecasting. For nonnegative variables with mass at zero, \cite{gneiting2006rst} introduced the cut-off Gaussian distribution, which was adopted in the regime-switching space-time model to improve forecast calibration. More recently, \cite{salinas2020deepar} introduced DeepAR, an RNN that maps previous hidden states and covariates directly to the parameters of a chosen emission distribution, such as a Gaussian for real-valued targets or a negative binomial for count data. To achieve this, the model maximizes the log-likelihood over the prediction window and employs specialized activation functions, such as a softplus layer, to satisfy structural constraints like positive scale parameters. 
More generally, frameworks like generalized additive models for location, scale, and shape (GAMLSS) allow all parameters of these chosen distributions to vary flexibly as functions of covariates \citep{merder2026distributional}. 

Despite these developments, each parametric family remains limited by the assumptions of its chosen distribution and may struggle to capture complex predictive behaviors. 
A further limitation is the assumption that errors are temporally independent. \cite{zheng2024correlated} addressed this by learning the covariance structure of errors over multiple steps using a low-rank-plus-diagonal parameterization and latent temporal processes, improving calibration without increasing parameter size. An alternative non-parametric direction constructs predictive distributions by reweighting the empirical distribution of past observations, as in non-parametric time series forecaster (NPTS) and its deep learning extension, DeepNPTS \citep{rangapuram2023deepnpts}. However, DeepNPTS' improved flexibility for non-Gaussian data comes at the expense of sample efficiency. These advances highlight the need for more flexible probabilistic models capable of capturing richer distributional structures.

\subsubsection{Mixture Density Networks}\label{sec_parametric_mdn}

While computationally simple, single parametric distributions lack the flexibility to capture complex, input-dependent predictive uncertainties such as multimodality or asymmetric tails. Mixture density networks (MDNs) \citep{bishop1994mdns} overcome this by modeling the conditional predictive density as a weighted combination of multiple parametric components. Given the observed history $\mathbf{Y}_{1:T}$, a neural network estimates the parameters of the mixture for a future state $y_{T+h}$:
$$p(y_{T+h} \mid \mathbf{Y}_{1:T}) = \sum_{k=1}^{K} \alpha_k(\mathbf{Y}_{1:T}) \, \phi_k(y_{T+h} \mid \mathbf{Y}_{1:T}),$$
where the mixing coefficients $\alpha_k(\cdot)$ are constrained such that $\alpha_k \geq 0$ and $\sum_{k=1}^K \alpha_k = 1$. Each component density $\phi_k$ is parameterized by a mean vector $\mu_k(\cdot)$ and a covariance matrix $\Sigma_k(\cdot)$. Trained by minimizing the negative log-likelihood, MDNs generalize from a standard single distribution ($K = 1$) to a highly flexible approximator of complex conditional densities as the number of components $K$ increases \citep{bishop1994mdns}.

MDNs have been applied extensively in forecasting problems where uncertainty is inherently non-Gaussian. For wind power forecasting, \cite{zhang2020idmdn} proposed an improved deep MDN that replaces Gaussian components with beta distributions to respect the bounded nature of power generation. 
Similarly, for directional forecasting problems, \cite{zhi2019directional} combined LSTM networks with von Mises mixture distributions to model circular variables such as pedestrian movement directions, where Gaussian assumptions are inappropriate due to the periodic structure of the output space. In electricity price forecasting, \cite{afrasiabi2022DeepGabor} integrated Gabor feature extraction with a mixture density output layer to capture the multimodal and volatile nature of price distributions. 
Recent work has focused on improving the training and scalability of MDNs. Maximum likelihood training can become unstable when mixture components collapse or when data is sparse. \cite{yang2022waimdn} addressed this by replacing the negative log-likelihood with a Wasserstein distance-based adversarial objective that compares the generated and observed distributions. 
For spatiotemporal forecasting, \cite{he2026hybrid} combined MDNs with GCNs to output multivariate Gaussian mixtures with low-rank covariance. 
The number of mixture components in MDNs must be carefully selected, as insufficient components limit expressiveness while excessive components increase computational cost and training instability \citep{bishop1994mdns}. Moreover, the number of parameters can grow rapidly in high-dimensional tasks, further requiring efficient covariance parameterizations \citep{he2026hybrid}. 

\subsection{Distributional Regression}\label{sec_distributional}
\begin{figure}[!t]
    \centering
    \includegraphics[width=\linewidth]{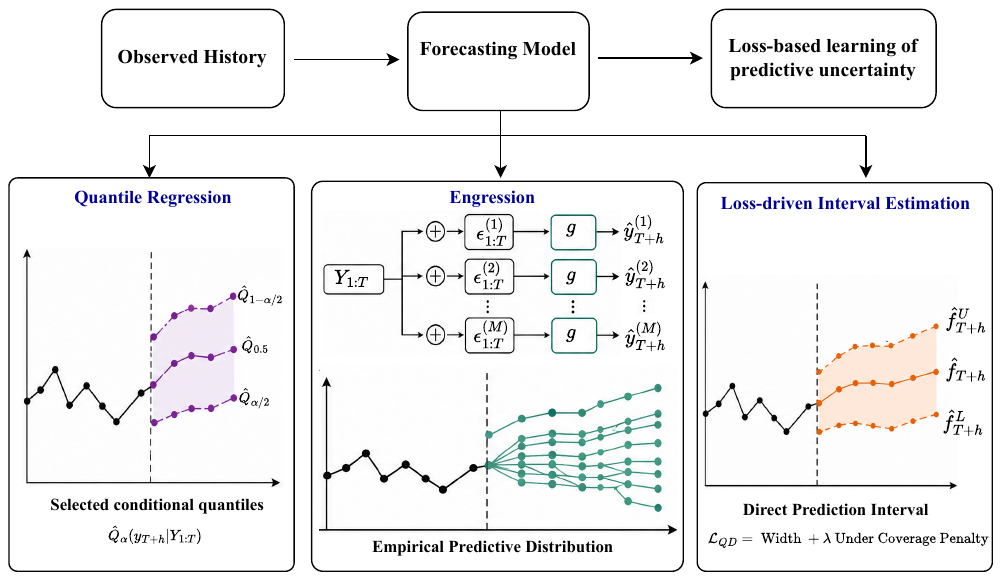}
    \caption{Overview of distributional regression approaches, including quantile regression, engression, and loss-driven interval estimation.}
    \label{fig:Distributional_Regression_Overview}
\end{figure}
Distributional regression methods learn predictive uncertainty through the training objective, rather than by specifying a parametric predictive distribution as in Section~\ref{sec_parametric} \citep{kneib2023distributional}. 
The forecasting model is trained to predict distributional quantities of interest, such as quantiles, intervals, or entire predictive distributions, making uncertainty estimation an integral part of the learning process. Fig.~\ref{fig:Distributional_Regression_Overview} illustrates the three representative approaches reviewed in this section. 
Quantile regression (Section~\ref{sec_distributional_quantile}) estimates conditional quantiles by minimizing the pinball loss. Engression (Section~\ref{sec_distributional_engression}) learns the predictive distribution by optimizing the energy score under a pre-additive noise model. 
Loss-driven interval estimation (Section~\ref{sec_loss_driven}) optimizes interval quality without explicitly estimating densities. These methods differ in the predictive quantities they recover and the associated modeling complexity, as summarized in Table~\ref{tab:dist_reg_comparison}.

\begin{table}[!ht]
\caption{Comparison of distributional regression methods.
Check marks (\ding{51}) denote recoverability of the full predictive distribution; crosses (\ding{55}) denote otherwise.}
\begin{adjustbox}{width=\textwidth}
   \centering
    \begin{tabular}{lccccc} \toprule
   \textbf{Method} & \textbf{Predictive Representation} & \textbf{Full Distribution} & \textbf{Primary Limitation} & \textbf{Computational Cost} \\\midrule

   Quantile Regression & Quantiles & \ding{55}\textsuperscript{$\dagger$} & Quantile Crossing & Low-Moderate \\

   Engression & Forecast Ensemble & \ding{51}\textsuperscript{$\ddagger$} & Noise Specification & Low-Moderate\textsuperscript{$\S$} \\


   Loss-Driven Interval Estimation & Prediction Interval & \ding{55} & Interval Optimization & Low \\
    \bottomrule
    \end{tabular}
    \label{tab:dist_reg_comparison}
    \end{adjustbox}
\footnotesize{\textsuperscript{$\dagger$}Only a finite set of quantiles is estimated; recovering the full density requires additional assumptions.\\
\textsuperscript{$\ddagger$}Recovered empirically through forward simulation, as an ensemble of sampled trajectories.\\
\textsuperscript{$\S$}Training cost scales quadratically in the number of noise realizations $M$ per instance, but $M$ is typically small.\\
}
\end{table}

\subsubsection{Quantile Regression}\label{sec_distributional_quantile}
Quantile regression (QR) estimates selected quantiles of the predictive distribution without requiring the specification of a parametric distributional family \citep{koenker1978regression}, offering a flexible alternative to the approaches discussed in Section~\ref{sec_parametric}. 
The conditional quantile for the future states at level $\alpha\in(0,1)$ is estimated by minimizing the pinball loss over the training set \citep{ingo2011pinball}. 
Estimating multiple quantiles enables the construction of prediction intervals without specifying a parametric predictive distribution, making QR suitable when the underlying distribution is unknown or difficult to model \citep{koenker1978regression,koenker2005quantile}. QR has been widely adopted in probabilistic forecasting, particularly in energy applications. 
\cite{wan2017quantile} proposed direct quantile regression for wind power forecasting, 
improving both calibration and sharpness over kernel density estimation. Quantile regression neural networks integrate QR with deep learning, replacing the linear predictor with a neural network to capture complex nonlinear relationships \citep{cannon2017qrnn}. At the regional scale, \cite{yu2019quantile} combined convolutional and recurrent neural networks with quantile loss to capture spatial and temporal dependencies in aggregated wind power generation. For large spatiotemporal systems, the high-dimensionality of the predictor space presents an additional challenge. \cite{agoua2019photovoltaic} addressed this using an $\ell_1$-penalized quantile regression model that performs variable selection while estimating conditional quantiles, enabling scalable probabilistic forecasting for photovoltaic power generation across hundreds of spatially distributed sites. 

A persistent challenge in QR is quantile crossing, where independently estimated quantiles violate their natural ordering. Existing solutions include constrained optimization, post-processing methods \citep{chernozhukov2010rearranging}, and architectures that jointly estimate multiple quantiles while enforcing monotonicity \citep{moon2021learning}. \cite{fakoor2023flexible} proposed an alternative approach that combines a differentiable crossing penalty during training with post-hoc isotonization operators such as sorting and the min-max sweep. They also proposed deep quantile aggregation, where a neural network learns input-dependent weights to combine multiple base quantile models. 
To forecast multiple quantiles over extended time horizons, \cite{wen2018mqrnn} combined a recurrent encoder with a shared decoder for multi-horizon forecasting. At extreme quantile levels where observations are scarce, \cite{pasche2024eqrn} proposed extreme quantile regression neural networks (EQRN), which combine QR with extreme value theory (EVT). EQRN combines a recurrent network for quantile estimation with RNN-parameterized generalized Pareto distributions to model conditional tails for flood risk forecasting. 
However, QR estimates only a finite set of quantiles rather than the full predictive density \citep{koenker2005quantile}. Recovering the full distribution therefore requires estimating many quantiles or introducing additional assumptions, motivating more flexible probabilistic forecasting frameworks.

\subsubsection{Engression}\label{sec_distributional_engression}
Engression, introduced by \cite{shen2025engression}, is a neural network-based distributional regression framework that learns the conditional distribution of the target variable by optimizing the energy score over a pre-additive noise model. Engression injects noise into the historical observations before passing them through the network: $y_{T+h} = g(\mathbf{Y}_{1:T} + \epsilon_{1:T}),$ rather than adding noise after the nonlinear transformation. This allows the model to generate stochastic forecast trajectories from perturbed input sequences. For a time step $t$, given a ground truth target sequence $y_{t+1:t+H}$ of length $H$ and the corresponding in-sample predictions $\{\widehat{y}_{t+1:t+H}^{(m)}\}_{m=1}^M$ for $M$ independent noise realizations from a predefined distribution, training proceeds by minimizing the empirical energy score loss:
\begin{equation}\label{eq:energy_score_loss}
\mathcal{L}_{\text{ES}} = \frac{1}{|\mathcal{D}|}\sum_{\mathcal{D}}\left[\frac{1}{M}\sum_{m=1}^{M}\left\|y_{t+1:t+H} - \widehat{y}_{t+1:t+H}^{(m)}\right\| - \frac{1}{2M(M-1)}\sum_{m=1}^{M}\sum_{k=1\neq m}^{M}\left\|\widehat{y}_{t+1:t+H}^{(m)} - \widehat{y}_{t+1:t+H}^{(k)}\right\|\right],
\end{equation}
where $|\mathcal{D}|$ denotes the size of the training dataset. The first term penalizes deviation from the ground truth and the second prevents mode collapse. Since the energy score is a strictly proper scoring rule \citep{gneiting2007proper}, minimizing its negative formulation (Eq. \ref{eq:energy_score_loss}) recovers the true conditional distribution. Distinct plausible forecast trajectories can then be sampled from the trained model via forward simulation using different noise samples, resulting in an empirical approximation of the predictive distribution. 
This characteristic has motivated its application to several forecasting problems. For instance, \cite{kraft2026engression} integrated engression with LSTM networks for rainfall runoff forecasting.
For multivariate and spatiotemporal epidemiological forecasting, \cite{pathak2026engression} proposed the multivariate engression network (MVEN), the graph convolutional engression network (GCEN), and the spatiotemporal engression network (STEN), extending engression to jointly capture temporal and spatial dependencies. More recently, \cite{pathak2026entransformer} introduced Enformer and graph Enformer (GEnformer), which leverage Transformer architectures to model long-range temporal dependencies and cross series interactions while improving computational efficiency. 
The training procedure incurs a quadratic computational complexity in the number of generated samples $M$ \citep{pathak2026entransformer}, although the models are typically lightweight for small values of $M$.

\subsubsection{Loss-Driven Interval Estimation}\label{sec_loss_driven}
Loss-driven interval estimation encodes the desired properties of the predictive distribution into the training objective. These approaches treat prediction interval estimation as an empirical risk minimization (ERM) problem where the loss function balances interval coverage and width without requiring explicit quantile estimation. The resulting models can be optimized with gradient-based methods, making them particularly suitable for neural forecasting architectures. The quality driven (QD) loss introduced by \cite{pearce2018high} represents one such approach to optimizing prediction intervals. The objective combines interval width minimization with a penalty for insufficient coverage:
$$\mathcal{L}_{\text{QD}} = \frac{\sum_{t=1}^{T}\left(\widehat{f}_{t}^{U} - \widehat{f}_{t}^{L}\right)k_t}
{\sum_{t=1}^{T}k_t} + \lambda \frac{T}{\alpha (1- \alpha)} \max\left(0,(1-\alpha)-\frac{\sum_{t=1}^{T}k_t}{T}\right)^2,
$$
where $\widehat{f}_{t}^L$ and $\widehat{f}_{t}^U$ denote the estimated lower and upper forecasting bounds at time index $t$, respectively, $k_t$ is a differentiable approximation of whether the observation lies inside the estimated interval, and $\lambda$ balances the trade-off between interval width and coverage. 
$k_t$ is approximated using sigmoid functions, enabling the objective to be optimized through backpropagation. 

Several subsequent methods extend this coverage-width optimization approach. For instance, DeepPIPE \citep{wang2020DeepPIPE} adapts the QD formulation to multi-horizon settings by introducing a unified loss function that jointly penalizes central point deviations and interval inefficiencies.
The hybrid objective balances a weighted Mean Absolute Error (MAE) for point accuracy with a quadratic penalty that is triggered during undercoverage, while minimizing the mean prediction interval width. 
This framework can be incorporated into a wide range of forecasting models, making it suitable for temporal and spatiotemporal applications involving long forecasting horizons \citep{wang2020DeepPIPE}. Despite the distribution-free nature and the ability of these approaches to yield adaptive interval widths, the QD loss and DeepPIPE frameworks introduce additional tuning hyperparameters and complicate gradient-based optimization due to their non-convex objectives. The Tube Loss \citep{anand2026tube} addresses this limitation by encoding the target coverage through a piecewise quadratic loss function that is differentiable almost everywhere. For a target confidence level $1-\alpha$, the Tube Loss is formulated in terms of the lower and upper prediction errors, defined respectively as $l_{t} = y_t - \widehat{f}_{t}^L$ and $u_{t} = y_t - \widehat{f}_{t}^U$:
\[
\mathcal{L}_{1-\alpha}^r(l_{t}, u_{t}) = 
\begin{cases} 
(1-\alpha)u_{t}, &  u_{t} > 0, \\ 
-\alpha u_{t}, &  u_{t} \le 0, \; l_{t} \ge 0, \text{ and } r\;u_{t} + (1-r)\;l_{t} \ge 0, \\ 
\alpha l_{t}, &  u_{t} \le 0, \; l_{t} \ge 0, \text{ and } r\;u_{t} + (1-r)\;l_{t} < 0, \\ 
-(1-\alpha)\;l_{t}, & l_{t} < 0,
\end{cases},
\]
where $r \in (0,1)$ is a tunable parameter that controls the positioning of the interval along the response trajectory.
Optimization via ERM ensures that the resulting predictive intervals asymptotically achieve the target coverage level. For forecasting applications, the Tube Loss has been integrated with LSTM for probabilistic forecasting, achieving superior performance, competitive interval sharpness and reduced training cost compared to quantile-based and QD-based approaches across several datasets \citep{anand2026tube}.

\subsection{Generative Models}\label{sec_generative}
Generative models represent predictive uncertainty by learning a stochastic mechanism for the joint distribution $p(y_{T+1:T+H}\mid \mathbf{Y}_{1:T})$, 
from which future trajectories can be generated \citep{bishop2006prml}. This differs from the approaches discussed in Section~\ref{sec_parametric}, where uncertainty is represented through an explicit parametric form for the predictive density, and Section~\ref{sec_distributional}, where models optimize objectives associated with specific distributional functionals, such as quantiles. Depending on the model, this distribution may admit tractable likelihood evaluation or may be represented empirically through generated samples \citep{goodfellow2014gan, rezende2015vnf, ho2020diffusion}. 

\begin{figure}[!t]
    \centering
    \includegraphics[width=\linewidth]{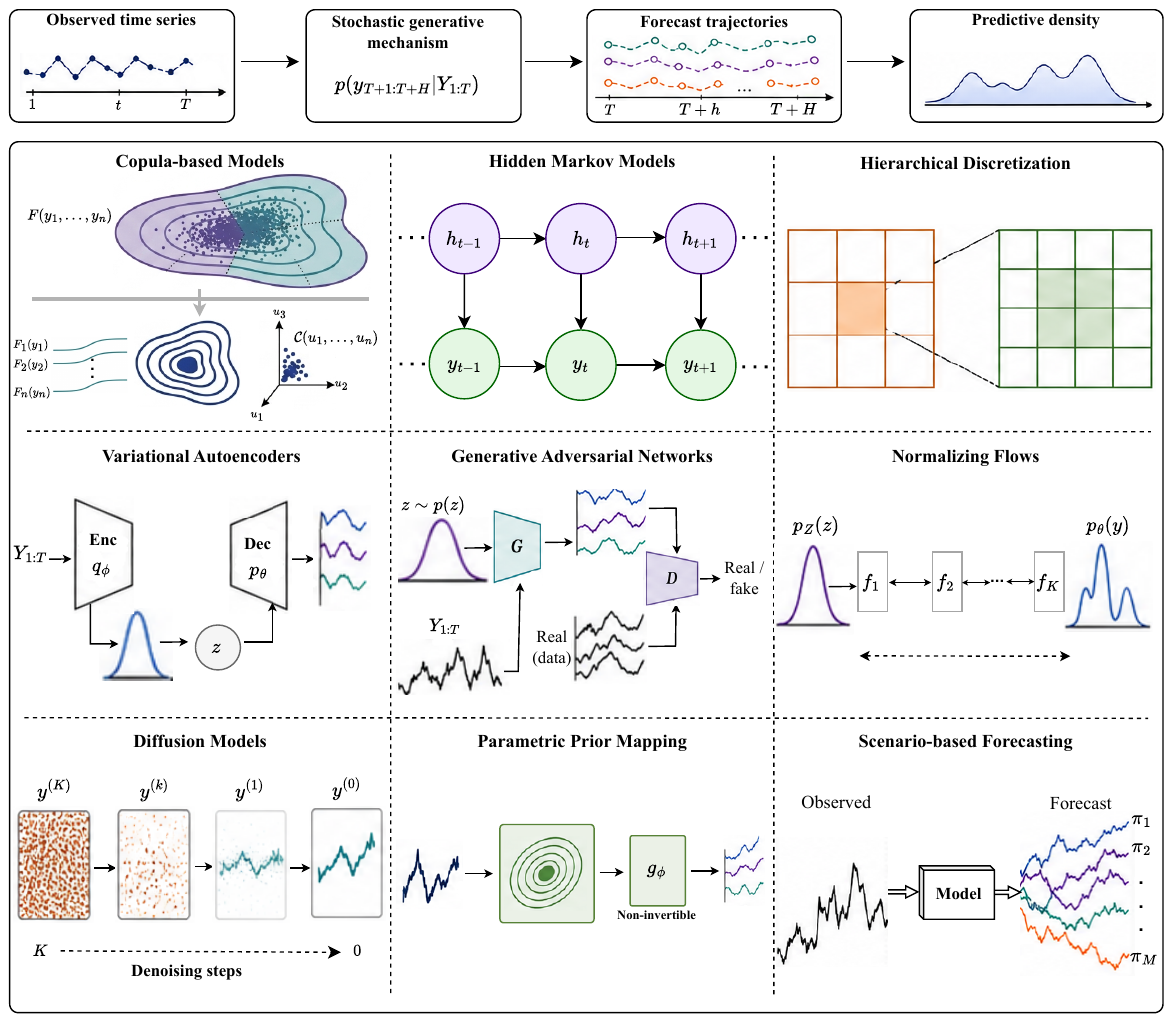}
    \caption{Overview of generative models for probabilistic forecasting.}
    \label{fig_generative_forecasting}
\end{figure}

We organize this section into nine families. Copula-based models (Section~\ref{sec_generative_copula}) separate the joint predictive distribution into marginal distributions and an explicit dependence structure. Hidden Markov models (Section~\ref{sec_generative_hmm}) generate observations conditional on an evolving sequence of discrete latent states. Hierarchical discretization (Section~\ref{sec_distributional_discretization}) represents the predictive distribution through a coarse-to-fine partitioning of the target space. Variational autoencoders (Section~\ref{sec_generative_vae}) learn continuous latent representations from which forecast samples are generated, whereas generative adversarial networks (Section~\ref{sec_generative_gan}) learn an implicit sampling mechanism. Normalizing flows (Section~\ref{sec_generative_nf}) transform a simple base density through a sequence of invertible mappings, allowing complex predictive distributions to be modeled with exact likelihood evaluation. Diffusion models (Section~\ref{sec_generative_diffusion}) generate samples by reversing a progressive noising process. Parametric prior mapping (Section~\ref{sec_generative_prior}) combines a learned latent prior with a non-invertible generative mapping. Scenario-based methods (Section~\ref{sec_generative_scenario}) produce a finite set of future trajectories with explicit probabilities in a single forward pass. These families differ in how they represent the joint predictive distribution, whether exact likelihood evaluation is tractable, and the cost of sampling at inference time, as summarized in Table~\ref{tab:generative_comparison}. Figure~\ref{fig_generative_forecasting} provides a visual overview of the nine families. 

The distinction between sampling-based uncertainty estimation and generative modeling is important for two boundary cases in this taxonomy. 
Engression (Section~\ref{sec_distributional_engression}) represents a boundary case, as it generates forecast samples by passing noise-perturbed inputs through a learned nonlinear mapping. However, it is classified as a distributional regression framework because it models the conditional distribution through a regression framework optimized with the energy score, rather than learning a latent generative process for the data distribution. Although its uncertainty is recovered from sampled trajectories, its formulation and theoretical properties align with distributional regression. Copula-based models (Section~\ref{sec_generative_copula}) represent another boundary case. According to Sklar's theorem \citep{sklar1959fonctions}, a copula combined with specified marginal distributions defines a multivariate joint distribution, which could motivate their classification as parametric models. However, the primary role of copulas in probabilistic forecasting is to construct the joint dependence structure among multiple correlated series, rather than to specify the marginal uncertainty of an individual target. This distinction is particularly relevant for modern neural copula models, which learn flexible dependence structures beyond predefined copula families while preserving the objective of modeling complex multivariate uncertainty. Therefore, copula-based models are grouped with generative approaches, where the focus is on constructing joint predictive distributions.

\begin{table}[!t]
\caption{Comparison of generative models for probabilistic forecasting.
Check marks (\ding{51}) denote availability of exact density evaluation; crosses (\ding{55}) denote otherwise. Abbreviation: Exact L. = Exact Likelihood.}
\begin{adjustbox}{width=\textwidth}
\centering
\begin{tabular}{lccccc}
\toprule
\textbf{Method} &
\textbf{Generative Mechanism} &
\textbf{Exact L.} &
\textbf{Sampling Procedure} &
\textbf{Inference Cost} &
\textbf{Key Limitation} \\
\midrule

Copula-based Models & Dependence Modeling & \ding{51}\textsuperscript{$\dagger$} & Simulation-based & Low-Moderate & High-dimensional dependence structure \\

Hidden Markov Models & Latent State Modeling & \ding{51} & Simulation-based & Low-Moderate\textsuperscript{$\bullet$} & Limited capacity by discrete latent states\\

Hierarchical Discretization & Discrete Target Representation & \xmark & Autoregressive & Moderate-High & Performance issues with limited data\\

Variational Autoencoders & Latent Variable Modeling
 & \ding{55} & Single Forward Pass & Low & Mode collapse \& limited generation quality\\

Generative Adversarial Networks & Adversarial Learning & \ding{55} & Single Forward Pass & Low & Mode collapse \& training instability\\

Normalizing Flows & Invertible Transformations & \ding{51} & Single Forward Pass & Low-Moderate & Invertibility \& expensive computation \\

Diffusion Models & Iterative Denoising & \ding{55} & Iterative Denoising & Moderate-High\textsuperscript{$\P$} & Slow sampling \& high inference cost\\

Parametric Prior Mapping & Push-forward Mapping & \ding{55} & Single Forward Pass  & Low & Bandwidth-sensitive density estimation\\

Scenario-based Forecasting & Discrete Scenario Generation & \ding{55} & Single Forward Pass & Low & Fixed number of scenarios; no density outside support\\

\bottomrule
\end{tabular}
\label{tab:generative_comparison}
\end{adjustbox}

\footnotesize{
\textsuperscript{$\dagger$}Exact evaluation is restricted to copula models with tractable marginals, neural copulas require approximate inference.\\
\textsuperscript{$\bullet$}Cost increases with the sequence length and number of latent states.\\
\textsuperscript{$\P$}Dependent on diffusion steps; latent diffusion reduces this via space compression (Section~\ref{sec_diffusion_ldm}).\\
}
\end{table}

\subsubsection{Copula-Based Models}\label{sec_generative_copula}
Copula-based methods separate the modeling of marginal distributions from the dependence structure among variables. 
According to Sklar's theorem \citep{sklar1959fonctions}, any multivariate distribution can be decomposed into its univariate marginals and a copula capturing their dependence:
$$F(y_1, \dots, y_N) = \mathcal{C}\big(F_1(y_1), \dots, F_N(y_N)\big),$$
where $F_i$ are the marginal cumulative distribution functions and $\mathcal{C}$ is the copula capturing their dependence \citep{nelsen2006copula}. This decomposition provides a flexible framework that allows marginals to be modeled independently while preserving the joint dependencies. Therefore, copulas have become a prominent approach for probabilistic forecasting. 
However, explicitly modeling dependencies becomes challenging as the number of variables grows, since the number of pairwise relationships scales quadratically with dimensionality. \cite{salinas2019gpcopula} addressed this by combining an autoregressive recurrent network for modeling temporal dynamics with a low-rank Gaussian copula process, in which a low-rank-plus-diagonal covariance reduces the parameter count.
In a different direction, \cite{sun2024copula} developed copula conformal prediction for time series (CopulaCPTS),
which constructs marginal conformal predictive distributions at each horizon step and subsequently fits an empirical copula to capture dependencies among future prediction errors. 

The ability of copulas to represent complex dependence structures has also motivated their integration into spatiotemporal forecasting. Among the various copula constructions, vine copulas \citep{joe1996families} provide a flexible framework for modeling high-dimensional dependence by decomposing a multivariate distribution into a hierarchy of bivariate copulas. 
For instance, \cite{erhardt2015rvine} extended regular-vine (R-vine) copulas \citep{joe2010vine} to spatiotemporal forecasting by parameterizing pairwise dependencies as functions of geographical distance, yielding superior probabilistic temperature forecasts across German weather stations compared to Gaussian spatial models. 
However, scaling R-vine copulas to large spatiotemporal systems remains challenging due to the increasing complexity of the vine structure.
\cite{zhang2026joint} addressed this limitation by combining R-vine copulas with Kolmogorov-Arnold Networks (KANs) \citep{liu2025kan} for regional wind power forecasting.
Copula models have also been adopted for incomplete spatiotemporal observations. \cite{an2025stmpnet} employed copulas to estimate the joint distribution of missing traffic measurements, enabling forecasting and data imputation within a unified probabilistic framework. 

Another challenge in copula-based forecasting is jointly improving point prediction accuracy and uncertainty calibration. \cite{kim2026tsconet} proposed a two-stage copula network (TSCoNet), which combines a hybrid convolutional neural network (CNN)-LSTM predictor with a Gaussian copula to jointly model point forecasts and multivariate uncertainty. 
As forecasting problems become increasingly high-dimensional and irregular, recent work has explored neural parameterizations of copulas to improve scalability and flexibility. \cite{ashok2024tactis} introduced an improved version of Transformer-attentional copulas for time series (TACTiS-2), which addresses the difficulty of learning valid neural copulas by separating the estimation of marginal distributions and dependence structures. TACTiS-2 reduces the number of distributional parameters
in TACTiS \citep{drouin22tactis} 
while enabling scalable probabilistic forecasting for high-dimensional and irregularly sampled time series. Another active direction focuses on dynamic copulas whose dependence structure evolves over time. \cite{zito2025dynamic} proposed a Bayesian dynamic copula model that jointly learns nonparametric marginal distributions and time-varying dependence through a Gaussian copula with a dynamic factor structure, providing fully Bayesian probabilistic forecasts for multivariate time series. Similarly, \cite{li2026uncertainty} combined LSTM and generalized autoregressive conditional heteroskedasticity difference-residual modeling with R-vine copulas to characterize the evolving dependence between wind and solar power generation, enabling joint uncertainty quantification for renewable energy forecasting.


\subsubsection{Hidden Markov Models}\label{sec_generative_hmm}
Hidden Markov models (HMMs) are latent variable models that represent a time series as observations generated from a sequence of unobserved states that evolve according to a Markov process \citep{zucchini2016hidden}. For the observed history $\mathbf{Y}_{1:T}=\{y_t\}_{t=1}^T$ and a corresponding sequence of hidden states $\mathbf{H}_{1:T} = \{h_t\}_{t=1}^T$, an HMM is defined by a state transition model $p(h_t \mid h_{t-1})$ and an observation model $p(y_t \mid h_t)$. The joint distribution factorizes as:
$$p(\mathbf{Y}_{1:T}, \mathbf{H}_{1:T}) = p(h_1) \prod_{t=2}^{T} p(h_t \mid h_{t-1}) \prod_{t=1}^{T} p(y_t \mid h_t).$$
The model parameters are typically estimated via EM through the Baum-Welch procedure \citep{baum1970maximization} and forecasts are produced by propagating uncertainty over future state transitions.
%

Early spatiotemporal applications, such as \cite{zucchini19911hidden}, modeled spatial dependencies by assuming observations across locations were conditionally independent given a shared hidden (weather) regime to avoid the computational burden of modeling the full joint distribution.
Later works relaxed the assumptions of fixed transition dynamics and a predefined number of states of standard HMMs. For instance, non-homogeneous HMMs allow transition probabilities to depend on external covariates, enabling the model to adapt to changing environmental conditions and seasonal effects \citep{tezcan2025training}. The infinite hidden Markov model (InfHMM) further removes the need to specify the number of states in advance 
by placing nonparametric Bayesian priors over state transitions \citep{frimane2022infhmm}. 
More flexible variants relax the model's conditioning structure. Input-output HMMs (IOHMMs) \citep{bengio1994input} incorporate external inputs into both the transition and observation models. For example, they can combine environmental covariates and historical activity data with latent behavioral states to forecast trajectories while retaining interpretability through the learned transitions \citep{liu2026iohmm}. 
Multi-observation HMMs (MOHMMs) instead allow several correlated variables to be generated from the same hidden state. This approach is useful in solar forecasting, where multiple meteorological variables jointly reflect a latent weather condition \citep{zhang2025mohmm}. 
Each of these extensions relaxes a specific assumption of standard HMMs, while retaining the same core representation of discrete latent regimes with state-specific observation models. 

\subsubsection{Hierarchical Discretization}\label{sec_distributional_discretization}
Hierarchical discretization 
models the predictive distribution over a discretized representation of the target variable. The core idea is to partition the support of the target into a finite set of bins and predict the probability of falling into each bin using a categorical distribution \citep{bergsma2022c2far,liu2024deep}. 
This approach places no shape constraints on the resulting density and naturally accommodates multimodality, discrete targets, and mixed data types. A basic implementation uses uniform binning over the range of the target variable, but this introduces a trade-off between resolution and efficiency, since higher precision requires a rapidly growing number of bins. \cite{bergsma2022c2far} addressed this through the coarse-to-fine autoregressive (C2FAR) framework, which introduces a generative hierarchical discretization strategy. A scalar target value $y$ is represented as a vector of $B$ indices, $(y^1, \ldots, y^B)$, where $y^1$ identifies a coarse bin and each subsequent $y^i$ refines the estimate within the bin identified by $y^{i-1}$. 
In practice, C2FAR uses an LSTM to capture temporal dependencies while the hierarchical categorical outputs characterize predictive uncertainty at each level of the hierarchy \citep{bergsma2022c2far}. To handle unbounded support, a parametric tail distribution, such as the Pareto distribution, is appended to the finest level. A conceptually similar approach is the hierarchical discrete Transformer (HDT) \citep{shibo2025hdt}, which reformulates multivariate forecasting as a token generation task by encoding time series into discrete representations via $\ell^2$-normalized vector quantization. This hierarchical framework first extracts low-level tokens to capture overarching long-term trends, which subsequently condition the generation of high-level tokens to accurately model fine-grained dynamics over extended prediction horizons. The extension of the framework by \cite{bergsma2022c2far} to spatiotemporal settings remains an open direction.
Additionally, the hierarchical factorization introduces a sequential dependency across precision levels that increases inference cost relative to methods that produce distributional outputs in a single forward pass.

\subsubsection{Variational Autoencoders}\label{sec_generative_vae}
Variational autoencoders (VAEs), introduced by \cite{kingma2014autoencoding} and \cite{rezende2014stochastic}, are latent variable generative models that learn a probabilistic representation of the data by mapping observations to a latent distribution \citep{kingma2019vae}.
Given the observations $\mathbf{Y}_{1:T}$, an encoder approximates the posterior $q_\phi(z \mid \mathbf{Y}_{1:T})$, from which a latent variable $z$ is sampled and decoded to generate the future trajectory. Model training maximizes the ELBO \citep{kingma2014autoencoding},
which balances reconstruction accuracy with regularization of the latent representation toward a prior distribution. Learning a distribution over these latent variables enables probabilistic forecasting by sampling from the learned latent space. 

Early temporal latent variable models extended VAEs with recurrent architectures to capture sequential dynamics. Frameworks such as variational recurrent autoencoders \citep{fabius2015vrae} and variational RNNs \citep{chung2015recurrent} introduced recurrent states to model time-varying latent distributions. 
ProTran \citep{tang2021probabilistic} 
replaces recurrent latent dynamics with self-attention, allowing each latent variable to depend on the entire history rather than only the previous state. 
Later, \cite{koyuncu2024eprotran} proposed an efficient variant, E-ProTran, which removes autoregressive attention and restricts sampling to the final layer, achieving comparable performance with substantially faster inference. Vector autoregression-VAE (VAR-VAE) \citep{leushuis2025varvae} takes a different approach by constraining the latent dynamics with a first-order VAR process, improving robustness to noisy observations and producing more stable probabilistic forecasts.
To stabilize training under data scarcity, the diffusion, denoise, and disentanglement VAE (D$^3$VAE) model \citep{li2022d3vae} combines a bidirectional VAE with a diffusion-based noise-augmentation scheme (see Section~\ref{sec_generative_diffusion}) and a score-matching denoising module. For long-term forecasting, \cite{wu2025kvae} proposed K$^2$VAE, which transforms nonlinear time series into a linear dynamical system using a learnable Koopman operator (KoopmanNet) and refines predictions and uncertainty through a Kalman-like filtering mechanism (KalmanNet) to mitigate error accumulation over extended horizons. 

VAEs have also been extended to spatiotemporal forecasting by integrating GNNs to capture spatial dependencies. Graph-based ensemble forecasting model (Graph-EFM) \citep{oskarsson2024probabilistic} introduces stochastic latent variables on a hierarchical graph representation of the atmosphere, allowing uncertainty to propagate across multiple spatial resolutions while producing calibrated ensemble weather forecasts. \cite{holmberg2026njord} extended this approach to global and regional ocean forecasting with Njord, adapting the hierarchical graph structure to irregular ocean geometries and incorporating physical constraints to model ocean and sea ice dynamics. However, VAE-based forecasting models remain susceptible to posterior collapse and the smoothing effects associated with the ELBO objective, discussed in Section~\ref{sec_approximate_vi}, which can limit their ability to capture complex predictive distributions \citep{macKay2003information,maldonado2026analysis}. These limitations have motivated alternative generative approaches, including GANs \citep{goodfellow2014gan} and diffusion models \citep{ho2020diffusion}, which rely on adversarial learning and iterative denoising, respectively. 

\subsubsection{Generative Adversarial Networks}\label{sec_generative_gan}
Generative adversarial networks (GANs), introduced by \cite{goodfellow2014gan}, learn a data distribution through an adversarial game between two neural networks. 
Training is formulated as a minimax objective, where a generator $\mathbb{G}$ maps a latent variable $z \sim p(z)$ to a synthetic sample, while a discriminator $\mathbb{D}$ distinguishes these generated samples from real observations. GANs learn to generate samples without defining an explicit predictive distribution, making them highly suitable for probabilistic forecasting \citep{gao2022gan}. However, the adversarial objective was originally developed for independent observations and does not account for temporal or spatial dependence, motivating forecasting-specific adaptations \citep{pacchiardi2024gan}. 

Early applications of GANs in forecasting focused on predicting multimodal trajectories. Social GAN \citep{gupta2018social} combined an LSTM generator with a social pooling mechanism to generate multiple plausible pedestrian trajectories, introducing a diversity loss to encourage multimodal predictions.  
Spatiotemporal interaction graph attention GAN (STI-GAN) \citep{huang2021stigan} subsequently replaced the pooling mechanism with graph attention, allowing the generator to model spatial interactions jointly with temporal dynamics and improving trajectory prediction in crowded scenes. GANs were also extended to a broader range of spatiotemporal problems. For example, ImaGINator \citep{wang2020imagenet} incorporated a spatiotemporal fusion mechanism through encoder-decoder skip connections to preserve appearance while generating realistic video dynamics. However, adversarial training remains a major limitation of GAN-based frameworks because maintaining training stability and calibrated uncertainty is challenging in high-dimensional settings \citep{gao2022gan}. \cite{pacchiardi2024gan} addressed this limitation by replacing the adversarial objective with the optimization of proper scoring rules, such as the energy and kernel scores \citep{gneiting2007proper}, over sequential forecasts. The proposed framework accounts for temporal dependence through sequential conditioning and was later extended to spatiotemporal forecasting using patched scoring rules that evaluate localized spatial regions. 

\subsubsection{Normalizing Flows}\label{sec_generative_nf}
Normalizing flows, introduced by \cite{tabak2013nf} and comprehensively reviewed by \cite{papamakarios2021nf}, are a family of generative models that construct expressive probability distributions by transforming a simple base distribution, $p_\mathcal{Z}(z)$, into a complex data distribution, $p_\theta(y)$, through a sequence of invertible and differentiable transformations. Because the mapping is bijective, it can be evaluated in reverse for exact density estimation. Given an observation sequence $y$, the flow learns the mapping $f_\theta$ that transforms the data back into the latent base space. The exact density of the observed sequence is then obtained via the change-of-variables formula:
$$ p_\theta(y)  = p_\mathcal{Z}(f_\theta(y)) \cdot \left| \det \frac{\partial f_\theta(y)}{\partial y} \right|,$$
where the Jacobian determinant accounts for the change in volume induced by the transformation \citep{papamakarios2021nf}. This enables exact likelihood evaluation and efficient sampling, making normalizing flows a powerful alternative to GANs and VAEs.
In the context of probabilistic forecasting, this framework is typically conditioned on the observed history $\mathbf{Y}_{1:T}$ to model the highly flexible conditional predictive density $p(y_{T+h} \mid \mathbf{Y}_{1:T})$ of future states. 
As the model is autoregressive, the predictive distribution factorizes over the forecast horizon, with each factor obtained by applying the change-of-variables at a single time step \citep{rasul2021multivariate}.

In practice, conditional normalizing flows are commonly implemented using affine coupling layers, such as real-valued non-volume preserving 
\citep{dinh2017density}, whose triangular Jacobian enables efficient likelihood computation and exact inversion. Early forecasting models combined conditional normalizing flows with autoregressive sequence models. Transformer-MAF \citep{rasul2021multivariate} integrates a masked autoregressive flow (MAF) \citep{papamakarios2017maf} with a Transformer decoder, using masked self-attention to condition each flow transformation on the complete observed history. 
Subsequent works extended flow-based models to spatiotemporal forecasting by incorporating explicit spatial representations into the conditioning mechanism. Spatial-temporal graph conditionalized normalizing flow (STGNF) \citep{an2024graphnf} integrates GNNs with conditional normalizing flows, allowing local and global spatiotemporal dependencies learned from traffic networks to parameterize the flow transformations and yield probabilistic traffic forecasts. MotionFlow \citep{Zand2023stsp} instead employs masked convolutions to model spatial autoregressive dependencies for probabilistic trajectory prediction. Recent work has expanded normalizing flows to non-Euclidean domains. For land deformation forecasting, DyLand \citep{xu2024dyland} combines neural ordinary differential equations with topology-preserving mappings to model evolving spatial manifolds. 
Normalizing flows have also been adapted to extreme value analysis. For instance, \cite{demonte2025generative} introduced hyperspherical flow transformations to model multivariate extremes, mapping the data onto spherical manifolds to estimate the directional distributions of rare events. Irregularly sampled multivariate time series present another challenge, as conventional coupling layer architectures assume a fixed-dimensional input. ProFITi \citep{yalavarthi2025ProFITi} addresses this limitation by replacing fixed-dimensional coupling layers with an invertible attention mechanism conditioned on graph-based representations of observed and queried time-channel pairs. Hence, the flow can model joint distributions over a variable number of targets. 
However, the computational cost of invertible transformations and Jacobian evaluations remains a fundamental challenge for applying normalizing flows to high-dimensional forecasting tasks \citep{papamakarios2021nf}.

\subsubsection{Diffusion Models}\label{sec_generative_diffusion}
\begin{figure}[!t]
    \centering
    \includegraphics[width=\linewidth]{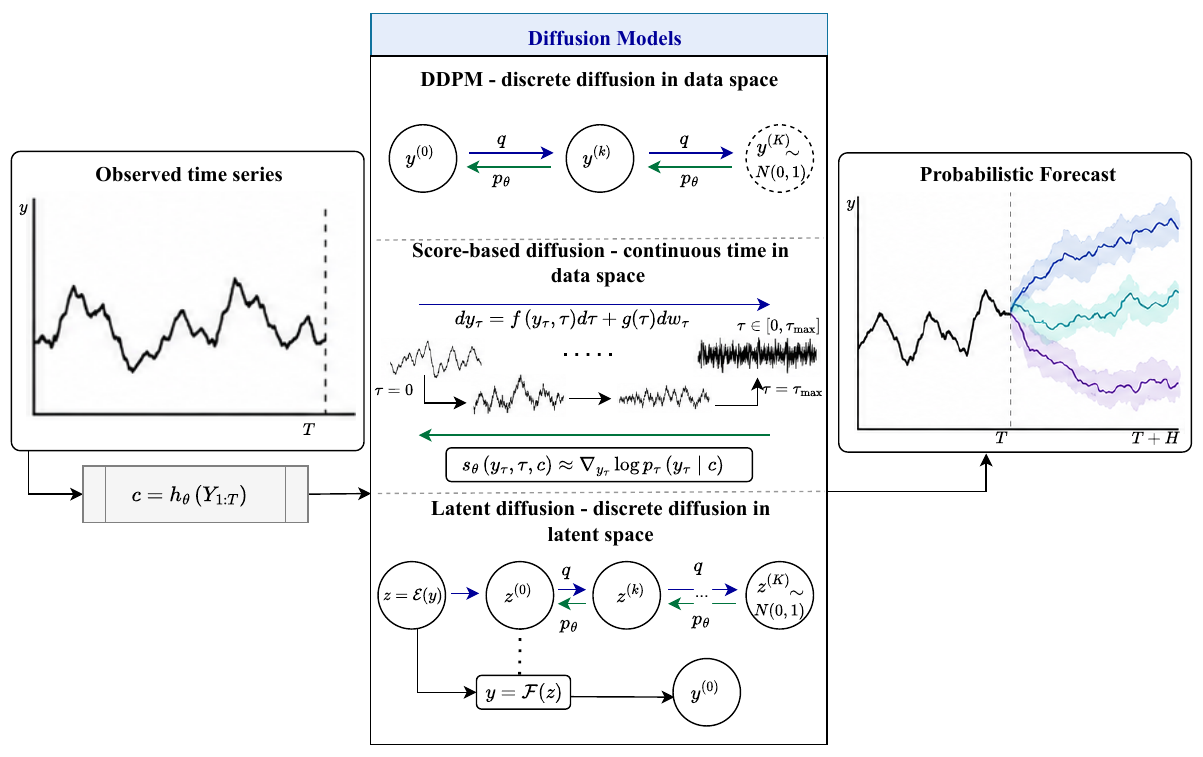}
    \caption{Overview of diffusion models, including DDPM, score-based, and latent diffusion models.}
    \label{fig:diffusion}
\end{figure}
Diffusion models are latent variable generative models originally introduced from a thermodynamic perspective \citep{dickstein2015deep} and later formalized for deep generative modeling \citep{ho2020diffusion}. They generate samples by reversing a predefined diffusion process that progressively perturbs the target with Gaussian noise.
In time series forecasting, this process is conditioned on historical observations to model the distribution of future trajectories, enabling uncertainty quantification by generating multiple plausible forecasts \citep{rasul2021timegrad}. Readers are referred to \cite{yang2026diffusion} for a comprehensive survey on diffusion models for time series and spatiotemporal forecasting. Diffusion models appeal for their distributional flexibility and the stability of their training objective compared to adversarial approaches. However, their iterative sampling procedure introduces high computational costs, motivating a range of approaches to improve sampling efficiency, conditioning mechanisms, and the representation of temporal uncertainty \citep{rasul2021timegrad, tang2024udm, ye2025nsdiff}.  More recently, these models have been extended to spatiotemporal forecasting by incorporating spatial dependencies into the diffusion process \citep{wen2023diffstg, Chen2024diffusion, cachay2023dyffusion}. We categorize diffusion-based forecasting methods into three groups, as illustrated in Fig.~\ref{fig:diffusion}: denoising diffusion probabilistic models (Section~\ref{sec_diffusion_ddpm}), 
score-based diffusion models (Section~\ref{sec_diffusion_score}), and latent diffusion models (Section~\ref{sec_diffusion_ldm}). Table~\ref{tab:diffusion_comparison} summarizes the key characteristics of these diffusion models.

\begin{table}[!htbp]
\caption{Comparison of diffusion model formulations for probabilistic forecasting. 
}
\begin{adjustbox}{width=\textwidth}
   \centering
    \begin{tabular}{lccc} \toprule
    \textbf{Property} & \textbf{Denoising Diffusion Probabilistic Model} & \textbf{Score-based Diffusion Model} & \textbf{Latent Diffusion Model} \\ \midrule
    
    \textbf{Key Idea} & Discrete Diffusion Process & Continuous-time Diffusion & Latent-Space Compression \\
    
    \textbf{Time Parameterization} & Discrete Steps & Stochastic Differential Equation  & Discrete Steps \\

    \textbf{Training Objective} & Denoising Objective & Denoising Score Matching & Denoising Objective \\
    
    
    \textbf{Generation Space} & Data Space & Data Space & Latent Space \\
    
    \textbf{Sampling Cost} & High & High & Low-Moderate \\\bottomrule
    \end{tabular}
    \label{tab:diffusion_comparison}
    \end{adjustbox}
\end{table}

\paragraph{Denoising Diffusion Probabilistic Models.}\label{sec_diffusion_ddpm} 
Denoising diffusion probabilistic models (DDPMs) represent the foundational class of diffusion-based generative models \citep{ho2020diffusion}. 
They define a forward diffusion process that corrupts the target sample with Gaussian noise over multiple diffusion steps.
A key property of this process is that the noisy sample at any diffusion step can be obtained directly from the original sample,
enabling efficient training by avoiding the sequential simulation of the forward process. 
Instead of predicting reverse transition probabilities directly, DDPMs parameterize the reverse process by training a neural network to estimate the injected noise.
This defines the reverse process used during inference to iteratively transform Gaussian noise into samples from the learned data distribution \citep{ho2020diffusion}. 

The first application of DDPMs to probabilistic forecasting was TimeGrad \citep{rasul2021timegrad}, which conditions the denoising network on an RNN hidden state summarizing historical observations.
TimeGrad models the conditional distribution of each future time step through a diffusion process. 
However, its autoregressive formulation requires a complete diffusion process for every predicted time step, scaling inference cost linearly with both the forecasting horizon and the number of diffusion steps. 
Subsequent work addressed this limitation with conditional diffusion models, by jointly diffusing the entire forecast horizon and incorporating the conditioning information $c$ into the reverse process. $c$ encodes information used to guide the generation process, such as historical observations, graph structure, or auxiliary covariates \citep{yang2026diffusion}. 
TimeDiff \citep{shen2023nonautoregressive} was the first conditional diffusion model designed for forecasting. It constructs the conditioning signal $c$ using future mixup and autoregressive initialization to better capture temporal dependencies during denoising. Furthermore, it departs from conventional DDPMs by predicting the clean sample $y^{(0)}$ directly instead of the injected noise, to account for the fact that real-world time series noise is more irregular than standard Gaussian noise \citep{shen2023nonautoregressive}.
Transformer-modulated diffusion model (TMDM) \citep{li2024transformermodulated} simplifies distribution learning by using a Transformer to estimate the conditional mean forecast, thus restricting the forward and reverse diffusion processes to modeling only the residual distribution. DiffSTG \citep{wen2023diffstg} extends conditional diffusion to spatiotemporal forecasting 
by conditioning on both a masked spatiotemporal sequence and spatial structure. The masked sequence enables the entire forecast horizon to be generated through a single reverse diffusion process, while the denoising network combines a U-Net architecture with graph convolutions \citep{ronneberger2015unet} to model temporal and spatial dependencies jointly.

A limitation of early DDPM forecasters is the assumption of stationary noise variance, which fails to capture the changing uncertainty patterns common in real-world time series. To address this limitation, nonstationary diffusion (NsDiff) \citep{ye2025nsdiff} replaces the conventional additive noise formulation with a location scale noise model, allowing the diffusion process to capture time-varying uncertainty. A second direction modifies the denoising process itself. 
Multi-granularity time series diffusion (MG-TSD) \citep{fan2024mgtsd} introduces intermediate supervision across multiple temporal granularities during the denoising phase, encouraging the model to preserve long-term trends while progressively refining local dynamics. 
Extending DDPMs to spatiotemporal forecasting, dynamic spatial-temporal (DST)-DDPM \citep{Chen2024diffusion} combines a dynamic graph encoder that learns time-varying adjacency matrices with a Transformer-based denoising network for air quality forecasting.
Alternatively, spectral diffusion for spatiotemporal graph (SpecSTG) \citep{lin2025SpecSTG} performs diffusion in the graph spectral domain, where spatial correlations are represented through graph Fourier coefficients and conditioned using a spectral GRU. This formulation reduces the cost of modeling spatial structure while maintaining competitive probabilistic forecasting performance. For trajectory forecasting, UTD-PTP \citep{tang2024udm} adapts DDPMs to pedestrian movement by pairing a Transformer sequence encoder with a denoising diffusion implicit model (DDIM) sampler. This compresses the multi-step reverse process into a few iterations, accelerating inference while preserving sample diversity. 
The iterative reverse process of DDPM models is computationally expensive for long horizons, although accelerated samplers like DDIM offer partial mitigation. This 
motivates alternative diffusion formulations designed for more efficient forecast generation.

\paragraph{Score-Based Diffusion Models.}\label{sec_diffusion_score} Score-based diffusion models formulate the diffusion process in continuous time through a stochastic differential equation (SDE) rather than a discrete sequence of diffusion steps \citep{song2021score}. 
The forward process gradually perturbs the data according to:
$$
dy_\tau = f(y_\tau, \tau) \, d\tau + g(\tau) \, dw_\tau,
$$
where $\tau \in [0, \tau_{\max}]$ denotes the continuous diffusion time, $w_\tau$ is a standard Wiener process, and $f(\cdot, \tau)$ and $g(\tau)$ define the drift and diffusion coefficients, respectively. Sample generation is performed by solving the corresponding reverse-time SDE, which depends on the score function $\nabla_{y_\tau} \log p_\tau(y_\tau)$ estimated through denoising score matching. Compared to DDPMs, the continuous formulation removes the need to specify a discrete diffusion schedule and provides greater flexibility in the design of diffusion trajectories \citep{song2021score, yan2021scoregrad}. ScoreGrad \citep{yan2021scoregrad} first introduced score-based diffusion to probabilistic forecasting by replacing TimeGrad's discrete diffusion process with a continuous SDE formulation. 
The model integrates temporal feature extraction with conditional score matching to reduce sensitivity to diffusion hyperparameters. 

Continuous score-based diffusion has also been extended to spatiotemporal forecasting. ProGen \citep{gong2024progen} integrates GNNs within the SDE framework to capture spatial dependencies for traffic forecasting. GenCast \citep{price2024gencast} applies the same continuous-time formulation, following the parameterization of \cite{karras2022elucidating}, to global weather forecasting. GenCast conditions each diffusion process on the two most recent atmospheric states to model the distribution of the next state. 
Alternative approaches modified the forward diffusion process itself. For instance, dynamics-informed diffusion (DYffusion) \citep{cachay2023dyffusion} defines the forward process through temporal interpolation rather than Gaussian noise injection. The model operates on spatiotemporal dynamical systems, such as spatial grids or meshes, and captures spatial dependencies using CNN and U-Net architectures. A stochastic interpolator learns the forward process, while a forecasting network approximates the reverse process, reducing the number of diffusion steps required for sampling without sacrificing probabilistic forecasting performance.
For forecasting under extreme conditions, \cite{he2025offshore} proposed a score-based conditional diffusion model for offshore wind power prediction during typhoons. The framework employs a mean-reverting SDE alongside deterministic forecast networks to reduce the complexity of the denoising process. However, the need to repeatedly solve the reverse SDE remains a computational limitation of such models.

\paragraph{Latent Diffusion Models.}\label{sec_diffusion_ldm} Latent diffusion models reduce the computational cost by performing the denoising process in a compressed latent space rather than in the original data space \citep{rombach2022high}. The framework first learns an encoder-decoder pair that maps observations into a lower-dimensional representation $z = \mathcal{E}(y)$ and reconstructs the original data through $y=\mathcal{F}(z)$. Diffusion is then applied to the latent representation instead of the original signal.
This reduces the dimensionality of the generative process and enables more efficient sampling for high-dimensional forecasting tasks. \cite{feng2024ldt} introduced latent
diffusion Transformer (LDT) for multivariate time series forecasting. 
LDT uses an adaptive normalization mechanism to improve robustness under distribution shifts and performs non-autoregressive generation in the latent space, reducing the computational burden of long-horizon forecasting. Latent diffusion is also used in trajectory forecasting \citep{yang2026diffusion}. For instance, learning autoencoder diffusion model (LADM) \citep{lv2024ladm} pairs a VAE with a diffusion refiner. In this setup, a generator produces coarse trajectories, which the diffusion model then sharpens within the latent space, simplifying the refinement process and generating higher-quality future movements.
For spatiotemporal forecasting, solar high-resolution adaptive diffusion ensemble forecasting (SHADECast) model \citep{Carpentieri2025SHADECast} applies latent diffusion to high-dimensional solar irradiance fields by compressing spatiotemporal observations into a latent representation. 
Although latent models mitigate the computational burden of high-dimensional forecasting, 
their performance depends on the quality of the learned representation and the ability of the latent space to preserve the original data characteristics.

\subsubsection{Parametric Prior Mapping}\label{sec_generative_prior}
Parametric prior mapping (PPM) \citep{li2026ppm} relaxes the invertibility requirement of normalizing flows while avoiding the iterative sampling procedure of diffusion models. The framework combines a learned context-dependent prior with a non-invertible neural mapping, replacing the fixed latent distributions used in diffusion models and the bijective transformations required by normalizing flows. Given the observed history $\mathbf{Y}_{1:T}$, an encoder estimates the parameters of a multivariate Gaussian latent prior:$$p_\theta(z \mid \mathbf{Y}_{1:T}) = \mathcal{N}\left(\mu_\theta(\mathbf{Y}_{1:T}\right), \Sigma_\theta(\mathbf{Y}_{1:T})),$$where $z$ is the latent vector, and $\mu_\theta(\cdot)$ and $\Sigma_\theta(\cdot)$ represent the learned mean vector and covariance matrix, respectively. The joint predictive distribution over the forecast horizon is then obtained as the push-forward distribution:
$$ q_\phi(y_{T+1:T+H} \mid \mathbf{Y}_{1:T}) = \left(g_\phi\right)_\# p_\theta(z \mid \mathbf{Y}_{1:T}), $$
where $(g_\phi)_\#$ denotes the push-forward measure induced by the neural mapping $g_\phi$.
Forecasts are generated in a single forward pass through $g_\phi$, eliminating the iterative denoising procedure of diffusion models. At the same time, removing the invertibility constraint permits the use of more flexible neural architectures than those employed in normalizing flows. This combination substantially reduces inference time while maintaining competitive probabilistic forecasting performance. Since $g_\phi$ is not invertible, exact likelihood evaluation is intractable. Training, therefore, combines a kernel density estimation-based negative log-likelihood
with a mean squared error term that regularizes the sample mean and stabilizes optimization \citep{li2026ppm}.

\subsubsection{Scenario-Based Probabilistic Forecasting}\label{sec_generative_scenario}
An alternative to sampling-based generative models is to predict a finite set of future trajectories, or scenarios, together with their associated probabilities. In contrast to diffusion and other sampling-based approaches, scenario-based methods produce explicit \{scenario, probability\} pairs in a single forward pass. Thus, such approaches resolve major sampling limitations, specifically the absence of explicit probabilities for trajectories, inadequate coverage of rare events, and high inference cost \citep{dai2026timeprism}. Most existing approaches build on multiple choice learning \citep{guzmanrivera2012mcl}, where a multi-head network is trained under a winner-takes-all (WTA) loss so that only the head closest to the ground truth receives a gradient update, encouraging each head to specialize in a distinct mode of the conditional distribution \citep{rupprecht2017learning}. TimeMCL \citep{cortes2025timemcl} first adapted multiple choice learning to probabilistic time series forecasting by training a multi-head autoregressive network with a WTA loss. However, strict WTA objective hinders training stability by causing hypothesis collapse, where only few heads receive sufficient gradient updates while the rest remain underutilized. Subsequent work has focused on mitigating this limitation. TimePre \citep{jiang2026timepre} introduces stabilized instance normalization to reduce scale disparities between variables, improving optimization stability and preserving hypothesis diversity while retaining the efficiency of lightweight linear forecasting backbones. TimePrism \citep{dai2026timeprism} further simplifies the architecture by decomposing inputs into trend and seasonal components, which are combined to generate a larger set of candidate scenarios. A parallel linear layer then predicts scenario probabilities, while a relaxed WTA objective improves the utilization of all prediction heads. TimePrism outperformed generative baselines across benchmarks at a fraction of the inference cost since its cost remains fixed regardless of the number of requested scenarios.

Overall, the model-intrinsic methods differ primarily in how they represent predictive uncertainty. Bayesian approaches place distributions over unknown quantities, parametric likelihoods specify a predictive law, quantile methods estimate selected conditional functionals, and generative models produce samples from flexible distributions. These choices determine which aspects of uncertainty can be represented and evaluated, but none alone guarantees calibrated forecasts. In particular, representing multivariate and spatiotemporal dependence and establishing calibrated coverage remain distinct problems.

\section{Foundation Models for Probabilistic Forecasting}\label{sec_foundation_models}
\begin{figure}[!t]
    \centering
    \includegraphics[width=\linewidth]{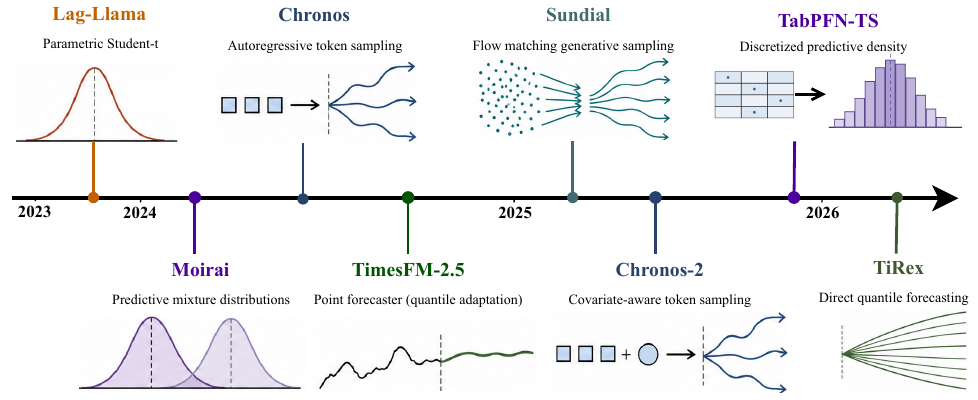}
    \caption{Evolution of time series foundation models for probabilistic forecasting.}
    \label{fig:Foundation_TS}
\end{figure}

Originating from natural language processing, recent advances in large-scale pretraining have led to the emergence of foundation models for time series forecasting \citep{kottapalli2025foundation}. In contrast to conventional forecasting models that are trained separately for individual datasets, foundation models are pretrained on large and diverse collections of time series spanning multiple domains, sampling frequencies, and forecasting horizons. They are subsequently adapted to forecasting tasks through zero-shot inference or lightweight fine-tuning, enabling strong transferability with minimal task-specific training \citep{liang2024foundation, zhao2026survey}. While several prominent foundation models such as Timer-XL \citep{liu2025timer} and IBM's TTM \citep{ekambaram2024tiny} remain dedicated to deterministic point forecasting, the field is increasingly shifting its focus toward capturing uncertainty. Consequently, foundation models have accelerated progress in probabilistic forecasting by estimating predictive uncertainty as an intrinsic part of the forecasting process. Instead of relying on post-hoc uncertainty estimation, modern foundation models directly produce predictive distributions through probabilistic mechanisms such as parametric distribution prediction, quantile regression, generative sequence modeling, and mixture distributions. 
Fig.~\ref{fig:Foundation_TS} summarizes the evolution of probabilistic time series foundation models.

\subsection{Uncertainty Quantification in Foundation Models}\label{sec_foundation_uq}
Foundation models differ not only in their architectures but also in how they estimate predictive uncertainty. While post-hoc calibration can be applied to any deterministic foundation model, this section focuses exclusively on models with intrinsic, pre-trained uncertainty representations. We organize these approaches according to their probabilistic formulation into six categories. Table~\ref{tab:tax_foundation} provides an overview of major open-source foundation models based on their base architecture and probabilistic output mechanism.

\textbf{Sequence models.} These models formulate forecasting as an autoregressive sequence-generation problem, learning the conditional distribution of future observations given historical context. Chronos \citep{ansari2024chronos}, for example, represents time series as discrete token sequences and generates future trajectories through autoregressive decoding. Prediction intervals and arbitrary quantiles are obtained by repeatedly sampling from the learned token distribution, whereas Chronos-2 \citep{ansari2025chronos2} extends this framework with covariate support and improved calibration via a quantile head. The absence of explicit distributional assumptions provides considerable flexibility, although repeated sampling increases inference cost.

\textbf{Mixture-distribution models.} These approaches represent predictive uncertainty as a weighted mixture of probability distributions. Masked encoder-based
universal time series forecasting Transformer (Moirai) \citep{woo2024unified} flattens multivariate time series into a single sequence and projects non-overlapping patches into vector representations via a multi-patch-size masked encoder. Patches within the forecast horizon are replaced by learnable mask embeddings, which are subsequently decoded to parameterize a predictive mixture distribution. Moirai learns mixtures of low-variance Gaussian, Student-$t$, log-normal, and negative binomial components, allowing the predictive distribution to adapt to heterogeneous noise characteristics. Its any-variate attention mechanism supports variable-dimensional multivariate forecasting, while Moirai-MoE \citep{liu2024moirai} improves scalability through sparse mixture-of-experts routing.

\textbf{Parametric probabilistic models.} Parametric approaches estimate the parameters of a predefined predictive distribution (see Section~\ref{sec_parametric} for details). A foundation model utilizing this technique is Lag-Llama \citep{rasul2023lag}. The terminal layer of the Lag-Llama architecture employs a dedicated distribution head, which maps the learned latent features onto the parameter space of a target probability distribution. The model's robust representational capacity can be flexibly coupled with various distribution heads to estimate the parameters of an alternative parametric family. In the original work, Lag-Llama predicts the location, scale, and degrees of freedom of a Student-$t$ distribution at each forecasting horizon, enabling efficient generation of prediction intervals and probabilistic forecasts. This formulation is computationally efficient, although its performance depends on the suitability of the assumed distribution.

\textbf{Quantile prediction models.} These models estimate conditional quantiles without specifying a parametric likelihood (see Section~\ref{sec_distributional_quantile}). Representative examples include TimesFM \citep{das2024timesfm}, TiRex \citep{auer2026tirex}, and Chronos-2 \citep{ansari2025chronos2}, which differ primarily in their backbone architectures but share the same probabilistic formulation. TimesFM is based on a decoder-only Transformer architecture and was originally introduced as a point forecaster. TimesFM-2.5 subsequently added an optional continuous quantile head, enabling direct probabilistic forecasting. TiRex, based on xLSTM \citep{beck2024xlstm}, predicts nine equidistant quantile levels: $\mathcal{Q}_{\text{set}} =\{0.1, 0.2, \ldots, 0.9\}$, and the model’s parameters are optimized by minimizing the quantile loss. Chronos-2 \citep{ansari2025chronos2} further predicts a set of 21 quantiles, resulting in a richer representation of the predictive distribution compared to the 9-quantile grid $\mathcal{Q}_{\text{set}}$. Direct quantile prediction is computationally efficient and straightforward to optimize, although coverage may deteriorate under distribution shift. The \texttt{darts} Python library \citep{herzen2022darts} implements various foundation models with a quantile regression likelihood, in order to turn them into probabilistic zero-shot forecasters.


\textbf{Generative models.} Generative modeling has become a focal point in the development of foundation models, primarily due to its efficacy in estimating complex target distributions during pre-training \citep{zhao2026survey}. Sundial \citep{liu2025sundial} is a flow-matching-based generative time-series foundation model that learns complex predictive distributions without prescribing a fixed parametric predictive family. Sundial directly processes continuous-valued time series using the TimeFlow Loss, which trains the autoregressive Transformer to map simple Gaussian noise to the target data distribution. To generate probabilistic forecasts, the model performs repeated sampling from independent Gaussian noises while efficiently reusing the shared representation of the historical context.

\textbf{Bayesian in-context models.} Prior-data fitted networks (PFNs) \citep{muller2022pfn} enable efficient in-context Bayesian prediction through pretraining on synthetic tasks sampled from a specified prior. Unlike MCMC or variational methods, which require costly posterior computation for each new dataset, a pre-trained PFN receives observed examples and query inputs to approximate the posterior predictive distribution in context without task-specific parameter updates. Importantly, this Bayesian interpretation governs the simulated task prior rather than a posterior over the network's weights. Originally introduced for tabular classification in TabPFN \citep{hollmann2023tabpfn} and extended to regression in TabPFN-v2 \citep{hollmann2025accurate}, this paradigm has recently been adapted to time series via TabPFN-TS \citep{hoo2025tables}. TabPFN-TS recasts forecasting as tabular regression by combining a running time index with calendar encodings, automatically extracted seasonal features, and known future covariates. Using the TabPFN-v2 backbone, it outputs a discretized predictive density for each future target, from which arbitrary quantiles and prediction intervals are readily derived. Complementing this approach, TempoPFN \citep{moroshan2025tempopfn} couples synthetic-prior pre-training with linear recurrent networks and quantile-loss optimization, demonstrating that PFN-based forecasting is restricted neither to Transformer backbones nor to likelihood-based output heads.

\begin{table}
\caption{Comparative overview of major open-source foundation models for probabilistic forecasting. ``Multivariate: \cmark'' denotes native support or pre-training for multivariate time series reported in the original work.}
\begin{adjustbox}{width=\textwidth}
   \centering
\begin{tabular}{cccccc}
\toprule Model & Architecture & Probabilistic Output & Zero-shot & Multivariate & \makecell{Possible\\Spatiotemporal Potential} \\
\midrule Chronos-2 & T5 & Quantiles & \ding{51} & \cmark & Indirect \\
Moirai & Transformer & Mixture distributions & \ding{51} & \ding{51} & Moderate \\
Lag-Llama & Decoder LLM & Student-$t$ distribution & \ding{51} & \xmark & Limited \\
TimesFM-2.5 & Decoder Transformer & Quantiles & \ding{51} & \ding{51} & Moderate \\
TiRex & xLSTM & Quantiles & \ding{51} & \xmark & Moderate \\
Sundial & Transformer & Generative sampling & \ding{51} & \xmark & High future potential \\

TabPFN-TS & Tabular Transformer & Discretized density & \ding{51} & \xmark\textsuperscript{$\dagger$} & Indirect \\

TempoPFN & Linear RNN & Quantiles & \ding{51} & \xmark & Moderate \\
\bottomrule
\end{tabular}
\label{tab:tax_foundation}
\end{adjustbox}
\footnotesize{
\textsuperscript{$\dagger$}Supports covariate-informed forecasting.}
\end{table}

\subsection{From Temporal to Spatiotemporal Foundation Models}
Most existing foundation models have been developed for temporal forecasting, where large-scale pretraining has enabled strong zero-shot and few-shot performance across diverse datasets. These models \citep{ansari2024chronos, das2024timesfm, auer2026tirex} demonstrate that pretrained temporal representations can generalize effectively across forecasting horizons, domains, and sampling frequencies with little or no task-specific training. Recent work has extended these capabilities beyond univariate forecasting toward multivariate settings, where models must capture temporal dynamics together with cross-variable dependencies and exogenous information. Moirai \citep{woo2024unified} addresses this challenge through any-variate attention, allowing a variable number of input channels, whereas Chronos-2 \citep{ansari2025chronos2} and TimesFM \citep{das2024timesfm} incorporate contextual covariates to improve forecasting performance. These developments broaden the applicability of foundation models but remain focused on collections of temporal sequences rather than explicitly modeling spatial structure.

Probabilistic spatiotemporal foundation models that transfer across spatial structures and application domains have yet to emerge. Existing deep spatiotemporal forecasting methods continue to rely primarily on application-specific GNNs \citep{yu2018spatio, panja2026estgcn, pathak2026engression}, Transformers \citep{wang2024spatiotemporal}, diffusion models \citep{wen2023diffstg, Chen2024diffusion}, and physics-informed architectures trained for individual tasks. Although such recent advances in multivariate representation learning, long-context modeling, and probabilistic generation provide important building blocks, extending foundation models to full spatiotemporal forecasting introduces additional challenges. These include scalable pretraining over heterogeneous spatial graphs, evolving dependency structures, irregular spatial observations, uncertainty propagation across networks, and the integration of physical constraints. Concurrently, graph foundation models (GFMs) have emerged, designed to learn generalizable, transferable structural representations across diverse and heterogeneous graph topologies \citep{liu2025graph, wang2025graph}. One possible direction is to combine GFMs, which encode spatial structure, with time-series foundation models that learn temporal dynamics. Such architectures must also model space-time interactions explicitly rather than treating spatial and temporal dependence as separable. For probabilistic forecasting, they must represent joint uncertainty across locations and forecast horizons while maintaining spatial coherence and calibration.

Overall, the foundation models do not introduce a distinct form of uncertainty representation. Instead, pretraining shifts much of the fitting effort and inductive structure to an earlier stage, while the deployed model still produces densities, quantiles, samples, and point forecasts. These outputs must be evaluated in the target domain, and external calibration may still be required. Foundation models should therefore be compared by their predictive mechanism, adaptation requirements, calibration conditions, and inference cost. These dimensions provide the basis for the cross-paradigm comparison in Section \ref{sec_cross_paradigm_synthesis}.

\section{Cross-Paradigm Synthesis of Uncertainty Representations}\label{sec_cross_paradigm_synthesis}

\begin{table}[p]
\centering
\caption{Cross-paradigm comparison of probabilistic forecasting methods in terms of uncertainty representation, dependence
structure, validity conditions, computational requirements, and practical considerations.}
\label{tab:cross_paradigm_synthesis}
\begingroup
\fontsize{6.7}{7.7}\selectfont
\setlength{\tabcolsep}{2pt}
\renewcommand{\arraystretch}{1.04}
\begin{tabularx}{\linewidth}{
 >{\RaggedRight\arraybackslash}p{0.08\linewidth}
 >{\RaggedRight\arraybackslash}p{0.12\linewidth}
 >{\RaggedRight\arraybackslash}p{0.15\linewidth}
 >{\RaggedRight\arraybackslash}p{0.16\linewidth}
 >{\RaggedRight\arraybackslash}p{0.16\linewidth}
 >{\RaggedRight\arraybackslash}p{0.15\linewidth}
 >{\RaggedRight\arraybackslash}p{0.14\linewidth}}
\toprule
Family & Taxonomic placement & Uncertainty representation &
Dependence and expressiveness & Validity conditions &
Computational requirements & Limitations and suitable conditions \\
\midrule
Pre-control (\ref{sec_precontrol_limits}) & \textbf{Primary:} model-agnostic \textit{Overlap:} parametric &
Historical residual variability; intervals or an assumed residual law. &
Usually marginal; joint dependence needs an additional residual model. &
Validity depends on residual law, stability, and estimation; no general coverage guarantee. & 
\textbf{Fitting:} Base fit plus residual estimation \textbf{Inference:} Direct interval construction. &
Shifting variance or tails; simple, stable error structures. \\
Ensembles and BMA (\ref{sec_ensemble})& \textbf{Primary:} model-agnostic \textit{Overlap:} Bayesian, distributional &
Response/model variation; empirical conditional laws (QRF), mixtures or samples. &
Joint only if members/resampling retain it; diversity limits spread. &
No automatic calibration; resampling validity or Bayesian interpretation depends on its construction. &
\textbf{Fitting:} Multiple fits or external weights \textbf{Inference:} Aggregate predictions. &
Correlated members or misspecified components; available diverse forecasters. \\
CP and CQR (\ref{sec_distribution_free}) & \textbf{Primary:} model-agnostic \textit{Overlap:} quantile regression &
Calibration scores modify uncertainty; usually sets or intervals. &
Standard coordinate coverage is marginal; no joint law supplied by the correction alone. &
Split finite-sample marginal coverage under exchangeability and adjusted quantile; extensions need their own assumptions. &
\textbf{Fitting:} Base fit plus held-out calibration \textbf{Inference:} corrected predictions. &
Unrepresentative scores; interval reliability when validity conditions are defensible. \\
Bayesian models (\ref{sec_bayesian}) & \textbf{Primary:} model-intrinsic \textit{Overlap:} parametric &
Specified parameter, function, state and observation uncertainty; posterior predictive law. &
Joint through the specified process/covariance; support and prior restrict flexibility. &
Model-based posterior uncertainty; no generic distribution-free finite-sample guarantee. &
\textbf{Fitting:} Posterior inference \textbf{Inference:} integration or sampling; approximation-dependent. &
Prior/likelihood mismatch; scientific structure and parameter/state uncertainty. \\
Parametric heads (\ref{sec_parametric}) & \textbf{Primary:} model-intrinsic \textit{Overlap:} mixtures, Bayesian &
Predicted likelihood parameters; density and summaries. &
Marginal unless a multivariate or autoregressive law is specified. &
Likelihood family and estimation govern validity; fitting is not a coverage guarantee. &
\textbf{Fitting:} Likelihood optimization \textbf{Inference:} direct evaluation or draws. &
Support/tail misspecification; manageable structured likelihoods. \\
Quantile regression (\ref{sec_distributional_quantile}) & \textbf{Primary:} model-intrinsic \textit{Overlap:} ensembles &
Conditional quantiles; selected functionals and intervals. &
Finite grids leave interpolation, tails and joint dependence unspecified. &
Quantile identification at population level; finite-data coverage not automatic. &
\textbf{Fitting:} Quantile-loss or forest fitting \textbf{Inference:} quantile extraction. &
Crossing, sparse tails; decisions needing specific quantiles. \\
Engression (\ref{sec_distributional_engression}) & \textbf{Primary:} model-intrinsic \textit{Overlap:} generative &
Injected noise and conditional transformation; forecast samples. &
Joint if vector outputs/noise are modeled jointly; generator family limits fit. &
Proper-score identification subject to score/model conditions; not a finite-sample coverage guarantee. &
\textbf{Fitting:} Sample-based score fitting \textbf{Inference:} repeated generator evaluations. &
Choice of base noise distribution; distributional regression with samples. \\
Loss-driven intervals (\ref{sec_loss_driven}) & \textbf{Primary:} model-intrinsic \textit{Overlap:} calibration &
Interval objective balances coverage/width; bounds. &
Typically marginal interval output, not a full predictive law. &
Empirical objective quality; theoretical validity requires a specific result and assumptions. &
\textbf{Fitting:} Interval-loss optimization \textbf{Inference:} direct bounds. &
Training coverage fails to transfer; interval-focused objectives. \\
Structured generative (\ref{sec_generative}) & \textbf{Primary:} model-intrinsic \textit{Overlap:} Bayesian, parametric &
Copulas, states, categorical laws or weighted scenarios; joint construction. &
Explicit dependence, state, resolution or scenario constraints. &
Validity depends on component laws, structure and estimation; no automatic calibration. &
\textbf{Fitting:} Structure/state or scenario fitting \textbf{Inference:} density, sequential or finite-scenario generation. &
Misspecified dependence or limited resolution; interpretable joint structure. \\
Latent/ implicit generators (\ref{sec_generative}) & \textbf{Primary:} model-intrinsic \textit{Overlap:} distributional &
VAE, GAN or prior mapping; latent/noise-induced samples. &
Flexible joint vectors when trained jointly; support and architecture still matter. &
Objective dependent; not all have tractable exact likelihood or coverage. &
\textbf{Fitting:} Latent or adversarial fitting \textbf{Inference:} generator draws. &
Lost modes or inaccurate tails; sample-based nonlinear futures. \\
Normalizing flows (\ref{sec_generative_nf}) & \textbf{Primary:} model-intrinsic \textit{Overlap:} density regression &
Invertible transformations of a base law; density and samples. &
Flexible joint densities within invertibility/support constraints. &
Exact evaluation of the fitted likelihood, not exact knowledge of the true law or calibrated coverage. &
\textbf{Fitting:} Flow/Jacobian fitting \textbf{Inference:} direction- and architecture-dependent generation. &
Restrictive transformation or costly direction; density access and joint sampling. \\
Diffusion/ score models (\ref{sec_generative_diffusion}) & \textbf{Primary:} model-intrinsic \textit{Overlap:} generative &
Conditional stochastic generation; trajectories or latent futures. &
Flexible joint outputs if conditioning/objective model them jointly. &
Score/ denoising and numerical approximations; no generic coverage guarantee. &
\textbf{Fitting:} Score/ denoising fit \textbf{Inference:} repeated steps and draws; solver-dependent. &
Sampling budget or weak conditioning; complex multimodal trajectories. \\
Foundation models (\ref{sec_foundation_models}) & Foundation models &
Pretrained density, quantile, prior-fitting or generative mechanism. &
Marginal or joint according to the head/generator, not pretraining status. &
Transfer depends on task/prior alignment; native outputs may still need calibration. &
\textbf{Fitting:} Upstream pretraining \textbf{Inference:} local setup and optional adaptation. &
Domain shift or unsuitable outputs; reuse across tasks with appropriate validation. \\
\bottomrule
\end{tabularx}
\endgroup
\end{table}
The previous sections individually review model-agnostic, model-intrinsic, and foundation model approaches; however, their practical differences are not fully captured by the taxonomy alone. The methods vary in the form of uncertainty they represent, the dependencies they capture, the assumptions required for validity, and their computational demands. This section compares these dimensions and identifies where the different families overlap and complement each other. Table~\ref{tab:cross_paradigm_synthesis} summarizes the findings across cross-paradigm models. The taxonomy reflects the stage at which uncertainty is primarily introduced into the forecasting pipeline, but several methods span more than one family. 
For instance, as discussed in Section~\ref{sec_generative}, copula-based models and engression exhibit characteristics of parametric and generative approaches, respectively, despite their primary classifications within the generative and distributional regression families. BMA (Section~\ref{sec_bma}), CQR (Section~\ref{sec_cqr}), and QRF (Section~\ref{sec_randomization_based}) represent additional boundary cases. BMA is Bayesian in its treatment of model weights, yet it is classified as a model-agnostic technique because uncertainty is aggregated at the output level of the component forecasters. Similarly, CQR integrates intrinsic quantile estimation with a post-hoc conformal correction, placing it among calibration methods based on its final adjustment step. Finally, QRF is classified within the ensemble family due to its random forest construction, even though its capacity to estimate conditional response distributions closely links it to distributional regression.

A foundational distinction across the paradigms in Table~\ref{tab:cross_paradigm_synthesis} is the separation between representational expressiveness and statistical calibration. While a generative objective or specified likelihood determines how a predictive law is learned, calibration procedures govern how its reported uncertainty is subsequently adjusted. Increasing architectural flexibility can enlarge the class of representable
predictive distributions, but neither accurate estimation from finite data nor finite-sample coverage follows from this capacity alone. Conversely, distribution-free calibration (such as standard conformal prediction) guarantees marginal coverage under exchangeability, yet provides no explicit predictive density, conditional coverage, or joint trajectory law. Consequently, highly expressive generative forecasters and post-hoc calibration wrappers could be viewed as complementary rather than competing strategies.

This complementarity is especially critical when modeling spatiotemporal dependence. Conditioning a neural network on spatial historical features does not intrinsically capture the dependence structure among future forecast errors. Furthermore, marginal predictive distributions and standard evaluation metrics such as univariate CRPS or the pinball loss cannot distinguish between independent and perfectly correlated simultaneous outcomes. Resolving the full joint law across forecast horizons and spatial locations requires explicit structural mechanisms, such as joint stochastic simulation, copulas, multivariate likelihoods, or shared latent dynamics, evaluated via proper multivariate scoring rules \citep{gneiting2007proper} discussed in Section~\ref{sec_metrics}.

Methodological choices provide such representational flexibility for statistical and computational tractability. Parametric likelihoods restrict the class of representable distributions to enable direct probability calculations. Finite quantile grids provide distribution-free marginal flexibility but leave interpolation, tail behavior, and joint dependence entirely unspecified. Bayesian integration offers rigorous uncertainty propagation but remains sensitive to prior specification and posterior approximation quality, while flexible conditional samplers may still overlook rare modes in finite datasets. Even when exact likelihoods are preserved alongside high expressiveness, as in invertible normalizing flows \citep{papamakarios2021nf}, this is achieved through strict architectural constraints.


Ultimately, synthesizing these cross-paradigm characteristics yields a structured sequence for model selection. Rather than searching for a universally dominant architecture, practitioners must identify the decision-relevant output, verify defensible dependence and calibration assumptions, and allocate feasible computational budgets. These foundational checks motivate the practical model guidance detailed in Section~\ref{sec_model_guidance}. Moreover, the comparison above identifies several limitations that persist across otherwise different uncertainty mechanisms. First, a joint predictive law conditional on
an estimated graph or covariance structure need not propagate uncertainty in that structure, and neither joint representation nor marginal calibration alone ensures reliability when dependencies change. Second, flexible output
distributions do not automatically respect the support, topology, or physical constraints of the forecasting target. Third, expressive likelihoods and
generators still face limited information about rare events, while marginal scores cannot assess joint tail dependence. Fourth, richer representations and uncertainty propagation can increase inference and calibration costs, and
comparisons remain difficult when outputs and software interfaces differ. 
Section~\ref{sec_challenges} derives research objectives from these
limitations and discusses several open challenges.

\section{Evaluation and Model Selection}\label{sec_evaluation_model_selection}
This section provides a practical guide to the aspects of evaluating probabilistic forecasting methods. Section~\ref{sec_datasets} covers data resources, benchmarks, and open-source implementations available. Section~\ref{sec_metrics} presents probabilistic metrics to measure forecast accuracy, while Section~\ref{sec_robustness} discusses statistical significance tests to evaluate model comparisons. Finally, Section~\ref{sec_model_guidance} offers practical guidance on method selection. All tables in this section are deferred to the Appendix for brevity in the main text.

\subsection{Data Resources and Software Tools}\label{sec_datasets}
The empirical evaluation of probabilistic forecasting methods requires access to diverse datasets, standardized benchmarks for fair comparison, and open-source implementations for reproducibility. This section provides the key resources available to researchers and practitioners across temporal and spatiotemporal settings. We first describe widely used datasets and evaluation benchmarks (Section~\ref{sec_datasets_benchmark}), followed by a summary of open-source libraries and model implementations (Section~\ref{sec_datasets_lib}).

\subsubsection{Datasets and Evaluation Benchmarks}\label{sec_datasets_benchmark}
The rich literature in the domain of (probabilistic) forecasting has led to the development and availability of high quality datasets and benchmarks, both for temporal and spatiotemporal settings. Table~\ref{tab:datasets} presents a non-exhaustive list of publicly available datasets for temporal and spatiotemporal forecasting, along with their sources and works utilizing them. These datasets span a wide range of domains including transportation (traffic forecasting), healthcare (epidemic forecasting), energy, computer vision (human pose forecasting), and finance. Table~\ref{tab:benchmarks} lists a set of resources including data repositories, benchmarks, and model evaluation papers for univariate, multivariate, and spatiotemporal forecasting tasks along with links to the respective public repositories.

\subsubsection{Open-Source Libraries and Repositories}\label{sec_datasets_lib}

Table~\ref{tab:UQ_Methods_Implementation} summarizes publicly available implementations of representative probabilistic forecasting methods discussed throughout this survey. The implementations are organized according to the forecasting setting, including univariate, multivariate, and spatiotemporal models. For each method, we report the programming environment, official repository when available, and the corresponding reference to support reproducing published results and comparing state-of-the-art approaches. In addition to model-specific implementations, several open-source libraries provide general frameworks for developing and evaluating probabilistic forecasting models. For Bayesian inference, libraries such as \texttt{PyMC} \citep{pymc2023}, \texttt{Pyro} \citep{bingham2018pyro}, \texttt{NumPyro} \citep{phan2019composable}, and \texttt{TensorFlow Probability}\footnote{\url{https://github.com/tensorflow/probability}} in Python, alongside \texttt{Stan}, \texttt{brms} \citep{burkner2021bayesian}, and \texttt{nimble} \citep{de2017programming} in R, support MCMC, variational inference, and probabilistic programming. For latent Gaussian and spatial models, \texttt{R-INLA} \citep{lindgren2015bayesian} and \texttt{pyINLA} \citep{fattah2026pyinla} enable efficient approximate Bayesian inference using INLA (Section~\ref{sec_approximate_inla}). Gaussian process modeling is available through packages such as \texttt{GPyTorch} \citep{gardner2018gpytorch} and \texttt{GPflow} \citep{matthews2017gpflow} in Python and \texttt{GPfit} \citep{macdonald2015gpfit}, \texttt{DiceKriging} \citep{roustant2012dicekriging}, \texttt{GpGp}\footnote{\url{https://cran.r-project.org/package=GpGp}}, and \texttt{spBayes} \citep{finley2007spbayes} in R. For general probabilistic forecasting frameworks, several open-source libraries offer comprehensive implementations. In Python, libraries such as \texttt{Darts} \citep{herzen2022darts}, \texttt{NeuralForecast}\footnote{\url{https://github.com/nixtla/neuralforecast}}, \texttt{GluonTS} \citep{alexandrov2020gluonts}, and \texttt{PyTorch Forecasting}\footnote{\url{https://github.com/sktime/pytorch-forecasting}}, alongside R packages including \texttt{forecast} \citep{hyndman2008automatic} and \texttt{fable} \citep{fable2026}, support prediction intervals, likelihood estimation, and deep learning architectures. R packages such as \texttt{forecast} and \texttt{fable} focus on classic prediction intervals and distributional models. Python libraries including \texttt{Darts}, \texttt{NeuralForecast}, \texttt{GluonTS}, \texttt{PyTorch Forecasting}, \texttt{stengression} \citep{pathak2026engression}, and \texttt{Genformer} \citep{pathak2026entransformer} implement parametric likelihood models, deep regression-based approaches, and deep generative architectures. Model-agnostic approaches are supported across both ecosystems. In R, packages such as \texttt{ensembleBMA}\footnote{\url{https://cran.r-project.org/package=ensembleBMA}} and \texttt{conformalForecast}\footnote{\url{https://cran.r-project.org/package=conformalForecast}} enable Bayesian model averaging and conformal prediction, respectively. In Python, libraries including \texttt{MAPIE} \citep{taquet2022mapie} and \texttt{TorchCP} \citep{huang2025torchcp} provide flexible distribution-free uncertainty quantification.

\subsection{Probabilistic Evaluation Metrics}\label{sec_metrics}
Probabilistic forecasts are evaluated differently from point forecasts because the objective is to assess an entire predictive distribution, rather than a single prediction for each time step over the forecast horizon. Evaluation therefore combines proper scoring rules for overall distributional quality with separate diagnostics for calibration and sharpness \citep{gneiting2007proper, gneiting2014probabilistic}. Calibration measures the statistical consistency between forecasts and observations, whereas sharpness quantifies the concentration of predictive distributions \citep{gneiting2007probForecastandCalibration}. The appropriate metric depends on the form of the probabilistic forecast, including predictive distributions, quantiles, prediction intervals, or ensembles. For univariate probabilistic forecasts, the continuous ranked probability score (CRPS) is one of the most widely used proper scoring rules \citep{gneiting2007proper}. Given a predictive cumulative distribution function $F_t$ and observation $y_t$, it is defined as:
$$ \mathrm{CRPS}(F_t,y_t) = \frac{1}{H} \sum_{t = T+1}^{T+H}\int_{-\infty}^{\infty} \left(F_t(x)-\mathbb{I}_{\{x\geq y_t\}}\right)^2 dx.
$$
CRPS generalizes the absolute error to probabilistic forecasts. For multivariate forecasts, the (negative) energy score (ES) extends CRPS to vector-valued predictions:
$$
\mathrm{ES}(F_t,y_t) = \mathbb{E}\|Y_t-y_t\|^\beta - \frac{1}{2}\mathbb{E}\|Y_t-Y'_t\|^\beta,
$$
where $Y_t$ and $Y'_t$ are independent samples from the predictive distribution and $\beta\in(0,2)$ is a hyperparameter. However, the ES can be insensitive to dependence misspecification \citep{michael2015variogram}. The variogram score (VS) addresses this limitation by evaluating pairwise dependence structures:
$$\mathrm{VS}_{p}(F,\mathbf{y}) = \sum_{s_1=1}^{N} \sum_{t_1=1}^{T} \sum_{s_2=1}^{N} \sum_{t_2=1}^{T} w_{s_1,t_1,s_2,t_2} \left(\vert{}y_{s_1,t_1}-y_{s_2,t_2}\vert{}^{p} - \mathbb{E}_{F}\vert{}Y_{s_1,t_1}-Y_{s_2,t_2}\vert{}^{p}\right)^2,$$
where $\mathbf{y}$ denotes the full set of spatiotemporal observations over the entire domain and time horizon, $F$ is the joint predictive distribution of the spatiotemporal forecast, $y_{s_1,t_1}, y_{s_2,t_2}$ are the realized observations at two spatiotemporal points $(s_1, t_1)$ and $(s_2, t_2)$, respectively, $Y_{s_1,t_1}, Y_{s_2,t_2}$ are the corresponding forecast random variables drawn from $F$, $p$ is the order of the VS, and 
$w_{s_1,t_1,s_2,t_2}$ are non-negative weights assigned to the pair of spatiotemporal points \citep{michael2015variogram}. 
The weights can be tuned to emphasize specific distance scales or downweight distant pairs to reduce sampling error \citep{michael2015variogram, jung2008scale}. For forecasts represented as quantiles or prediction intervals, performance is commonly evaluated using the pinball loss:
$$
L_{\tau}(q_{t,\tau},y_t) = \left(\tau-\mathbb{I}_{\{y_t < q_{t, \tau}\}}\right)(y_t - q_{t, \tau}),
$$
where $q_{t,\tau}$ is the $\tau$-quantile forecast at time $t$ \citep{gneiting2023quantile}. To evaluate overall distributional coverage, the $\rho$-risk metric normalizes the pinball loss and aggregates it across a range of quantile levels $\tau$:
$$\rho\text{-risk} = \frac{2\sum_{t=T+1}^{T+H}\max\left[ \tau(y_t-\widehat{y}_{t}), (1-\tau)(\widehat{y}_{t} - y_t)\right]}{\sum_{t=T+1}^{T+H}\vert{}y_t\vert{}},$$
where lower $\rho$-risk values indicate more accurate quantile forecasts across the predictive distribution \citep{seeger2016bayesian, salinas2020deepar}. For prediction intervals, the Winkler score provides a proper scoring rule that jointly evaluates interval sharpness and coverage \citep{winkler1996scoring}. Given a prediction interval $\left[\widehat{f}_{t}^{L}, \widehat{f}_{t}^{U}\right]$ with a coverage level $1-\alpha$ and an observation $y_t$, it is defined as:
$$
W_{\alpha,t}
\begin{cases}
\left(\widehat{f}_{t}^{U}-\widehat{f}_{t}^{L}\right)
+\frac{2}{\alpha}\left(\widehat{f}_{t}^{L}-y_t\right),
& y_t<\widehat{f}_{t}^{L},\\[6pt]
\widehat{f}_{t}^{U}-\widehat{f}_{t}^{L},
& \widehat{f}_{t}^{L}\leq y_t\leq \widehat{f}_{t}^{U},\\[6pt]
\left(\widehat{f}_{t}^{U}-\widehat{f}_{t}^{L}\right)
+\frac{2}{\alpha}\left(y_t-\widehat{f}_{t}^{U}\right),
& y_t>\widehat{f}_{t}^{U}.
\end{cases}
$$
The score reduces to the interval width when the observation lies within the interval and adds penalties proportional to the distance outside the bounds; otherwise, penalizing under-dispersion so that lower scores reflect sharper, better-calibrated intervals \citep{winkler1996scoring, pathak2026engression}. In addition to the Winkler score, the prediction interval coverage probability (PICP) and mean prediction interval width (MPIW) are commonly reported \citep{anand2026tube}:
$$
\mathrm{PICP} = \frac{1}{H} \sum_{t=T+1}^{T+H} \mathbb{I}_{\left\{\widehat{f}_{t}^{L}\leq y_t\leq \widehat{f}_{t}^{U}\right\}} \qquad \text{and} \qquad \mathrm{MPIW} = \frac{1}{H}\sum_{t=T+1}^{T+H}\left(\widehat{f}_{t}^{U}-\widehat{f}_{t}^{L}\right).
$$
PICP quantifies the proportion of observations falling within the predicted intervals, whereas MPIW assesses forecast sharpness. Reliable interval forecasts achieve nominal coverage while minimizing interval width \citep{gneiting2007probForecastandCalibration}. 
For ensemble forecasts, the probability integral transform (PIT) assesses calibration by computing the normalized rank of the true observation $y_t$ within an ensemble $\{\widehat{y}_t^{(m)}\}_{m=1}^{M}$:
$$
\mathrm{PIT} = \frac{1}{M+1} \sum_{m=1}^{M} \mathbb{I}\left(\widehat y_t^{(m)}< y_t\right).
$$ 
For well-calibrated forecasts, PIT values follow a uniform distribution, whereas deviations signal bias or miscalibrated uncertainty dispersion \citep{gneiting2007proper}.
In practice, evaluation pairs proper scoring rules for overall predictive quality with PIT, coverage, or interval-width metrics to assess calibration.

\subsection{Robustness and Statistical Significance Tests}\label{sec_robustness}
The metrics presented in Section~\ref{sec_metrics} quantify predictive performance but do not determine whether observed differences between forecasting methods are statistically significant. 
Statistical significance tests distinguish meaningful performance improvements from random sampling variation. For probabilistic forecasts, conventional hypothesis tests apply by evaluating differences in proper scoring rules, such as CRPS, rather than traditional point-forecast errors. The Diebold-Mariano (DM) test \citep{diebold1995dm} remains the standard approach for pairwise comparison of forecasting methods. For two competing forecasters, $F$ and $G$, evaluated across $J$ cases, let $d_t = S(F_t, y_t) - S(G_t, y_t)$ denote the score differential. Under the null hypothesis of equal predictive performance ($\mathbb{E}[d_t] = 0$), the test statistic:
$$ T_J = \sqrt{J}\,\frac{\bar{d}_J}{\widehat\sigma_J} $$
is asymptotically standard Gaussian, where $\bar{d}_J$ and $\widehat\sigma_J^2$ are the sample mean and variance estimators of $d_i$, respectively. 
The Giacomini-White (GW) test \citep{giacomini2006gw} extends the DM framework to evaluate conditional predictive ability by testing whether loss differences depend on time-$t$ information $h_t$. The test statistic follows an asymptotic chi-square distribution with $\dim(h_t)$ degrees of freedom and reduces to the standard DM test when $h_t = 1$. The GW framework supports rolling and recursive re-estimation, making it ideal for testing adaptive forecasting systems; though it remains limited to pairwise comparisons. Comparing larger collections of forecasting methods requires procedures that control multiple comparisons. \cite{mariano2012multidm} generalized the DM framework to jointly test the equal predictive ability of several forecasting models using a Wald-type statistic, avoiding repeated pairwise testing. The model confidence set (MCS) procedure \citep{hansen2011mcs} sequentially eliminates inferior models until a superior set of statistically indistinguishable forecasting methods remains. A complementary alternative is the multiple comparisons with the best (MCB) test \citep{nemenyi1963distribution, koning2005mcb}, which ranks competing methods relative to the best-performing model and constructs confidence intervals for rank differences. 
Extending significance testing to spatiotemporal forecasting presents additional challenges because forecast errors exhibit both temporal and spatial dependence. For panel data, \cite{akgun2024panel} proposed two equal predictive ability tests, namely $S$-statistics and $C$-statistics. The former tests whether two forecasters perform equally well on average across all panel units and time periods, while the latter evaluates whether equal predictive performance holds jointly across predefined clusters of units. The AZ-whiteness test \citep{zambon2022az} offers a complementary diagnostic for graph-structured spatiotemporal data. It uses sign-based statistics across adjacent time steps and connected nodes to evaluate whether model residuals retain exploitable spatial or temporal dependencies.

\subsection{Method Guidance}\label{sec_model_guidance}
Selecting an appropriate probabilistic forecasting method requires navigating several competing considerations, including data availability, distributional characteristics, uncertainty requirements, computational constraints, and interpretability. No single approach is optimal across all settings. Methods with strong theoretical guarantees, such as conformal prediction (Section~\ref{sec:agnostic_conformal}), provide finite-sample marginal coverage under exchangeability but may require additional calibration strategies to adapt to nonstationary uncertainty (Section~\ref{sec_cqr}). Approaches with explicit probabilistic structure, such as Bayesian hierarchical models and Gaussian processes (Section~\ref{sec_classical_bayesian}), offer interpretability and principled uncertainty decomposition but face scalability challenges in large spatiotemporal systems compared with deep generative approaches (Section~\ref{sec_generative}). Conversely, highly expressive generative models, such as diffusion models (Section~\ref{sec_generative_diffusion}), can represent complex predictive distributions but often introduce substantial computational costs during inference. To summarize these trade-offs, Section~\ref{sec_model_guidance_data_driven} presents a decision flowchart that maps forecasting requirements onto a specific recommended methodology, while Section~\ref{sec_model_guidance_domain_specific} examines how these considerations translate into practice across different application domains.

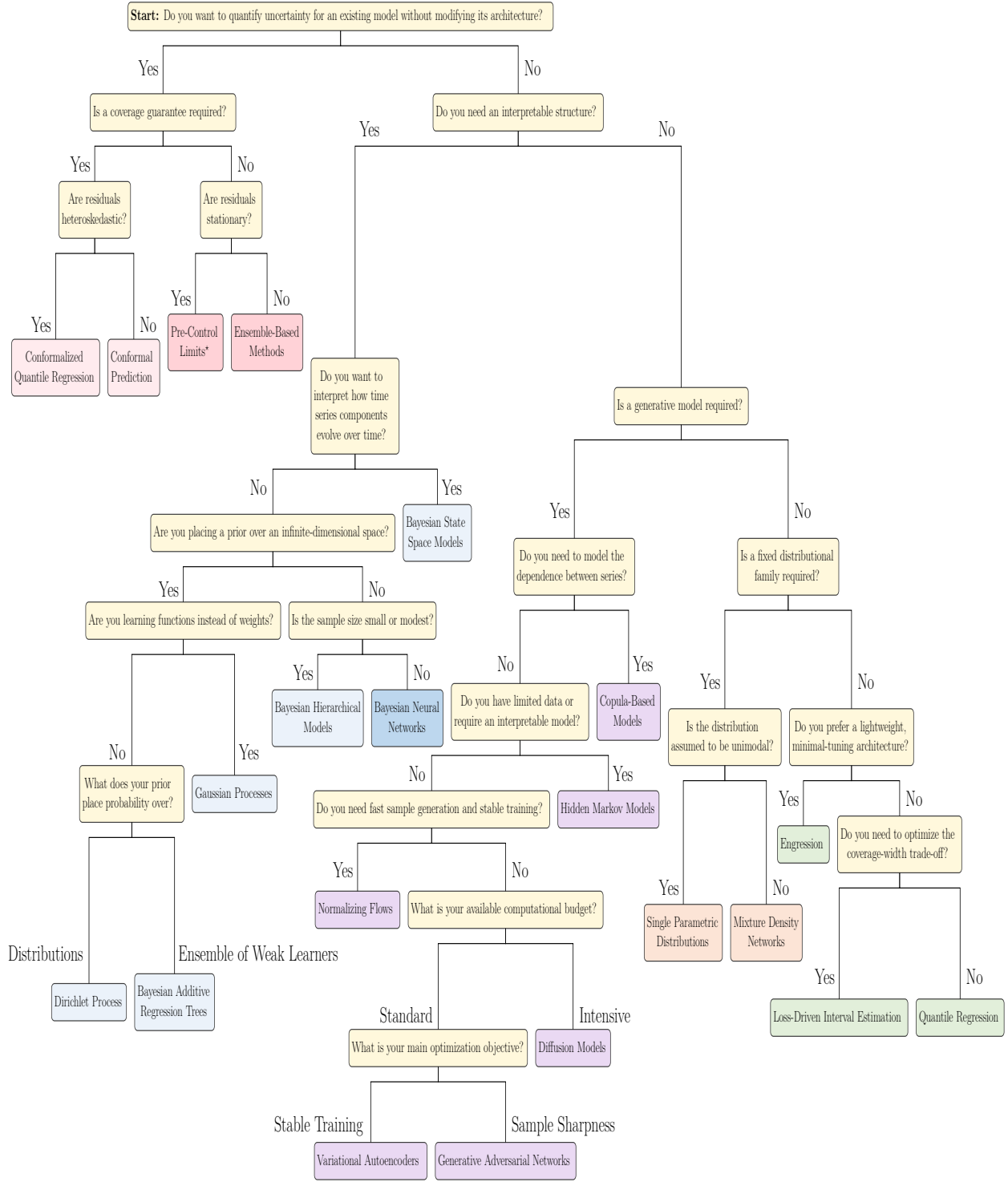
\begin{figure}[p]
\centering
\resizebox{\textwidth}{0.85\textheight}{
\begin{forest}
  guidance tree,
  for tree={
    grow=south,
    parent anchor=south,
    child anchor=north,
    anchor=center,
    align=center,
    draw,
    rounded corners=5pt,
    line width=0.5pt,
    edge={draw, line width=0.5pt},
    edge path={
    \noexpand\path[\forestoption{edge}] (!u.parent anchor) -- ++(0,-6mm) -| (.child anchor)\forestoption{edge label};
    },
    l sep=40mm,
    s sep=5mm,
    font = \LARGE,
    outer sep       = 0pt,
    inner xsep      = 5pt,
    inner ysep      = 10pt,
  },
  where level=0{fill=Decision, text width=25.5cm, inner sep=5pt, l sep = 25mm}{},
[{\textbf{Start:} Do you want to quantify uncertainty for an existing model without modifying its architecture?}, name=root
  [{Is a coverage guarantee required?}, fill=Decision, text width=8.5cm, l sep = 20mm,
    edge label={node[midway, xshift=-25pt, yshift=-25pt, font=\Huge]{Yes}}
    [{Are residuals\\ heteroskedastic?}, fill=Decision, text width=4cm, l sep = 40mm,
        edge label={node[midway, xshift=-25pt, yshift=-20pt, font=\Huge]{Yes}}
        [{Conformalized \\Quantile Regression}, fill=AgnosticLeaf, text width=5cm,
          edge label={node[midway, xshift=-25pt, yshift=-80pt, font=\Huge]{Yes}}]
        [{Conformal\\ Prediction}, fill=AgnosticLeaf, text width=2.8cm,
          edge label={node[midway,xshift=25pt, yshift=-80pt, font=\Huge]{No}}]
      ]
      [{Are residuals\\ stationary?}, fill=Decision, text width=3.4cm, l sep = 30mm,
        edge label={node[midway, xshift=25pt, yshift=-20pt, font=\Huge]{No}}
        [{Pre-Control \\ Limits\textsuperscript{$\star$}}, fill=Agnostic, text width=3cm,
          edge label={node[midway, xshift=-25pt, yshift=-50pt, font=\Huge]{Yes}}]
        [{Ensemble-Based\\ Methods}, fill=Agnostic, text width=4cm,
          edge label={node[midway,xshift=25pt, yshift=-50pt,font=\Huge]{No}}]
      ]
  ]
  [{Do you need an interpretable structure?}, fill=Decision, text width=10cm, l sep = 90mm, 
    edge label={node[midway,xshift=25pt, yshift=-25pt,font=\Huge]{No}}
      [{Do you want to\\ interpret how time\\ series components\\ evolve over time?}, fill=Decision, text width=5cm, l sep = 20mm,
        edge label={node[midway, xshift=25pt, yshift=20pt, font=\Huge]{Yes}}
        [{Are you placing a prior over an infinite-dimensional space?}, fill=Decision, text width=14.5cm, l sep = 20mm,
          edge label={node[midway, xshift=-25pt, yshift=-20pt,font=\Huge]{No}}
          [{Are you learning functions instead of weights?}, fill=Decision, text width=11.5cm, l sep = 50mm,
            edge label={node[midway, xshift=-25pt, yshift=-25pt,font=\Huge]{Yes}}
            [{What does your prior\\ place probability over?}, fill=Decision, text width=6cm, l sep = 60mm, edge label={node[midway, xshift=-25pt, yshift=-110pt,font=\Huge]{No}}
            [{Dirichlet Process}, fill=BayesianLeaf, text width=4.2cm,
            edge label={node[midway, xshift=-75pt, yshift=-130pt,font=\Huge]{Distributions}}]
            [{Bayesian Additive \\ Regression Trees}, fill=BayesianLeaf, text width=4.5cm,
            edge label={node[midway, xshift=145pt, yshift=-130pt,font=\Huge]{Ensemble of Weak Learners}}]
            ]
            [{Gaussian Processes}, fill=BayesianLeaf, text width=4.8cm,
            edge label={node[midway, xshift=25pt, yshift=-110pt,font=\Huge]{Yes}}]
           ] 
          [{Is the sample size small or modest?}, fill=Decision, text width=8.6cm, l sep = 20mm, edge label={node[midway, xshift=25pt, yshift=-25pt,font=\Huge]{No}}
            [{Bayesian Hierarchical\\ Models}, fill=BayesianLeaf, text width=5.2cm,
              edge label={node[midway, xshift=-25pt, yshift=-20pt,font=\Huge]{Yes}}]
            [{Bayesian Neural\\ Networks}, fill=Bayesian, text width=4cm,
              edge label={node[midway, xshift=25pt, yshift=-20pt,font=\Huge]{No}}]
          ]
        ]
        [{Bayesian State\\ Space Models}, fill=BayesianLeaf, text width=3.8cm,
          edge label={node[midway, xshift=25pt, yshift=-20pt, font=\Huge]{Yes}}]
      ]
      [{Is a generative model required?}, fill=Decision, text width=7.8cm, l sep = 45mm,
        edge label={node[midway, xshift=-25pt, yshift=20pt, font=\Huge]{No}}
        [{Do you need to model the \\dependence between series?}, fill=Decision, text width=7cm, l sep = 35mm,
          edge label={node[midway,xshift=-25pt, yshift=-80pt, font=\Huge]{Yes}}
          [{Do you have limited data or\\ require an  interpretable model?}, fill=Decision, text width=8cm, l sep = 20mm,
            edge label={node[midway,xshift=-25pt, yshift=-60pt, font=\Huge]{No}}
            [{Do you need fast sample generation and stable training?}, fill=Decision, text width=14cm, l sep = 25mm,
              edge label={node[midway,xshift=-25pt, yshift=-20pt,font=\Huge]{No}}
              [{Normalizing Flows}, fill=Generative, text width=4.8cm,
                edge label={node[midway,xshift=-25pt, yshift=-30pt,font=\Huge]{Yes}}]
              [{What is your available computational budget?}, fill=Decision, text width=11.5cm, l sep = 40mm,
                edge label={node[midway,xshift=25pt, yshift=-30pt,font=\Huge]{No}}
                [{What is your main optimization objective?}, fill=Decision, text width=10.5cm, l sep = 30mm,
                  edge label={node[midway, xshift=-55pt, yshift=-80pt,font=\Huge]{Standard}}
                  [{Variational Autoencoders}, fill=Generative, text width=6.5cm,
                    edge label={node[midway,xshift=-90pt, yshift=-50pt,font=\Huge]{Stable Training}}]
                  [{Generative Adversarial Networks}, fill=Generative, text width=8.2cm,
                    edge label={node[midway,xshift=100pt, yshift=-50pt,font=\Huge]{Sample Sharpness}}]
                ]
                [{Diffusion Models}, fill=Generative, text width=4.2cm,
                  edge label={node[midway,xshift=55pt, yshift=-80pt,font=\Huge]{Intensive}}
                ]
              ]
            ]
            [{Hidden Markov Models}, fill=Generative, text width=5.8cm,
              edge label={node[midway,xshift=25pt, yshift=-20pt, font=\Huge]{Yes}}]
          ]
          [{Copula-Based\\ Models}, fill=Generative, text width=3.5cm,
            edge label={node[midway,xshift=25pt, yshift=-60pt, font=\Huge]{Yes}}]
        ]
        [{Is a fixed distributional\\ family required?}, fill=Decision, text width=5.8cm, l sep = 45mm, 
          edge label={node[midway,xshift=25pt, yshift=-80pt, font=\Huge]{No}}
          [{Is the distribution\\ assumed to be unimodal?}, fill=Decision, text width=6.5cm, l sep = 55mm,
            edge label={node[midway,xshift=-25pt, yshift=-80pt,font=\Huge]{Yes}}
            [{Single Parametric\\ Distributions}, fill=Parametric, text width=4.4cm,
              edge label={node[midway,xshift=-25pt, yshift=-120pt,font=\Huge]{Yes}}]
            [{Mixture Density\\ Networks}, fill=Parametric, text width=4cm,
              edge label={node[midway,xshift=25pt, yshift=-120pt,font=\Huge]{No}}]
          ]
          [{Do you prefer a lightweight,\\ minimal-tuning architecture?}, fill=Decision, text width=7.2cm, l sep = 20mm,
            edge label={node[midway,xshift=25pt, yshift=-80pt,font=\Huge]{No}}
            [{Engression}, fill=DistReg, text width=2.8cm,
              edge label={node[midway,xshift=-25pt, yshift=-20pt,font=\Huge]{Yes}}
            ]
            [{Do you need to optimize the\\ coverage-width trade-off?}, fill=Decision, text width=7.2cm, l sep = 50mm,
              edge label={node[midway,xshift=25pt, yshift=-20pt,font=\Huge]{No}}
              [{Loss-Driven Interval Estimation}, fill=DistReg, text width=8cm,
                edge label={node[midway,xshift=-25pt, yshift=-100pt, font=\Huge]{Yes}}]
              [{Quantile Regression}, fill=DistReg, text width=5cm,
                edge label={node[midway,xshift=25pt, yshift=-100pt,font=\Huge]{No}}]
            ]
          ]
        ]
      ]
  ]
]
\end{forest}
}
\caption{Decision flowchart for identifying candidate probabilistic forecasting families based on model access, coverage requirements, structural assumptions, and computational constraints. Recommendations are indicative rather than exclusive. \textsuperscript{$\star$}Rarely recommended beyond simple, low-resource deployments (Section~\ref{sec_precontrol_limits}).}
\label{fig:model_guidance_flowchart}
\end{figure}
 
\subsubsection{Data-Driven Selection}\label{sec_model_guidance_data_driven}

Fig.~\ref{fig:model_guidance_flowchart} summarizes the practical considerations discussed throughout this survey into a decision flowchart for selecting an appropriate probabilistic forecasting method. The method-selection framework follows the distinctions in Section~\ref{sec_cross_paradigm_synthesis}. The first requirement is the decision-relevant output: selected intervals or quantiles may suffice for single-target decisions, whereas network-wide risk or scenario planning requires a joint dependence representation. The next requirement is a defensible basis for uncertainty: usable priors or process structure and interpretability requirements favor
explicit Bayesian modeling; an adequate likelihood favors parametric heads; specific distributional functionals favor quantile objectives; and complex joint futures motivate a structured or sample-based predictive law. Calibration is then considered as an additional stage, with its validity conditions checked against the data's dependence and distributional stability. Finally, fitting and inference budgets constrain the feasible implementation. The flowchart therefore identifies candidate mechanisms from the required output and assumptions, rather than selecting a universally best family from the established taxonomy.

\subsubsection{Domain-Specific Method Guidance}\label{sec_model_guidance_domain_specific}
\begin{figure}[!t]
    \centering
    \includegraphics[width=\linewidth]{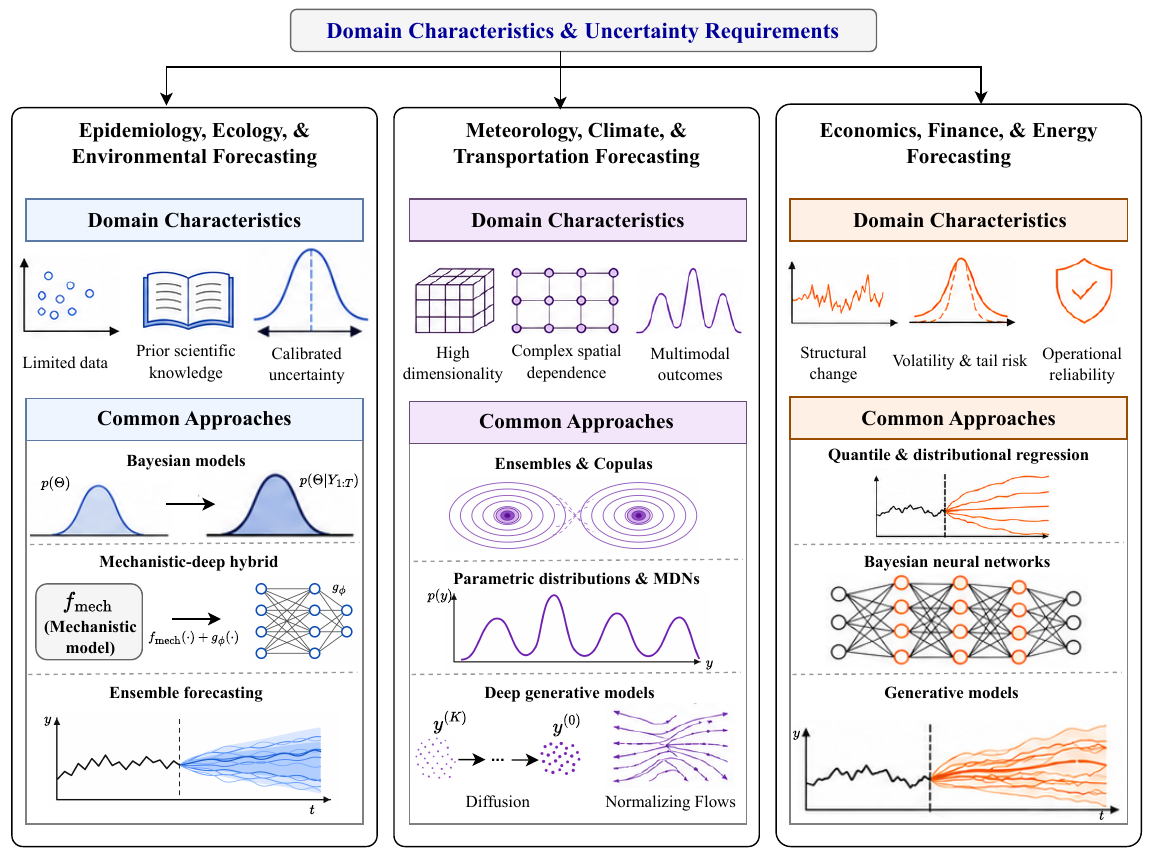}
    \caption{Overview of probabilistic forecasting applications across diverse domains alongside their primary methodological approaches. The associations are illustrative rather than exclusive.}
    \label{fig:domains_overview}
\end{figure}

The previous section discussed method selection based on statistical characteristics and modeling requirements. In practice, probabilistic forecasting is also shaped by domain-specific considerations, including the data-generating process, availability of prior knowledge, operational constraints, and regulatory requirements. These factors often determine whether forecasting systems prioritize calibrated prediction intervals, full predictive distributions, interpretability, or computational efficiency. This section identifies recurring methodological patterns across domains. Applications with limited observations, strong domain knowledge, or a need for interpretability tend to favor Bayesian formulations, whereas high-dimensional spatiotemporal systems increasingly rely on deep generative approaches to capture complex dependencies. 
Distribution-free calibration provides a complementary strategy for improving coverage across forecasting architectures; its theoretical validity depends on the assumptions of the chosen procedure, including exchangeability for standard split conformal methods and method-specific conditions for dependent data. A breakdown of these domains and their corresponding methodological frameworks is depicted in Fig.~\ref{fig:domains_overview}. 

Domains such as epidemiology, ecology, and environmental science prioritize interpretability  \citep{ferte2026epidemic, michael2017ecological, krapu2019environmental}. In particular, epidemic and public health forecasting requires balancing predictive performance with interpretability to support intervention strategies and policy decisions \citep{ferte2026epidemic}. Bayesian state-space and hierarchical formulations are, therefore, widely adopted to incorporate epidemiological knowledge through prior information \citep{karami2026comparative}. 
Recent approaches have further explored hybrid frameworks that integrate epidemiological principles into deep learning architectures to enhance forecasting performance \citep{barman2025egdl}. Under exchangeability, these methods can be combined with standard conformal prediction to obtain finite-sample marginal coverage; extensions for dependent data require additional method-specific assumptions. Similar considerations arise in ecological forecasting, where sparse observations and imperfect sampling processes motivate Bayesian models that separate latent ecological dynamics from observation uncertainty \citep{clark2005why,michael2017ecological}. Their ability to update predictions as new observations become available also aligns with iterative ecological forecasting practices \citep{michael2018iterative}. Environmental and hydrological forecasting extend these challenges to physical systems, where interpretability and uncertainty quantification are essential for supporting risk-sensitive decisions such as air quality and flood management \citep{krapu2019environmental, Papacharalampous2022hydrological}. Thus, Bayesian approaches are commonly employed for their principled uncertainty propagation and incorporation of domain knowledge \citep{krapu2019environmental,johnson2023bayesian}.  Other approaches combine deep learning models with distribution-free calibration methods  \citep{panja2025precipitation,panja2026estgcn}, while ensemble methods remain widely used in hydrology \citep{de2025beyond}. Recent advances have further explored flexible generative approaches, such as normalizing flows, for modeling complex predictive distributions in Earth system forecasting \citep{xu2024dyland}.

Forecasting domains dominated by large-scale spatiotemporal dynamics have increasingly adopted flexible probabilistic models capable of representing high-dimensional dependencies. Meteorological forecasting, for instance, relies on numerical weather prediction systems whose ensemble forecasts often suffer from miscalibration \citep{gneiting2005calibrated}. BMA and copula-based approaches have addressed this challenge by calibrating ensemble outputs and modeling dependencies between variables \citep{gneiting2014probabilistic}. Recent work, however, has shifted toward deep generative models for probabilistic spatiotemporal forecasting \citep{oskarsson2024probabilistic, price2024gencast} due to their flexibility in capturing high-dimensional, nonlinear dynamics. Similar challenges arise in transportation forecasting, where graph-structured dynamics and multimodal traffic states motivate the use of parametric distributional models such as Gaussian mixtures and MDNs \citep{zhi2019directional,chen2023prob}. Recent approaches have also explored loss-driven interval estimation and diffusion models for calibrated intervals and realistic probabilistic trajectory generation, respectively \citep{cheng2024recent}. Gaussian processes remain useful for smaller-scale mobility problems where uncertainty calibration outweighs scalability limitations \citep{steentoft2024quantifying}.

High-stakes forecasting in economic, financial, and energy markets requires adapting probabilistic architectures and calibration techniques to domain-specific constraints. Economic and macro-financial forecasting are characterized by limited historical observations, structural breaks, nonlinear dynamics, and a strong emphasis on tail risk \citep{elliott2006handbook}. Consequently, distributional methods such as quantile regression have gained widespread adoption \citep{adrian2019vulnerable,chavleishvili2024forecasting}. BNNs offer an alternative approach when uncertainty from model parameters and observations must be represented jointly \citep{blasco2024survey}, while conformal prediction provides a model-agnostic calibration framework for complex forecasting architectures \citep{sengupta2025fewnet, chakraborty2026narfima}. Energy forecasting shares these requirements, as weather variability and temporal dynamics make probabilistic modeling essential for managing financial risk. Distributional regression is frequently applied for load and price forecasting \citep{hirsch2025online}, while recent works have shifted towards generative approaches \citep{chen2025probabilistic, dudek2026recent}. 

\section{Empirical Demonstration}\label{sec_emp_demo}

This section provides an illustrative empirical comparison of representative probabilistic forecasting methods using point and probabilistic evaluation measures. 
Sections \ref{sec_demo_temporal} and \ref{sec_demo_spatiotemporal} detail the experimental results for the temporal and spatiotemporal settings, respectively, followed by a brief discussion on model performance in Section~\ref{sec_empirical_discussion}. Performance is evaluated across three standard public datasets, spanning distinct domains: the Air Passengers dataset \citep{herzen2022darts} for univariate forecasting, the PEMS-SF traffic dataset \citep{alexandrov2020gluonts} for multivariate analysis, and the Belgium COVID-19 dataset \citep{panja2026epicastbench} for spatiotemporal modeling. Table \ref{tab:models_demonstration} provides an overview mapping of the evaluated probabilistic forecasting models to their respective classes. Notably, spatiotemporal frameworks such as GpGp and GCEN require locational coordinates and spatial adjacency matrices; consequently, they are excluded from the temporal-only benchmarks due to the absence of spatial metadata. To reduce differences in the forecasting architecture within the model-agnostic subset, we fix LSTM \citep{hochreiter1997lstm} as the base model, and perform separate re-fitting for each method. To ensure a standardized comparison, all models employ the default hyperparameter configurations stipulated by their native libraries. For instance, the LSTM architectures utilize two layers with 64 hidden units. The BNN-MCD model is implemented using an LSTM network with active Monte Carlo dropout during inference \citep{hassan2024bitcoin}. The three foundation models, namely TimesFM-2.5 \citep{das2024timesfm}, Chronos-2 \citep{ansari2025chronos2}, and PatchTSTFM\footnote{\url{https://huggingface.co/ibm-research/patchtst-fm-r1}}, utilize quantile regression to generate probabilistic forecasts. All experiments are conducted on a single NVIDIA T4 GPU with 16GB of VRAM, and all reported training and inference times correspond to this hardware setup. Specifically, training time encompasses model fitting (or checkpoint loading for zero-shot foundation models) along with the calibration step for conformal methods, whereas inference time measures the wall-clock duration required to generate predictive samples from each fitted model. We report the probabilistic metrics listed in Section~\ref{sec_metrics} to compare the representative models. Additionally, we compute three point metrics, namely the Mean Absolute Error (MAE), Mean Absolute Scaled Error (MASE), and Root Mean Squared Scaled Error (RMSSE) \citep{hyndman2021forecasting} to assess the point forecast accuracy. Each dataset is evaluated at one terminal forecast origin, using a single fitted run with seed 42. Table~\ref{table:evaluation_protocol} provides a consolidated overview of the benchmarking protocol, detailing the context and prediction horizons, calibration window size, and training epochs. Each dataset is partitioned temporally into a training set and a held-out test set spanning the final prediction horizon. Across all models, we generate multi-step-ahead forecasts over the test horizon by drawing $100$ predictive trajectories from each fitted model to compute the evaluation metrics. Detailed hyperparameter configurations for each model are provided in the supplementary codes.

\begin{table}[!ht]
    \centering
    
    \caption{Benchmarking protocol across the three datasets.}
    \label{table:evaluation_protocol}
    \begin{adjustbox}{width=\textwidth}
    \begin{tabular}{cccccccccc}
        \toprule
        Setting & Dataset & Frequency & \# Time Points & \# Nodes & Pre-test/Test & \makecell{Context\\Length} & \makecell{Prediction\\Length} & \makecell{Calibration\\Size\textsuperscript{$\dagger$} (CP \& CQR)} & \# Epochs \\ 
        \midrule
        Univariate & Air Passengers & Monthly & 144 & 1 & 1-132 / 133-144 & 24 & 12 & 36 & 150 \\ 
        Multivariate & PEMS-SF Traffic & 1 hour & 4001 & 963 & 1-3977 / 3978-4001 & 24 & 24 & -- & 30 \\ 
        Spatiotemporal & Belgium COVID-19 & Daily & 776 & 11 & 1-746 / 747-776 & 60 & 30 & 90 & 100 \\ 
        \bottomrule
    \end{tabular}
    \end{adjustbox}
    \footnotesize{
\textsuperscript{$\dagger$}CP/CQR use the final pre-test observations for calibration, with preceding observations used for base-model fitting. \\
}
\end{table}

\begin{table}[!ht]
    \caption{Overview of the representative models used for empirical demonstration. `\textcolor{Blue}{\circmark}' indicates `partial support' for a specific data type (e.g., LSTM lacks built-in spatiotemporal capabilities but can still process these datasets by treating each node independently).}
    \begin{adjustbox}{width=\linewidth}
    \begin{tabular}{lccccc}
    \toprule
        Model & Class & Univariate & Multivariate & Spatiotemporal & Library \\ \midrule
        LSTM-PC & Pre-control limit & \textcolor{ForestGreen}{\ding{51}} & \textcolor{ForestGreen}{\ding{51}} & \textcolor{Blue}{\circmark} & \texttt{darts} \\ 
        LSTM-BE & Bootstrap ensemble & \textcolor{ForestGreen}{\ding{51}} & \textcolor{ForestGreen}{\ding{51}} & \textcolor{Blue}{\circmark} & \texttt{darts} \\ 
        LSTM-CP & Conformal prediction & \textcolor{ForestGreen}{\ding{51}} & \textcolor{ForestGreen}{\ding{51}} & \textcolor{Blue}{\circmark} & \texttt{darts} \\ 
        LSTM-CQR & Conformalized quantile regression & \textcolor{ForestGreen}{\ding{51}} & \textcolor{ForestGreen}{\ding{51}} & \textcolor{Blue}{\circmark} & \texttt{darts} \\ 
        BNN-MCD & Bayesian Monte Carlo dropout & \textcolor{ForestGreen}{\ding{51}} & \textcolor{ForestGreen}{\ding{51}} & \textcolor{Blue}{\circmark} & \texttt{torch} \\ 
        GpGp & Gaussian process & \textcolor{red}{\ding{55}} & \textcolor{red}{\ding{55}} & \textcolor{ForestGreen}{\ding{51}} & \texttt{gpgp} (R) \\ 
        DeepAR & Single parametric distribution & \textcolor{ForestGreen}{\ding{51}} & \textcolor{ForestGreen}{\ding{51}} & \textcolor{Blue}{\circmark} & \texttt{darts} \\ 
        DeepNPTS & Nonparametric distribution & \textcolor{ForestGreen}{\ding{51}} & \textcolor{ForestGreen}{\ding{51}} & \textcolor{Blue}{\circmark} & \texttt{neuralforecast} \\ 
        LSTM-QR & Quantile regression & \textcolor{ForestGreen}{\ding{51}} & \textcolor{ForestGreen}{\ding{51}} & \textcolor{Blue}{\circmark} & \texttt{darts} \\ 
        MVEN & Engression & \textcolor{ForestGreen}{\ding{51}} & \textcolor{ForestGreen}{\ding{51}} & \textcolor{Blue}{\circmark} & \texttt{stengression} \\ 
        GCEN & Engression & \textcolor{red}{\ding{55}} & \textcolor{red}{\ding{55}} & \textcolor{ForestGreen}{\ding{51}} & \texttt{stengression} \\ 
        K2VAE & VAE & \textcolor{ForestGreen}{\ding{51}} & \textcolor{ForestGreen}{\ding{51}} & \textcolor{Blue}{\circmark} & K2VAE\textsuperscript{$\dagger$} \\ 
        Transformer-MAF & Normalizing flows & \textcolor{red}{\ding{55}} & \textcolor{ForestGreen}{\ding{51}} & \textcolor{Blue}{\circmark} & \texttt{pytorch-ts} \\ 
        TimeGrad & Diffusion & \textcolor{red}{\ding{55}} & \textcolor{ForestGreen}{\ding{51}} & \textcolor{Blue}{\circmark} & \texttt{pytorch-ts} \\ 
        TimesFM-2.5 & Foundation (quantile regression) & \textcolor{ForestGreen}{\ding{51}} & \textcolor{ForestGreen}{\ding{51}} & \textcolor{Blue}{\circmark} & \texttt{darts} \\ 
        Chronos-2 & Foundation (quantile regression) & \textcolor{ForestGreen}{\ding{51}} & \textcolor{ForestGreen}{\ding{51}} & \textcolor{Blue}{\circmark} & \texttt{darts} \\ 
        PatchTSTFM & Foundation (quantile regression) & \textcolor{ForestGreen}{\ding{51}} & \textcolor{ForestGreen}{\ding{51}} & \textcolor{Blue}{\circmark} & \texttt{darts} \\ \bottomrule
    \end{tabular}
    \end{adjustbox}
    \footnotesize{
\textsuperscript{$\dagger$}\url{https://github.com/decisionintelligence/K2VAE}\\
}

    \label{tab:models_demonstration}
\end{table}

\subsection{Temporal Forecasting}\label{sec_demo_temporal}
In the univariate setting, we evaluate the compatible models outlined in Table~\ref{tab:models_demonstration} on the monthly Air Passengers dataset \citep{herzen2022darts}. All task-specific models are trained for 150 epochs using a standardized context window of 24 months to forecast a 12-month horizon; the results across evaluation metrics are detailed in Table~\ref{tab:results_airpassengers_data}. The evaluated \texttt{pytorch-ts}\footnote{\url{https://github.com/zalandoresearch/pytorch-ts}} implementations of Transformer-MAF (and TimeGrad) were omitted from the univariate experiment because they require a multivariate target representation. 
To evaluate probabilistic forecasting methods on multivariate temporal data, we use the PEMS-SF traffic dataset \citep{alexandrov2020gluonts}, which contains road occupancy rates for 4001 time points from 963 highway sensors in the San Francisco Bay Area. All compatible task-specific models are trained for 30 epochs under default hyperparameter configurations. Due to the dataset's high dimensionality, conformal prediction-based approaches, namely LSTM-CP and LSTM-CQR, are excluded because of memory constraints that result in out-of-memory failures during training. The resulting performance comparison is reported in Table~\ref{tab:results_pems_data}.

\subsection{Spatiotemporal Forecasting}\label{sec_demo_spatiotemporal}
\begin{figure}[!t]
    \centering
    \includegraphics[width=0.9\linewidth]{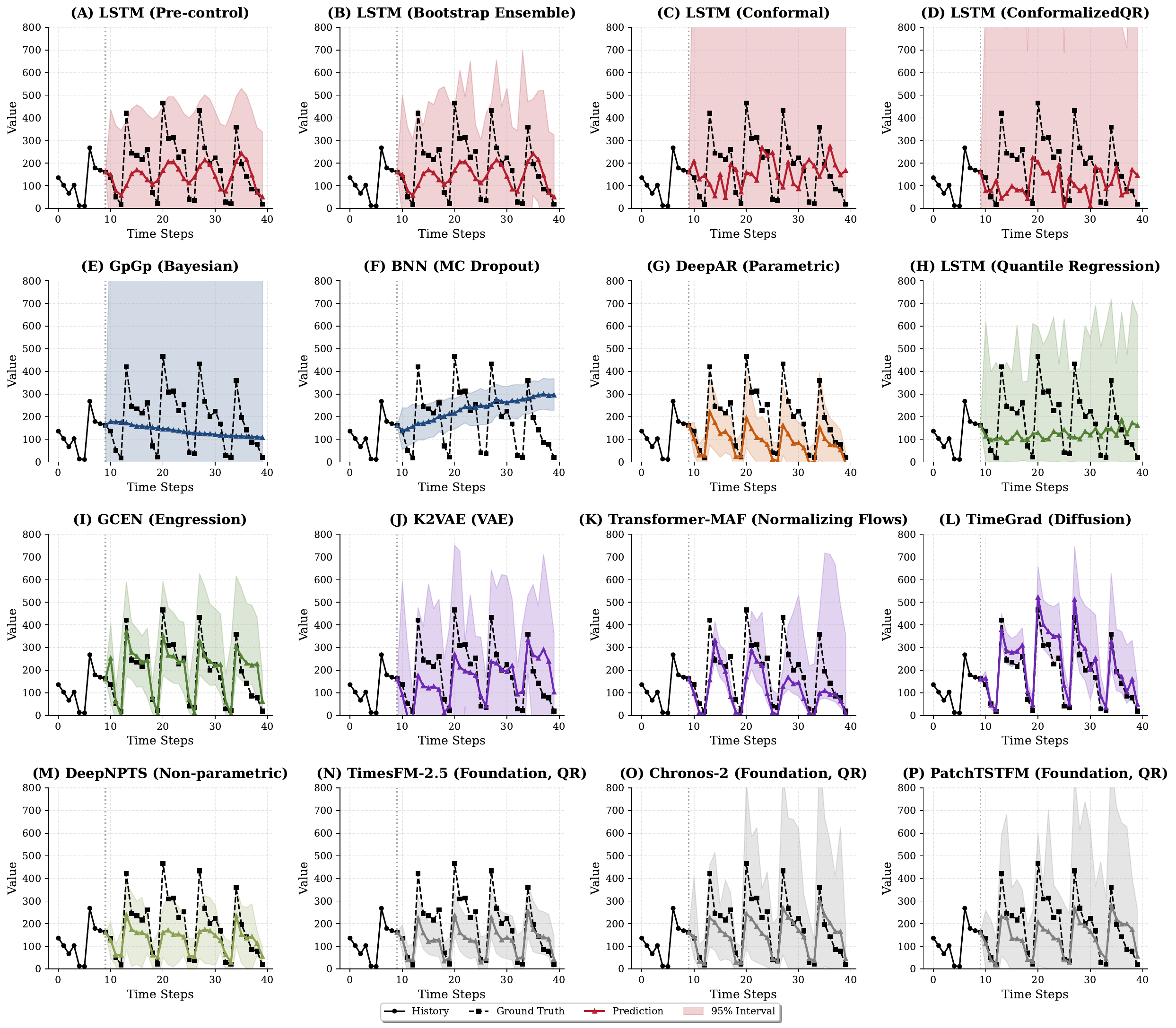}
    \caption{Forecasts and prediction intervals from representative models obtained on COVID-19 cases in Limburg, Belgium. The point predictions and prediction intervals are shown with a representative color in the legend; actuals are color-coded following the taxonomy in Table~\ref{tab:UQ_Methods_Summary}.}
    \label{fig:belgium_benchmark}
\end{figure}

To evaluate model performance on spatiotemporal data, we analyze daily COVID-19 case counts across 11 Belgian provinces from September 2020 to October 2022. Following the methodology in \cite{panja2026estgcn}, the spatial adjacency matrix is constructed by applying a Gaussian kernel to the pairwise Haversine distances between locations. For a standardized comparison, all models are configured with a 60-day context window to forecast across a 30-day prediction horizon. Quantitative evaluation metrics are summarized in Table~\ref{tab:results_belgium_data}, while Fig.~\ref{fig:belgium_benchmark} visualizes the ground truth against the out-of-sample point forecasts and their corresponding $95\%$ prediction intervals.

\subsection{Analysis of Results}\label{sec_empirical_discussion}
\begin{figure}[!t]
\centering
\includegraphics[width=\linewidth]{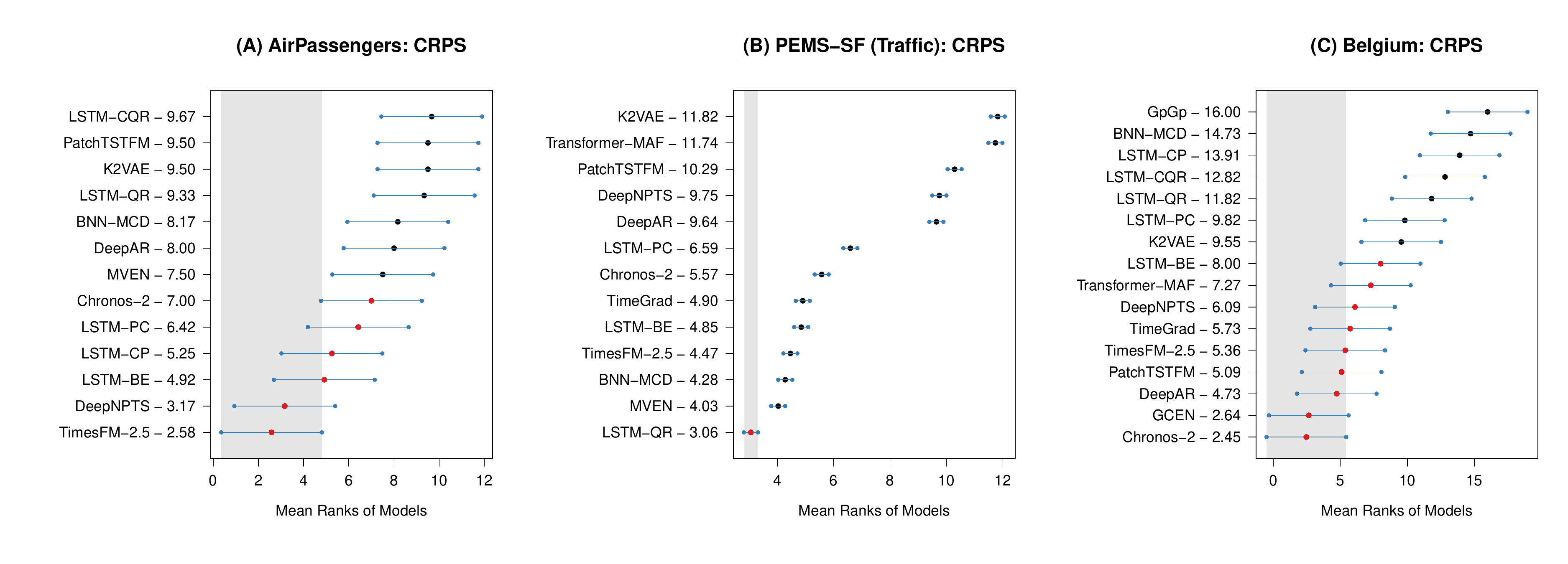}
\caption{MCB test results for the empirical evaluation based on CRPS for (A) Univariate Air Passengers, (B) Multivariate PEMS-SF, and (C) Spatiotemporal Belgium COVID-19 datasets.}
\label{fig:MCB_results}
\end{figure}

This section evaluates the predictive performance of the assessed representative models across the three diverse datasets. 
Following Section~\ref{sec_robustness}, we conduct MCB Nemenyi tests on CRPS at an $80\%$ confidence level ($\alpha = 0.20$). 
Fig.~\ref{fig:MCB_results} visualizes the mean CRPS ranks across models. For the multivariate datasets (PEMS-SF and Belgium), CRPS is first averaged over the forecast horizon, and models are then ranked across the spatial nodes. For the univariate Air Passengers series, models are ranked directly across the individual forecast horizon steps. The test treats these ranking units (nodes or horizon steps) as exchangeable blocks without explicit adjustments for residual spatial or serial correlation. Across these experiments, no single model performs best on every dataset, performance indicator, calibration, and computational cost. Given the limited number of datasets and the use of a single run per setting, the results can be interpreted as illustrative comparisons rather than general rankings of the methods. 


\begin{figure}[!htbp]
\centering
\includegraphics[width=0.7\linewidth]{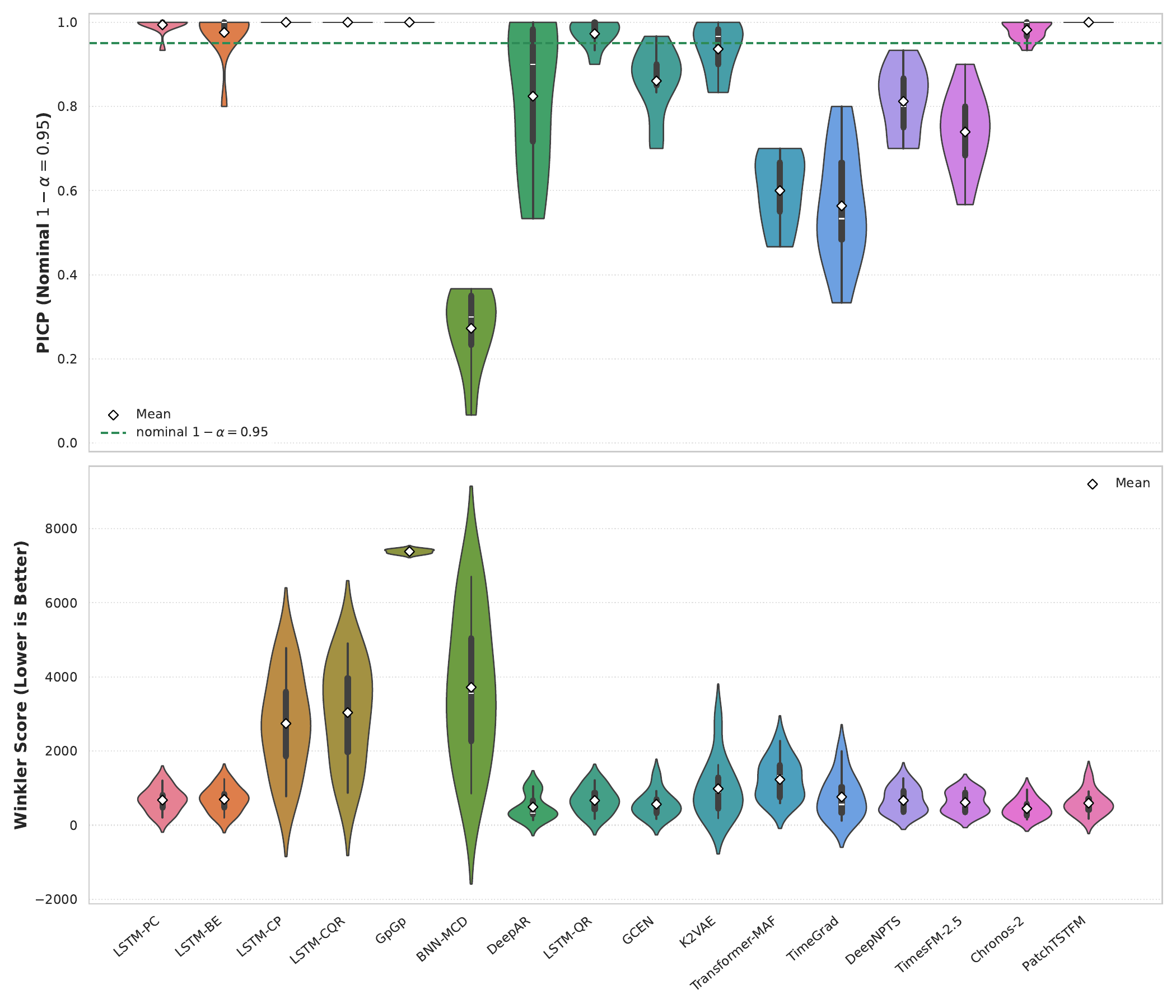}
\caption{PICP (top) and Winkler score (bottom) attained by the models on the Belgium COVID-19 dataset across 11 spatial nodes.}
\label{fig:coverage_violin_plot}
\end{figure}

\begin{figure}[!ht]
\centering
\includegraphics[width=\linewidth]{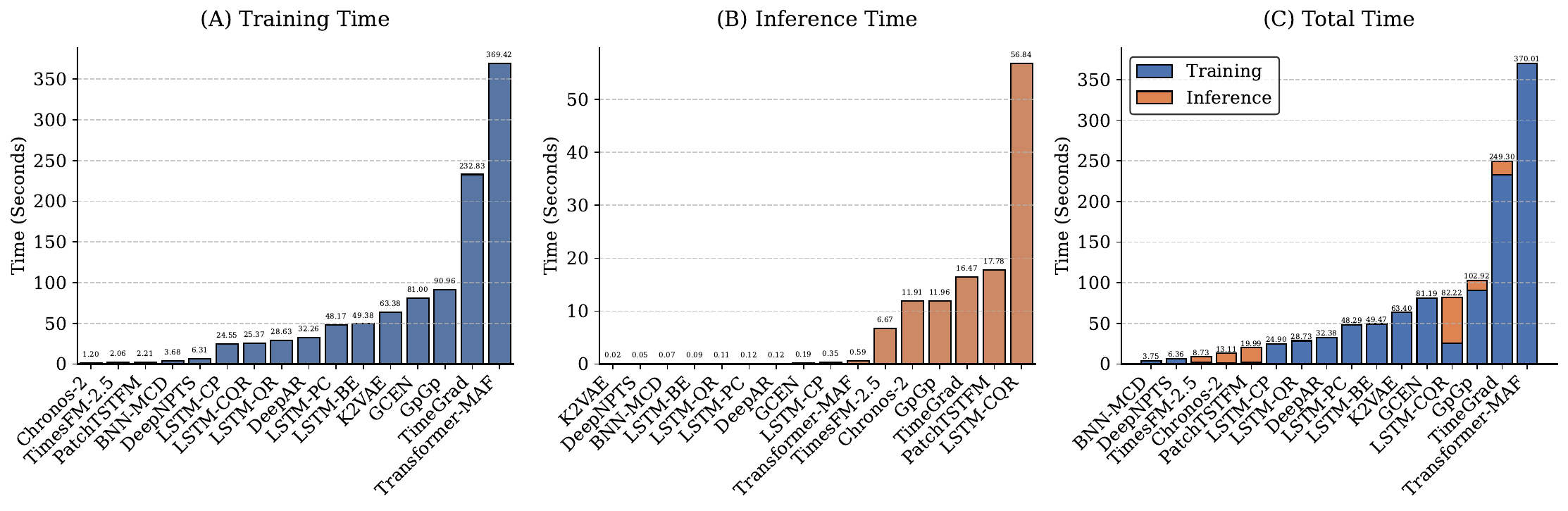}
\caption{(A) Training time, (B) inference time, and (C) total time taken by the representative models on the spatiotemporal Belgium COVID-19 dataset.}
\label{fig:training_inference_times}
\end{figure}

For the univariate Air Passengers dataset, TimesFM-2.5 demonstrates superior performance in terms of probabilistic metrics. Specifically, it obtains the best scores on CRPS, Winkler score, and pinball loss, achieving the best average rank in the MCB test. Obtaining the best Winkler score alongside near-nominal 91.67\% empirical coverage (PICP) demonstrates that TimesFM-2.5 generated robust predictive intervals for this univariate data. Following similar arguments as \cite{pathak2026engression}, we note that PICP alone can be a misleading indicator of prediction interval quality. Models can artificially achieve full coverage by generating excessively wide, uninformative intervals that lack practical utility. Consequently, the Winkler score serves as a more reliable metric by jointly penalizing both undercoverage and excessive interval width. For instance, K2VAE and LSTM-CQR on the Air Passengers data and GpGp on the Belgium COVID-19 data both achieve full empirical coverage, yet they incur excessively high Winkler scores, rendering their intervals suboptimal for decision-making. Applying vanilla conformal prediction to spatiotemporal data can yield similarly uninformative bounds. Dependence and nonstationarity can make historical calibration scores
unrepresentative of future errors, and failure of exchangeability removes the standard split-conformal coverage guarantee, which may lead to under- or over-coverage empirically. We observe that on the Belgium COVID-19 dataset (Fig.~\ref{fig:coverage_violin_plot}), LSTM-CP and LSTM-CQR achieve 100\% PICP at the cost of excessively high Winkler scores, indicating overly wide intervals.

The \texttt{pytorch-ts} implementation of generative methods leveraging normalizing flows (Transformer-MAF) and diffusion processes (TimeGrad) lack native support for univariate data. Moreover, as detailed in Tables~\ref{tab:results_pems_data}-\ref{tab:results_belgium_data} and Fig.~\ref{fig:training_inference_times}, these models incur substantial computational overhead during both training and inference compared to their competitors. 
Engression exhibits a different computational trade-off, as its resource demands depend directly on the backbone architecture. For example, MVEN with an LSTM backbone records the shortest inference time but the longest training duration on PEMS-SF, whereas GCEN shows more balanced training and inference runtimes on the Belgium COVID-19 dataset. On the latter, although GCEN achieves the best average CRPS across all spatial nodes, it secures only the second-best rank in the MCB test, trailing Chronos-2. 
Because arithmetic averages are highly susceptible to skewness from extreme outliers, a model might achieve a superior overall mean score by significantly outperforming on a few anomalous nodes, even if rank-based tests indicate less consistent performance across the broader dataset. Consequently, Chronos-2 proves more robust, consistently yielding lower CRPS values across a larger subset of spatial nodes compared to GCEN.

Finally, foundation models such as Chronos-2 and TimesFM-2.5 offer highly competitive predictive performance. An analysis of computational efficiency on the Belgium COVID-19 dataset (Fig.~\ref{fig:training_inference_times}) reveals significant architectural trade-offs. For the zero-shot foundation models, the small training time entries measure local checkpoint loading and wrapper setup rather than pretraining or dataset-specific gradient training. Their inference cost remains dependent on the evaluated workload and implementation. In contrast, deep generative architectures demand substantial computational resources; the normalizing-flow-based Transformer-MAF requires the most training time, followed by the diffusion-based TimeGrad. However, their inference behaviors diverge, as TimeGrad requires considerably more inference time than Transformer-MAF, which is highly efficient once trained. These operational profiles emphasize the necessary compromise between generative complexity and deployment speed in probabilistic forecasting tasks.

\section{Challenges and Future Directions}\label{sec_challenges}

The agenda below follows from the cross-paradigm comparison in
Section~\ref{sec_cross_paradigm_synthesis} and Table~\ref{tab:cross_paradigm_synthesis}, together with the evaluation
requirements in Section~\ref{sec_metrics} and the trade-offs in
Section~\ref{sec_model_guidance}. The central gaps concern uncertain and evolving dependence, predictive laws compatible with domain structure, reliable estimation of rare events, and feasible validation and deployment. Existing methods address important parts of these problems; the opportunities concern
what remains unresolved when these requirements must hold together.
Section~\ref{sec_open_challenges} pairs each limitation with a research objective and discusses open challenges in probabilistic forecasting. Section~\ref{sec_future_directions} considers foundation, adaptive, and agentic systems as possible future directions.


\subsection{Open Challenges}\label{sec_open_challenges}

\textbf{Uncertainty in dependency structures.} Section~\ref{sec_cross_paradigm_synthesis} distinguishes joint predictive laws from marginal outputs. However, joint forecasts
conditional on a fitted dependence structure need not integrate uncertainty in that structure. Most spatiotemporal forecasting methods assume that spatial dependencies are known or can be estimated deterministically. Whether represented through predefined graphs \citep{kipf2017gcn, velickovic2018graphattention}, adjacency matrices \citep{barman2025egdl,panja2026estgcn}, or covariance functions \citep{lindgren2011approximate}, these structures are treated as fixed during prediction, despite being inferred from incomplete and evolving observations. Consequently, current probabilistic forecasts account for uncertainty in model parameters or observations, but rarely quantify uncertainty in the dependency structure itself. Adaptive graph learning methods such as adaptive graph convolutional recurrent networks \citep{bai2020gcrn} alleviate the need for predefined connectivity by learning spatial dependencies directly from data. However, the learned dependency structure is treated as deterministic and uncertainty in these inferred relationships is not propagated into the predictive distribution. Developing probabilistic forecasting models that jointly quantify uncertainty in both spatial dependencies and future trajectories, therefore, remains an important research direction.

\textbf{Physics-informed probabilistic forecasting.} 
Many spatiotemporal systems are governed by known physical or mechanistic principles, including conservation laws, epidemiological dynamics, and power-system constraints. Recent methods enforce these constraints through domain-guided graph structures \citep{wang2026physics} or hybrid mechanistic-deep learning architectures \citep{barman2025egdl}. Although these approaches demonstrate the value of incorporating domain knowledge into forecasting models, domain constraints do not extend to the uncertainty quantification itself. Instead, predictive intervals are produced independently through post-hoc calibration. Embedding physical dynamics into the predictive distribution or calibration mechanism to enforce domain-constrained predictive bounds remains an open challenge.

\textbf{Forecasting extremes.} Reliable uncertainty estimation for rare events remains one of the most difficult problems in probabilistic forecasting. As highlighted in Section~\ref{sec_cross_paradigm_synthesis}, standard approaches frequently fail to capture these extremes due to restrictive parametric tails, finite quantile resolution, or an inability to resolve rare modes from sparse data. Extreme observations are scarce, causing existing probabilistic objectives to primarily fit the central part of the distribution, with extremes contributing minimally to the optimization signal. EVT provides a principled statistical framework for modeling distributional tails \citep{coles2001extremes} and several recent methods have incorporated these ideas through bulk-tail likelihoods for spatial processes \citep{castro2019spliced}, extreme-value-aware graph convolutional forecasting networks with conformal calibration \citep{panja2026estgcn}, quantile regression informed by EVT \citep{pasche2024eqrn}, and normalizing flows designed for multivariate tail dependence \citep{demonte2025generative}. Despite these advances, current approaches address different aspects of the problem in isolation. Simultaneously modeling temporal evolution, spatial dependence, and multivariate tail behavior at realistic spatial scales remains computationally demanding and evaluation is severely constrained by the scarcity of extreme events available for calibration and validation.

\textbf{Zero-inflated and count-valued spatiotemporal forecasting.} The representation restrictions in Table~\ref{tab:cross_paradigm_synthesis} have direct consequences for count data -- a continuous density
alone assigns no positive mass to an exact zero. Many spatiotemporal forecasting applications, including epidemiological surveillance \citep{francisco2024hybrid} and crime prediction \citep{wang2025stmgnnzinb}, involve count-valued observations with a high proportion of zeros arising from absence or infrequent event occurrence. These characteristics violate the assumptions underlying many probabilistic forecasting methods, which predominantly model continuous-valued targets; as a result, accurately representing the probability of observing zero events and the distribution of nonzero counts remains challenging \citep{wilson2022beyond}. Although zero-inflated count models are well established in statistical modeling, their integration with modern probabilistic forecasting architectures remains limited. For example, \cite{liu2025thompson} incorporated Poisson, negative binomial, and zero-inflated likelihoods into an uncertainty-aware online learning framework, producing calibrated posterior estimates under overdispersion and excess zeros. However, 
it models each time series independently and does not capture dependencies among interacting entities that may be represented through graph structures. Conversely, \cite{francisco2024hybrid} applied a two-stage hurdle model for zero-inflated dengue forecasting across fine spatial resolutions, albeit their approach targets point prediction. Recent work has begun to 
shift from point to probabilistic forecasting of zero-inflated spatiotemporal data. For instance, \cite{wilson2022beyond} extended the hurdle model with a generalized Pareto tail component to jointly capture zero-inflation and heavy-tailed extremes in a single spatiotemporal framework, while \cite{wang2025stmgnnzinb} combined diffusion graph convolutions with a zero-inflated negative binomial likelihood for spatial crime forecasting.
However, these approaches are restricted to regular grids with fixed spatial adjacency. Extending zero-inflated models to irregular spatiotemporal networks remains largely unexplored, particularly when both multi-horizon temporal dependencies and spatial structures must be jointly estimated at scale.

\textbf{Forecasting directional time series.} Directional or circular time series consist of sequential observations represented as angles, degrees, or compass headings. Their periodic topology, with values wrapping around at $360^\circ$, violates the Euclidean assumptions of standard forecasting models and requires methods that account for circular structure. 
Such data arises naturally across many scientific domains, such as spatiotemporal wind directions, ocean current vectors, and animal migration bearings. Although circular data have been studied extensively in temporal settings \citep{fisher1994time, di2012non, harvey2024modelling} and directional fields in spatiotemporal settings \citep{mastrantonio2016spatio}, existing approaches have focused predominantly on point forecasting. 
Recent work has introduced uncertainty quantification for circular data in regression settings \citep{prokudin2018deep, pathak2026angle}, but reliable probabilistic forecasting for complex spatiotemporal directional processes remains a significant open challenge. 

\textbf{Continuous-time probabilistic forecasting.} Many real-world temporal and spatiotemporal processes evolve continuously and are observed irregularly, including mobile-health signals such as ECG and wearable measurements \citep{li2026hearts}, financial transactions, network events, traffic systems, and environmental sensor networks. In these examples, asynchronous sampling, missing observations, and irregular spatial measurements challenge conventional discrete-time forecasting models \citep{liu2021event}. Neural ODEs and Neural SDEs offer a natural framework for such data by representing latent dynamics in continuous time, with the latter additionally modeling stochastic fluctuations through a diffusion process \citep{chen2018neural, oh2024stable}. However, existing continuous-time learning methods have been studied predominantly from the perspective of point prediction or latent-state reconstruction \citep{wan2025earth}, while reliable probabilistic forecasting under irregular sampling and evolving temporal or spatial dependencies remains comparatively underdeveloped. 
An important open challenge is therefore to develop scalable and interpretable continuous-time probabilistic models that jointly accommodate irregular observations, missingness, multivariate dependence, and dynamic spatial interactions in real-world applications.

\textbf{Lack of a comprehensive software package.} Software tools for probabilistic forecasting remain divided, especially for spatiotemporal settings. As detailed in Section~\ref{sec_datasets_lib}, most existing libraries focus on temporal data. Spatiotemporal approaches can be implemented using the PyTorch-Geometric-Temporal \citep{rozemberczki2021pytorch} and Torch Spatiotemporal (\texttt{tsl})\footnote{\url{https://github.com/TorchSpatiotemporal/tsl}} libraries in Python; however, their capabilities are restricted to point forecasting. Probabilistic spatiotemporal methods are typically released as research repositories with incompatible interfaces. Hence, no single library implements all probabilistic forecasting methods within a unified framework.

\subsection{Future Directions}\label{sec_future_directions}

In this section, we highlight three promising directions for future research. One prominent trend is the emergence of foundation models for forecasting (Section~\ref{sec_foundation_models}). The dependence, support, and transfer limitations identified in
Section~\ref{sec_cross_paradigm_synthesis} motivate foundation models whose uncertainty representations can be adapted and validated across new domains.
\cite{adler2025calibration} showed that many of these models produce well-calibrated probabilistic distributions in zero-shot settings, although calibration quality varies considerably across architectures. Similar ideas are now being extended to spatiotemporal forecasting. OpenCity \citep{li2025open}, for example, combines Transformer and GNN architectures to learn transferable representations that generalize across unseen cities, demonstrating the potential of foundation models for large-scale spatiotemporal forecasting. Recent evidence also suggests that foundation models can achieve strong point-forecasting performance on zero-inflated epidemiological datasets \citep{panja2026epicastbench}; however, whether these models can produce equally well-calibrated probabilistic forecasts for zero-inflated spatiotemporal count data remains an open question. A key research direction is extending foundation models to produce calibrated probabilistic forecasts across spatial domains under distribution shift at a feasible cost for real-world deployment. 

Sensitivity to nonstationarity in residual models, dependence structures, and calibration scores motivates adaptation of predictive
spread, tails, and calibration as well as model weights. An emerging direction is the development of adaptive probabilistic forecasting systems. Such systems are particularly valuable in nonstationary environments, where shifts in data distributions or operating conditions can rapidly degrade predictive performance. Recent work has explored adaptive ensemble strategies based on multi-armed bandit algorithms, including upper confidence bound (UCB) and the exponential-weight algorithm for exploration and exploitation (EXP3), which dynamically reweight candidate forecasting models according to their recent performance \citep{karami2026adaptive}. Although these approaches have demonstrated improved probabilistic forecasting in applications such as COVID-19 prediction, they primarily adapt the combination of existing forecasts without necessarily updating each component's uncertainty mechanism. Extending these ideas toward fully adaptive probabilistic forecasting frameworks in an online manner represents a promising research direction, particularly for spatiotemporal applications.

A third emerging direction is agentic probabilistic forecasting. Delegating the forecasting pipeline to autonomous agents makes the heterogeneous outputs and validation requirements identified in Section~\ref{sec_cross_paradigm_synthesis} highly consequential, as these systems must dynamically navigate differing structural assumptions. Recent agentic frameworks move beyond a single forecasting model and instead treat forecasting as a sequential decision process involving data diagnostics, preprocessing, model selection, validation, ensembling, contextual reasoning, and forecast synthesis. For example, TimeSeriesScientist \citep{zhao2025timeseriesscientist} employs specialized agents to automate the end-to-end forecasting workflow, whereas Nexus \citep{das2026nexus} combines multimodal contextual information with macro- and micro-level reasoning agents and an error-driven calibration mechanism. 
However, uncertainty quantification in such agentic systems remains largely unexplored. Existing frameworks primarily optimize and evaluate point-forecast accuracy, while uncertainty may arise not only from the future observations themselves, but also from model selection, external information, tool use, reasoning trajectories, and disagreement among agents. A promising direction is therefore to develop agentic forecasting systems that propagate these multiple sources of uncertainty into calibrated predictive distributions. 
Such systems could combine probabilistic forecasting models, conformal calibration, or generative sampling with agent-level reasoning and adaptive model selection, enabling agents to quantify when forecasts are reliable, recognize when uncertainty has increased, and revise their forecasting strategy accordingly. Extending this paradigm to spatiotemporal forecasting, where agents must additionally reason about uncertain and evolving spatial dependencies, represents an important open frontier.

\section{Conclusion}\label{sec_conclusion}

This survey provides a unified perspective on probabilistic forecasting for time series and spatiotemporal data by organizing existing methods according to how predictive uncertainty is represented and quantified. The resulting taxonomy discusses model-agnostic and model-intrinsic approaches. Beyond cataloging methods, our synthesis highlights a central message: no single probabilistic forecasting paradigm is universally preferable. 
The illustrative empirical study shows that different evaluated
implementations lead on different metrics in the selected test blocks,
consistent with the need to match uncertainty representations,
calibration requirements, and computational budgets to the task.

The field is now moving beyond the construction of prediction intervals around point forecasts toward richer predictive systems that must represent uncertainty over complex, evolving, and structured futures. At the same time, foundation models, adaptive forecasting systems, continuous-time probabilistic models, and emerging agentic forecasting frameworks are creating new opportunities and new sources of uncertainty that must themselves be quantified and calibrated. Progress will therefore require not only more expressive forecasting architectures, but also principled mechanisms for representing and propagating
uncertainty throughout the forecasting pipeline. We hope that the taxonomy, practical guidance, and research agenda developed in this survey provide a foundation for both informed methodological choice and the next generation of reliable probabilistic forecasting systems.


\bibliography{Survey}
\bibliographystyle{tmlr}

\clearpage
\appendix
\setcounter{table}{0}
\renewcommand{\thetable}{A.\arabic{table}}
\section*{Appendix}

\begin{table}[!ht]
    \raggedleft
    \caption{Publicly available datasets widely used for benchmarking in the forecasting literature. ``\#Series/Nodes'' indicates the number of time series excluding external covariates. ``Analysis \& Benchmarks'' consists of works that have analyzed and/or compared model performances on the respective datasets.}
    
    \newcounter{tabnotecounter}
    \setcounter{tabnotecounter}{0}
    \renewcommand{\thetabnotecounter}{\alph{tabnotecounter}}
    \newcommand{\newtabnote}[1]{\refstepcounter{tabnotecounter}\label{#1}\tnote{\thetabnotecounter}}
    \newcommand{\usetabnote}[1]{\tnote{\ref{#1}}}

    \begin{adjustbox}{width=\linewidth}
    \begin{threeparttable}
    \begin{tabular}{ccccc p{5.5cm} cc}
    \toprule
        \multicolumn{1}{c}{Dataset} & \multicolumn{1}{c}{Type} & \multicolumn{1}{c}{Domain} & \multicolumn{1}{c}{\#Series/Nodes} & \multicolumn{1}{c}{Frequency} & \multicolumn{1}{c}{Description} & \multicolumn{1}{c}{Source} & \multicolumn{1}{c}{Analysis \& Benchmarks} \\ \midrule

        Air Passengers & Univariate & Transportation & 1 & Monthly & Monthly air passengers dataset, from 1949 to 1960. & \newtabnote{tn:darts} & -\\

        Australian Beer & Univariate & Production & 1 & Quarterly & Total quarterly beer production in Australia from 1956 to 2008. & \usetabnote{tn:darts} & - \\

        Monthly Milk & Univariate & Production & 1 & Monthly & Milk production (in pounds per cow) between Jan 1962 and Dec 1975. & \usetabnote{tn:darts} & -\\

        Heart Rate & Univariate & Health & 1 & 0.5-second & 1800 evenly-spaced measurements of heart rate from a single subject. & \newtabnote{tn:ecg} & \cite{glass2012theory} \\

        \midrule
        Hospital & Multivariate & Health & 767 & Monthly & Monthly time series showing patient counts related to medical products. & \cite{expsmooth_manual} & \cite{montero2021principles}\\
        
        Solar & Multivariate & Energy & 137 & Hourly & Photovoltaic production of 137 stations in Alabama State. & \newtabnote{tn:solar} & \makecell{\cite{salinas2019gpcopula, fan2024mgtsd}; \\ \cite{pathak2026entransformer}} \\ 
        
        Electricity & Multivariate & Energy & 370 & 15-min & Time series of the electricity consumption of 370 customers. & \newtabnote{tn:elec} & \makecell{\cite{salinas2019gpcopula, fan2024mgtsd}; \\ \cite{pathak2026entransformer}} \\ 

        \makecell{London Smart\\Meters} & Multivariate & Energy & 5560 & 30-min &  Energy consumption readings of London households in kWh. & \newtabnote{tn:london} & \makecell{\cite{nugaliyadde2019predicting};\\\cite{kaur2019smart}}\\
        
        PEMS-SF & Multivariate & Transportation & 963 & Hourly & Occupancy rate between 0 and 1, of 963 San Francisco car lanes. & \newtabnote{tn:pemssf} & \makecell{\cite{salinas2019gpcopula, fan2024mgtsd}; \\ \cite{pathak2026entransformer}} \\ 
        
        Taxi & Multivariate & Transportation & 1214 & 30-min & Traffic time series of New York taxi rides taken at 1214 locations. & \newtabnote{tn:taxi} & \makecell{\cite{salinas2019gpcopula, fan2024mgtsd}; \\ \cite{pathak2026entransformer}} \\ 
        
        KDD-cup & Multivariate & Environmental & 270 & Hourly & Air-quality observations from monitoring stations in Beijing and London. & \newtabnote{tn:kdd} & \cite{fan2024mgtsd, pathak2026entransformer} \\ 
        
        Wikipedia & Multivariate & Behavioral & 2000 & Daily & Daily page view counts for a set of Wikipedia articles. & \newtabnote{tn:wiki} & \makecell{\cite{salinas2019gpcopula, fan2024mgtsd}; \\ \cite{pathak2026entransformer}} \\ 
        
        Exchange rate & Multivariate & Finance & 8 & Daily & Daily exchange rate between 8 currencies. & \newtabnote{tn:exchange} & \cite{salinas2019gpcopula} \\ 
        
        \midrule
        
        Human3.6M & Spatiotemporal & Computer Vision & - & 50 Hz & Large-scale motion capture dataset with 3.6 million video frames. & \cite{ionescu2013human3} & \cite{tang2021probabilistic} \\ 
        
        HumanEva-I & Spatiotemporal & Computer Vision & 3 & 60 Hz & Human motion dataset comprising 3 subjects recorded at 60 Hz. & \cite{sigal2010humaneva} & \cite{tang2021probabilistic} \\ 

        METR-LA & Spatiotemporal & Transportation & 207 & 5-min & Traffic readings from 207 loop detectors on highways in Los Angeles County. & \cite{jagadish2014big} & \cite{li2018diffusion, jiang2023spatio} \\

        PEMS-BAY & Spatiotemporal & Transportation & 325 & 5-min & Traffic readings from 325 traffic sensors in San Francisco Bay Area. & \newtabnote{tn:pemsbay} & \cite{li2018diffusion, jiang2023spatio} \\

        \makecell{PEMS 03, 04,\\07, 08} & Spatiotemporal & Transportation & \makecell{358, 307,\\883, 170} & 5-min & Traffic readings from sensors provided by CalTrans PeMS. & \usetabnote{tn:pemsbay} & \cite{song2020spatial, guo2021learning} \\

        LargeST & Spatiotemporal & Transportation & 8600 & 5-min & Large-scale traffic forecasting data from 8600 traffic sensors in California. & \cite{liu2023largest} & \cite{liu2023largest}  \\

        Japan TB & Spatiotemporal & Epidemiology & 47 & Monthly & Monthly tuberculosis cases from 47 prefectures in Japan. & \cite{sumi2019time} & \makecell{\cite{barman2025egdl};\\ \cite{pathak2026engression}} \\

        China TB & Spatiotemporal & Epidemiology & 31 & Monthly & Monthly tuberculosis cases from 31 provinces in China. & \cite{ma2023influential} & \makecell{\cite{barman2025egdl};\\ \cite{pathak2026engression}} \\

        USA ILI & Spatiotemporal & Epidemiology & 50 & Weekly & Weekly influenza-like illnesses (ILI) cases from 50 states in the US. & \newtabnote{tn:ili} & \makecell{\cite{pathak2026engression};\\\cite{panja2026epicastbench}}\\

        Belgium COVID-19 & Spatiotemporal & Epidemiology & 11 & Daily & Daily COVID-19 cases from 11 provinces in Belgium. & \newtabnote{tn:covid} & \makecell{\cite{pathak2026engression};\\\cite{panja2026epicastbench}}  \\

        Colombia Dengue & Spatiotemporal & Epidemiology & 33 & Weekly & Weekly dengue cases from 33 provinces in Colombia. & \cite{clarke2024global} & \makecell{\cite{pathak2026engression};\\\cite{panja2026epicastbench}}\\

        Hungary Chickenpox & Spatiotemporal & Epidemiology & 20 & Weekly & Weekly chickenpox cases from 20 provinces in Hungary. & \cite{rozemberczki2021chickenpox} & \makecell{\cite{pathak2026engression};\\\cite{panja2026epicastbench}}\\

        Shanghai Precipitation & Spatiotemporal & Environmental & 10 & Daily & Daily precipitation (1965-2015) at 0.1 mm resolution for 10 Shanghai stations. & \newtabnote{tn:shanghai} & \cite{qin2025urban}\\

        Delhi AQI & Spatiotemporal & Environmental & 37 & Daily & Air pollution data of 37 stations across Delhi, India. & \cite{panja2026estgcn} & \cite{panja2026estgcn}\\
        
        \bottomrule

    \end{tabular}
    \begin{tablenotes}[para]\footnotesize
        \item[\ref{tn:darts}] \url{https://unit8co.github.io/darts/generated_api/darts.datasets.html};
        \item[\ref{tn:ecg}] \url{http://ecg.mit.edu/time-series/};
        \item[\ref{tn:solar}] \url{https://www.nrel.gov/grid/solar-power-data.html};
        \item[\ref{tn:elec}] \url{https://archive.ics.uci.edu/dataset/321/electricityloaddiagrams20112014};
        \item[\ref{tn:london}] \url{https://www.kaggle.com/jeanmidev/smart-meters-in-london}
        \item[\ref{tn:pemssf}] \url{https://archive.ics.uci.edu/dataset/204/pems+sf};
        \item[\ref{tn:taxi}] \url{https://www.nyc.gov/site/tlc/about/tlc-trip-record-data.page};
        \item[\ref{tn:kdd}] \url{https://www.kdd.org/kdd2018/kdd-cup};
        \item[\ref{tn:wiki}] \url{https://github.com/mbohlkeschneider/gluon-ts/tree/mv_release/datasets};
        \item[\ref{tn:exchange}] \url{https://github.com/mbohlkeschneider/gluon-ts/blob/mv_release/datasets/exchange_rate_nips.tar.gz};
        \item[\ref{tn:pemsbay}] \url{https://dot.ca.gov/programs/traffic-operations/mpr/pems-source};
        \item[\ref{tn:ili}] \url{https://gis.cdc.gov/grasp/fluview/fluportaldashboard.html};
        \item[\ref{tn:covid}] \url{https://epistat.sciensano.be/covid/};
        \item[\ref{tn:shanghai}] \url{https://www.resdc.cn/}
    \end{tablenotes}
    \end{threeparttable}
    \end{adjustbox}
    \label{tab:datasets}
\end{table}

\begin{table}[!ht]
    \raggedleft
    \caption{Overview of prominent time-series and spatiotemporal forecasting resources. Probabilistic evaluation indicates whether the cited release explicitly evaluates predictive distributions. Dataset counts correspond to the cited versions.} 
    \begin{adjustbox}{width=\linewidth}
    \begin{threeparttable}
    \begin{tabular}{cccccccc}
    \toprule
        \multicolumn{1}{c}{Resource} & \multicolumn{1}{c}{Type} & \multicolumn{1}{c}{\makecell{Probabilistic Evaluation}} & \multicolumn{1}{c}{Data Type} & \multicolumn{1}{c}{Domain} & \multicolumn{1}{c}{\#Datasets} & \multicolumn{1}{c}{Repository} & \multicolumn{1}{c}{Reference}  \\ \midrule

        Monash TSF & Data Repo & \xmark & Uni-/Multivariate & Multiple & 20  & \href{https://forecastingdata.org/}{MTSF Repository} & \cite{godahewa2monash}\\

        LTSF & Benchmark & \xmark & Uni-/Multivariate & Multiple & 9  & \href{https://github.com/cure-lab/LTSF-Linear}{LTSF Repository} & \cite{zeng2023transformers} \\

        SynTSBench & Benchmark & \xmark & Uni-/Multivariate & Multiple & - & \href{https://github.com/TanQitai/SynTSBench}{SynTSBench} & \cite{tan2026syntsbench} \\

        EpiCastBench & Benchmark & \xmark & Multivariate & Epidemiology & 40 & \href{https://github.com/aimltsf/EpiCastBench}{EpiCastBench} & \cite{panja2026epicastbench} \\

        BasicTS & Benchmark &  \xmark & Multivariate & Multiple & 20 & \href{https://github.com/GestaltCogTeam/BasicTS}{BasicTS} & \cite{shao2024exploring}  \\

        LargeST &  Benchmark & \xmark & Spatiotemporal & Transportation & 4 & \href{https://github.com/liuxu77/LargeST}{LargeST} & \cite{liu2023largest} \\

        \midrule
        ProbTS & Benchmark &  \cmark & Multivariate & Multiple & 18 & \href{https://github.com/microsoft/ProbTS}{ProbTS}  & \cite{zhang2024probts} \\

        Chronos & Model Eval. &   \cmark & Univariate & Multiple & 42 & \href{https://github.com/amazon-science/chronos-forecasting}{Chronos-forecasting} & \cite{ansari2024chronos} \\

        GIFT-Eval & Benchmark & \cmark & Uni-/multivariate & Multiple & 23 & \href{https://github.com/SalesforceAIResearch/gift-eval}{GIFT-EVAL} & \cite{aksu2024gift}  \\

        fev-bench & Benchmark &  \cmark & Uni-/multivariate & Multiple & 96 & \href{https://github.com/autogluon/fev}{fev} & \cite{shchur2025fev} \\

        Sundial & Model Eval. & \cmark & Multivariate & Multiple & 23 & \href{https://github.com/thuml/Sundial}{Sundial} & \cite{liu2025sundial} \\

        stengression & Model Eval. & \cmark & Spatiotemporal & Epidemiology & 6 & \href{https://github.com/PyCoder913/stengression}{stengression} & \cite{pathak2026engression} \\

        AIATP & Data Repo & \cmark & Spatiotemporal & \makecell{Trajectory Prediction} & 39 & \href{https://github.com/jiachenli94/Awesome-Interaction-Aware-Trajectory-Prediction}{AIATP} & \cite{li2020evolvegraph} \\

        \bottomrule      
    \end{tabular}
    \end{threeparttable}
    \end{adjustbox}
    \label{tab:benchmarks}
\end{table}

\begin{table}[!ht]
    \centering
    \caption{Model evaluation results on the univariate Air Passengers dataset. The \textbf{best} results are highlighted.}
    \begin{adjustbox}{width=\linewidth}
    \begin{tabular}{cccccccccccccc}
    \toprule
        Metric & LSTM-PC & LSTM-BE & LSTM-CP & LSTM-CQR & BNN-MCD & DeepAR & DeepNPTS & LSTM-QR & MVEN & K2VAE & TimesFM-2.5 & Chronos-2 & PatchTSTFM \\ \midrule 
        MAE & 37.9239 & 37.9239 & 33.0622 & 39.2529 & 65.9036 & 53.7476 & 21.7857 & 72.1518 & 39.9051 & \textbf{18.0065} & 24.1588 & 32.9886 & 54.9515 \\ 
        MASE & 1.5747 & 1.5747 & 1.3728 & 1.6298 & 2.7364 & 2.2317 & 0.9046 & 2.9958 & 1.6569 & \textbf{0.7477} & 1.0031 & 1.3697 & 2.2817 \\ 
        RMSSE & 1.3269 & 1.3269 & 1.3351 & 1.5701 & 2.753 & 2.2981 & 1.0258 & 3.1015 & 1.4258 & \textbf{0.7813} & 0.905 & 1.2195 & 2.2387 \\ 
        PICP & 0.5833 & 0.5833 & 0.9167 & 1.0000 & 0.3333 & 0.8333 & 0.9167 & 0.8333 & 0.5833 & 1.0000 & 0.9167 & 1.0000 & 1.0000 \\ 
        CRPS & 27.4542 & 22.8708 & 24.2076 & 40.4744 & 56.4015 & 39.6796 & 16.104 & 56.4844 & 31.8976 & 46.22 & \textbf{15.5024} & 28.8727 & 42.8019 \\ 
        Winkler Score & 337.4479 & 253.2515 & 166.1978 & 455.091 & 1553.8936 & 358.7782 & 130.0086 & 629.789 & 510.5805 & 825.7117 & \textbf{126.1043} & 416.836 & 419.3702 \\ 
        Pinball-80 & 17.301 & 12.9625 & 11.1759 & 18.005 & 38.8983 & 20.0455 & 11.1319 & 28.711 & 19.3661 & 30.5515 & \textbf{5.4353} & 7.0093 & 15.9202 \\ 
        Training time & 10.1565 & 8.7225 & 8.4601 & 8.8623 & \textbf{0.7772} & 6.226 & 22.3078 & 5.5053 & 3.0041 & 12.9991 & 7.4556 & 4.2031 & 7.9223 \\ 
        Inference time & 0.0789 & 0.0735 & 0.2268 & 8.039 & 0.0462 & 0.1354 & 0.0404 & 0.1477 & 0.0463 & \textbf{0.0115} & 2.7472 & 2.2098 & 3.706 \\  \bottomrule
    \end{tabular}
    \end{adjustbox}
    \label{tab:results_airpassengers_data}
\end{table}

\begin{table}[!ht]
    \centering
    \caption{Model evaluation results on the multivariate PEMS-SF traffic dataset. The \textbf{best} results are highlighted.}
    \begin{adjustbox}{width=\linewidth}
    \begin{tabular}{cccccccccccccc}
    \toprule
        Metric & LSTM-PC & LSTM-BE & BNN-MCD & DeepAR & DeepNPTS & LSTM-QR & MVEN & Transformer-MAF & TimeGrad & K2VAE & TimesFM-2.5 & Chronos-2 & PatchTSTFM \\ \midrule
        MAE & \textbf{0.0103} & \textbf{0.0103} & 0.0107 & 0.0209 & 0.022 & 0.0108 & \textbf{0.0103} & 0.0308 & 0.0119 & 0.025 & 0.0134 & 0.0196 & 0.0269 \\ 
        MASE & 0.8765 & 0.8765 & 0.9057 & 1.7673 & 1.8622 & 0.9154 & \textbf{0.8740} & 2.6096 & 1.0068 & 2.1182 & 1.1339 & 1.6634 & 2.2799 \\ 
        RMSSE & 0.9731 & 0.9731 & 0.9675 & 1.4404 & 1.4668 & 1.0194 & \textbf{0.9508} & 2.6623 & 1.0296 & 1.5296 & 1.136 & 1.435 & 1.6535 \\ 
        PICP & 0.9699 & 0.9547 & 0.9114 & 0.9746 & 0.8596 & 0.928 & 0.9482 & 0.8 & 0.946 & 0.8889 & 0.9232 & 0.9941 & 0.9813 \\ 
        CRPS & 0.0099 & 0.0088 & 0.0085 & 0.0157 & 0.0163 & \textbf{0.0077} & 0.0082 & 0.025 & 0.009 & 0.0411 & 0.0098 & 0.0109 & 0.0181 \\ 
        Winkler Score & 0.158 & 0.1549 & 0.1534 & 0.1961 & 0.1816 & \textbf{0.0967} & 0.1306 & 0.3465 & 0.1169 & 0.6587 & 0.1316 & 0.2316 & 0.2429 \\ 
        Pinball-80 & 0.0061 & 0.0051 & 0.0051 & 0.009 & 0.009 & 0.0048 & \textbf{0.0047} & 0.0137 & 0.0053 & 0.0264 & 0.0063 & 0.0078 & 0.0113 \\ 
        Training time (sec.) & 83.7331 & 88.5309 & 16.2234 & 120.8953 & 22.4629 & 169.8363 & 1418.9248 & 390.5839 & 284.8944 & 94.2103 & 7.4582 & \textbf{4.7971} & 8.3082 \\ 
        Inference time (sec.) & 0.0805 & 0.0839 & 0.0948 & 0.1962 & 1.3771 & 0.3417 & \textbf{0.0514} & 26.7843 & 16.2917 & 0.0631 & 186.4986 & 661.4305 & 715.2338 \\ \bottomrule
    \end{tabular}
    \end{adjustbox}
    \label{tab:results_pems_data}
\end{table}

\begin{table}[!ht]
    \centering
    \caption{Model evaluation results on the spatiotemporal Belgium COVID-19 dataset. The \textbf{best} results are highlighted.}
    \begin{adjustbox}{width=\linewidth}
    \begin{tabular}{ccccccccccccccccc}
    \toprule
        Metric & LSTM-PC & LSTM-BE & LSTM-CP & LSTM-CQR & BNN-MCD & GpGp & DeepAR & DeepNPTS & LSTM-QR & GCEN & Transformer-MAF & TimeGrad & K2VAE  & TimesFM-2.5 & Chronos-2 (QR) & PatchTSTFM \\ \midrule
        MAE & 97.023 & 97.023 & 120.0787 & 121.3414 & 152.892 & 118.5176 & 65.0088 & 74.4091 & 115.6297 & \textbf{55.2497} & 66.6075 & 71.5354 & 98.1925 & 68.3451 & 61.3654 & 66.725 \\ 
        MASE & 0.4954 & 0.4954 & 0.6131 & 0.6196 & 0.7807 & 0.6051 & 0.3319 & 0.3799 & 0.5904 & \textbf{0.2821} & 0.3401 & 0.3653 & 0.5014 & 0.349 & 0.3133 & 0.3407 \\ 
        RMSSE & 0.2706 & 0.2706 & 0.327 & 0.3519 & 0.4118 & 0.3403 & 0.2009 & 0.2245 & 0.3172 & \textbf{0.1673} & 0.2228 & 0.2029 & 0.3081 & 0.1981 & 0.1832 & 0.1979 \\ 
        PICP & 0.9939 & 0.9758 & 1 & 1 & 0.2727 & 1 & 0.8242 & 0.8121 & 0.9727 & 0.8606 & 0.6 & 0.5636 & 0.9364 & 0.7394 & 0.9818 & 1 \\ 
        CRPS & 71.8659 & 67.2433 & 115.8783 & 103.3416 & 133.6601 & 499.112 & 48.3278 & 54.4688 & 84.7214 & \textbf{42.2133} & 55.2931 & 54.6017 & 80.151 & 50.3264 & 44.2722 & 51.3223 \\ 
        Winkler Score & 676.6248 & 692.9498 & 2741.2335 & 3036.1968 & 3718.1841 & 7380.7593 & 489.1249 & 666.5125 & 665.1303 & 563.2885 & 1230.1843 & 762.1248 & 983.2138 & 612.5202 & \textbf{446.9277} & 593.5463 \\ 
        Pinball-80 & 40.7961 & 37.0838 & 53.7525 & 46.0443 & 44.6283 & 246.8415 & 28.6325 & 27.7178 & 49.0047 & \textbf{21.3606} & 39.6408 & 22.2583 & 40.9953 & 29.1404 & 23.7537 & 26.8285 \\ 
        Training time & 48.1703 & 49.3807 & 24.5544 & 25.3741 & 3.676 & 90.9597 & 32.2584 & 6.3141 & 28.6281 & 81 & 369.4191 & 232.8301 & 63.3808 & 2.065 & \textbf{1.2000} & 2.2096 \\ 
        Inference time & 0.1188 & 0.087 & 0.3481 & 56.8413 & 0.0714 & 11.9556 & 0.1235 & 0.048 & 0.1054 & 0.1912 & 0.5926 & 16.466 & \textbf{0.0164} & 6.6694 & 11.9143 & 17.7775 \\ \bottomrule
    \end{tabular}
    \end{adjustbox}
    \label{tab:results_belgium_data}
\end{table}

\begin{table}[!t]
\fontsize{8}{9}\selectfont
\caption{Summary of open-source implementations for probabilistic forecasting methods. A dash (-) indicates methods without an explicit name in their original publication.}

\begin{adjustbox}{width=\textwidth, max height=\textheight}
   \centering
    \begin{tabular}{ccccc} \toprule
    Method Name & Data Type & Environment & Repository & Reference\\ 
    \multicolumn{5}{l}{\cellcolor{AgnosticDark}\textbf{\textcolor{white}{%
  Model-Agnostic}}}\\

   \ragnostic CSP & Univariate & \texttt{Python} & \url{https://github.com/valeman/csp-forecaster} & \cite{valery2026csp}\\

   \ragnostic FEWNet & Univariate & \texttt{R} & \url{https://github.com/ctanujit/FEWNet} & \cite{sengupta2025fewnet}\\

   \ragnostic EnbPI & Univariate & \texttt{Python} & \url{https://github.com/hamrel-cxu/EnbPI} & \cite{xu2021conformal}\\

   \ragnostic EnCQR & Multivariate & \texttt{Python} & \url{https://github.com/FilippoMB/Ensemble-Conformalized-Quantile-Regression} & \cite{jensen2024ensqr}\\

   \ragnostic STACI & Spatiotemporal & \texttt{Python} & \url{https://github.com/bf5124/STACI} & \cite{feng2026staci}\\

   \ragnostic S-ESN & Spatiotemporal & \texttt{Python} & \url{https://github.com/hhuang90/KSA-wind-forecast} & \cite{huang2022esn}\\

   \ragnostic D-EESN & Spatiotemporal & \texttt{R} & \url{https://github.com/PatrickMcDermottResearch/BD-EESN} & \cite{McDermott2019deen}\\

   \ragnostic QESN  & Spatiotemporal & \texttt{R} & \href{https://doi.org/10.1002/sta4.160}{Wiley Online Library} & \cite{McDermott2017qesn}\\

    \multicolumn{5}{l}{\cellcolor{BayesianDark}\textbf{\textcolor{white}{%
  Bayesian Modeling}}}\\

    \rbayesian BNN & Univariate & \texttt{R} & \url{https://github.com/nhauzenb/hhkm-joe-bnn} & \cite{Hauzenberger2025bnn}\\

    \rbayesian BARNN & Univariate & \texttt{Python} & \url{https://github.com/dario-coscia/barnn} & \cite{coscia2025barnn}\\

    \rbayesian BSTS & Univariate & \texttt{R} & \url{https://github.com/cran/bsts} & \cite{scott2014bsts}\\

    \rbayesian BSTS & Univariate & \texttt{Python} & \url{https://github.com/Focus/bsts}\textsuperscript{$\dagger$} & \cite{scott2014bsts}\\

    \rbayesian Bayesian LSTM & Multivariate & \texttt{Python} & \url{https://github.com/patichandanareddy/bayesian-lstm}\textsuperscript{$\dagger$} & \cite{hassan2024bitcoin}\\

    \rbayesian MBSTS & Multivariate & \texttt{R} & \url{https://github.com/cran/mbsts} & \cite{qiu2018mbsts}\\

    \rbayesian LKGP & Spatiotemporal & \texttt{Python} & \url{https://github.com/jandylin/Latent-Kronecker-GPs} & \cite{lin2025scalable}\\

    \rbayesian BayesNF & Spatiotemporal & \texttt{Python} & \url{https://github.com/google/bayesnf} & \cite{saad2024scalable}\\

    \rbayesian - & Spatiotemporal & \texttt{R} & \url{https://github.com/mindra-bit/HighResolutionForecasting} & \cite{jaya2024pm}\\

    \rbayesian Space-Time.DeepKriging & Spatiotemporal & \texttt{R} & \url{https://github.com/pratiknag/Space-Time.DeepKriging} & \cite{nag2023DeepKriging}\\

    \rbayesian VLR-STRF & Spatiotemporal & \texttt{MATLAB} & \url{https://github.com/pillowlab/VLR-STRF} & \cite{duncker2023scalable}\\

    \rbayesian - & Spatiotemporal & \texttt{R} & \url{https://github.com/swchouchen/DengueCR_Bayesian_ST_Prediction} & \cite{chen2023prob}\\

    \rbayesian AGCGRU+flow & Spatiotemporal & \texttt{Python} & \url{https://github.com/networkslab/rnn_flow} & \cite{pal2021rnnpf}\\

    \rbayesian ST-SVGP & Spatiotemporal & \texttt{Python} & \url{https://github.com/AaltoML/spatio-temporal-GPs} & \cite{hamelijnck2021stvgp}\\

    \rbayesian BSTIM & Spatiotemporal & \texttt{Python} & \url{https://github.com/ostojanovic/BSTIM} & \cite{stojanovic2019bayesian}\\

    \rbayesian BD-EESN & Spatiotemporal & \texttt{R} & \url{https://github.com/PatrickMcDermottResearch/BD-EESN} & \cite{McDermott2019deen}\\

    \rbayesian - & Spatiotemporal & \texttt{MATLAB} & \href{https://onlinelibrary.wiley.com/action/downloadSupplement?doi=10.1111%2Fbiom.12844&file=biom12844-sup-0002-SuppfMRI_space-time.zip}{Wiley Online Library} & \cite{castruccio2018spatiotemporalbrain}\\

    \rbayesian GpGp & Spatiotemporal & \texttt{R} & \url{https://github.com/cran/GpGp} & \cite{guinness2018gpgp}\\

    \rbayesian Sticky-HDP-HMM & Multivariate & \texttt{Python} & \url{https://github.com/MoonBlvd/Sticky-HDP-HMM} & \cite{fox2011sticky}\\

    \multicolumn{5}{l}{\cellcolor{ParametricDark}\textbf{\textcolor{white}{%
  Parametric Predictive Distributions}}}\\

  \rparametric WA-IMDN  & Univariate & \texttt{Python} & \url{https://github.com/IkeYang/WA-IMDN-} & \cite{yang2022waimdn}\\

  \rparametric -  & Multivariate & \texttt{Python} & \url{https://github.com/rottenivy/mv_pts_correlatederr} & \cite{zheng2024correlated}\\

  \multicolumn{5}{l}{\cellcolor{DistRegDark}\textbf{\textcolor{white}{%
  Distributional Regression}}}\\

  \rdistreg QRNN  & Univariate & \texttt{R} & \url{https://github.com/cran/qrnn} & \cite{graves2011practical}\\

  \rdistreg MQRNN   & Univariate & \texttt{Python} & \url{https://github.com/ErezSC42/qr_forcaster} & \cite{wen2018mqrnn}\\

  \rdistreg NMQN   & Univariate & \texttt{R} & \url{https://github.com/Monster-Moon/l1pm} & \cite{moon2021learning}\\

  \rdistreg DQA & Multivariate & \texttt{Python} & \url{https://github.com/amazon-science/quantile-aggregation} & \cite{fakoor2023flexible}\\

  \rdistreg LSTM-Engression & Multivariate & \texttt{Python} & \url{https://github.com/bask0/mach-flow-engression} & \cite{kraft2026engression}\\

  \rdistreg Enformer & Multivariate & \texttt{Python} & \url{https://github.com/yuvrajiro/Genformer} & \cite{pathak2026entransformer}\\

  \rdistreg Tube Loss & Multivariate & \texttt{Python} & \url{https://github.com/ltpritamanand/Tube_loss} & \cite{anand2026tube}\\

  \rdistreg DESQRUQ & Spatiotemporal & \texttt{Python} & \url{https://github.com/tanwimallick/DESQRUQ} & \cite{mallick2024desqruq}\\

  \rdistreg EQRN  & Spatiotemporal & \texttt{R} & \url{https://github.com/cran/EQRN} & \cite{pasche2024eqrn}\\

  \rdistreg GCEN/STEN & Spatiotemporal & \texttt{Python} & \url{https://github.com/PyCoder913/stengression} & \cite{pathak2026engression}\\

  \multicolumn{5}{l}{\cellcolor{GenerativeDark}\textbf{\textcolor{white}{%
  Generative Models}}}\\

  \rgenerative InfHMM & Univariate & \texttt{R} & \url{https://github.com/frimane/InfHMM-Infinit-Hidden-Markov-Model} & \cite{frimane2022infhmm}\\

  \rgenerative C2FAR & Univariate & \texttt{Python} & \url{https://github.com/huaweicloud/c2far_forecasting} & \cite{bergsma2022c2far}\\

  \rgenerative PPM & Multivariate & \texttt{Python}  & \url{https://github.com/ljl8336/PPM} & \cite{li2026ppm}\\

  \rgenerative TimePrism & Multivariate & \texttt{Python} & \url{https://github.com/Fifthky/TimePrism} & \cite{dai2026timeprism}\\

  \rgenerative DGFC & Multivariate & \texttt{R} & \url{https://github.com/johnczito/DynamicCopula} & \cite{zito2025dynamic}\\

  \rgenerative HDT & Multivariate & \texttt{Python} & \url{https://github.com/autonomousvision/hdt} & \cite{shibo2025hdt}\\

  \rgenerative K$^2$VAE & Multivariate & \texttt{Python} & \url{https://github.com/decisionintelligence/K2VAE} & \cite{wu2025kvae}\\

  \rgenerative - & Multivariate & \texttt{Python} & \url{https://github.com/lambertdem/geometricExtremesNF} & \cite{demonte2025generative}\\

  \rgenerative ProFITi & Multivariate & \texttt{Python} & \url{https://github.com/yalavarthivk/ProFITi} & \cite{yalavarthi2025ProFITi}\\

  \rgenerative NsDiff & Multivariate & \texttt{Python} & \url{https://github.com/wwy155/NsDiff} & \cite{ye2025nsdiff}\\

  \rgenerative TimeMCL & Multivariate & \texttt{Python} & \url{https://github.com/Victorletzelter/timeMCL} & \cite{cortes2025timemcl}\\

  \rgenerative TACTiS-2 & Multivariate & \texttt{Python} & \url{https://github.com/ServiceNow/TACTiS} & \cite{ashok2024tactis}\\

  \rgenerative CopulaCPTS & Multivariate & \texttt{Python} & \url{https://github.com/Rose-STL-Lab/CopulaCPTS} & \cite{sun2024copula}\\

  \rgenerative E-ProTran & Multivariate & \texttt{Python} & \url{https://github.com/bkoyuncu/eprotan} & \cite{koyuncu2024eprotran}\\

  \rgenerative TMDM & Multivariate & \texttt{Python} & \url{https://github.com/LiYuxin321/TMDM} & \cite{li2024transformermodulated}\\

  \rgenerative MG-TSD & Multivariate & \texttt{Python} & \url{https://github.com/Hundredl/MG-TSD} & \cite{fan2024mgtsd}\\

  \rgenerative D$^3$VAE & Multivariate & \texttt{Python} & \url{https://github.com/PaddlePaddle/PaddleSpatial/tree/main/research/D3VAE} & \cite{li2022d3vae}\\

  \rgenerative Transformer-MAF & Multivariate & \texttt{Python} & \url{https://github.com/zalandoresearch/pytorch-ts} & \cite{rasul2021multivariate}\\

  \rgenerative TimeGrad & Multivariate & \texttt{Python} & \url{https://github.com/zalandoresearch/pytorch-ts} & \cite{rasul2021timegrad}\\

  \rgenerative ScoreGrad & Multivariate & \texttt{Python} & \url{https://github.com/yantijin/ScoreGradPred} & \cite{song2021score}\\

  \rgenerative GP-Copula  & Multivariate & \texttt{Python} & \url{https://github.com/mbohlkeschneider/gluon-ts/tree/mv_release} & \cite{salinas2019gpcopula}\\

  \rgenerative SpecSTG & Spatiotemporal & \texttt{Python} & \url{https://github.com/lequanlin/SpecSTG} & \cite{lin2025SpecSTG}\\

  \rgenerative GenCast & Spatiotemporal & \texttt{Python} & \url{https://github.com/google-deepmind/weathernext} & \cite{price2024gencast}\\

  \rgenerative SHADECast & Spatiotemporal & \texttt{Python} & \url{https://github.com/EnergyWeatherAI/GenerativeNowcasting} & \cite{Carpentieri2025SHADECast}\\

  \rgenerative - & Spatiotemporal & \texttt{Python} & \url{https://github.com/LoryPack/GenerativeNetworksScoringRulesProbabilisticForecasting} & \cite{pacchiardi2024gan}\\

  \rgenerative STMPNet & Spatiotemporal & \texttt{Python} & \url{https://github.com/yang-1201/STMPNet} & \cite{an2025stmpnet}\\

  \rgenerative Graph-EFM & Spatiotemporal & \texttt{Python}  & \url{https://github.com/mllam/neural-lam/tree/prob_model_global} & \cite{oskarsson2024probabilistic}\\

  \rgenerative MotionFlow & Spatiotemporal & \texttt{Python}  & \url{https://github.com/MohsenZand/MotionFlow} & \cite{Zand2023stsp}\\

  \rgenerative DiffSTG & Spatiotemporal & \texttt{Python}  & \url{https://github.com/wenhaomin/DiffSTG} & \cite{wen2023diffstg}\\

  \rgenerative DYffusion & Spatiotemporal & \texttt{Python}  & \url{https://github.com/Rose-STL-Lab/dyffusion} & \cite{cachay2023dyffusion}\\

  \rgenerative Social GAN & Spatiotemporal & \texttt{Python}  & \url{https://github.com/agrimgupta92/sgan} & \cite{gupta2018social}\\

  \multicolumn{5}{l}{\cellcolor{FoundationDark}\textbf{\textcolor{white}{%
  Foundation Models}}}\\
  \rfoundation Lag-Llama & Univariate & \texttt{Python}  & \url{https://github.com/time-series-foundation-models/lag-llama} & \cite{rasul2023lag}\\
  \rfoundation TiRex & Univariate & \texttt{Python}  & \url{https://github.com/NX-AI/tirex} & \cite{auer2026tirex}\\
\rfoundation Moirai & Multivariate & \texttt{Python}  & \url{https://github.com/redoules/moirai} & \cite{liu2024moirai}\\
  \rfoundation TimesFM & Multivariate & \texttt{Python}  & \url{https://github.com/google-research/timesfm} & \cite{das2024timesfm}\\
  \rfoundation Chronos-2 & Multivariate & \texttt{Python}  & \url{https://github.com/amazon-science/chronos-forecasting} & \cite{ansari2024chronos}\\  \bottomrule

    \end{tabular}
    \label{tab:UQ_Methods_Implementation}
    \end{adjustbox}
    \footnotesize{\textsuperscript{$\dagger$}Third-party implementation; not created by the paper's authors.}
\end{table}

\end{document}